\documentclass{article} 
\usepackage{iclr2026_conference,times}

\usepackage{amsmath,amsfonts,bm}

\def\eqref#1{equation~\ref{#1}}

\def\1{\bm{1}}

\def\ra{{\textnormal{a}}}

\def\rx{{\textnormal{x}}}

\def\rva{{\mathbf{a}}}

\def\erva{{\textnormal{a}}}

\def\ervx{{\textnormal{x}}}

\def\rmA{{\mathbf{A}}}

\def\vmu{{\bm{\mu}}}
\def\vtheta{{\bm{\theta}}}
\def\va{{\bm{a}}}

\def\ve{{\bm{e}}}

\def\vx{{\bm{x}}}

\def\eva{{a}}

\def\mA{{\bm{A}}}

\def\mH{{\bm{H}}}
\def\mI{{\bm{I}}}
\def\mJ{{\bm{J}}}

\def\mX{{\bm{X}}}

\def\mSigma{{\bm{\Sigma}}}

\DeclareMathAlphabet{\mathsfit}{\encodingdefault}{\sfdefault}{m}{sl}
\SetMathAlphabet{\mathsfit}{bold}{\encodingdefault}{\sfdefault}{bx}{n}
\newcommand{\tens}[1]{\bm{\mathsfit{#1}}}
\def\tA{{\tens{A}}}

\def\tX{{\tens{X}}}

\def\gG{{\mathcal{G}}}

\def\sA{{\mathbb{A}}}
\def\sB{{\mathbb{B}}}

\def\sS{{\mathbb{S}}}

\def\emA{{A}}

\newcommand{\etens}[1]{\mathsfit{#1}}

\def\etA{{\etens{A}}}

\newcommand{\E}{\mathbb{E}}

\newcommand{\R}{\mathbb{R}}

\newcommand{\KL}{D_{\mathrm{KL}}}
\newcommand{\Var}{\mathrm{Var}}

\newcommand{\Cov}{\mathrm{Cov}}

\newcommand{\normltwo}{L^2}
\newcommand{\normlp}{L^p}

\newcommand{\parents}{Pa} 

\DeclareMathOperator*{\argmax}{arg\,max}

\usepackage[utf8]{inputenc} 
\usepackage[T1]{fontenc}    
\usepackage{url}            
\usepackage[english]{babel}
\usepackage{kotex}
\usepackage{booktabs}       
\usepackage{multicol,multirow}
\usepackage[svgnames,x11names,dvipsnames,table]{xcolor}
\usepackage{makecell}
\usepackage{colortbl}
\usepackage{diagbox}
\usepackage{slashbox}
\usepackage{wrapfig}
\usepackage{caption}
\usepackage{tabularx}
\usepackage{float}
\usepackage{enumitem}
\newcommand*\fullcirc[1][1ex]{\tikz\fill (0,0) circle (#1);} 
\usepackage{mathrsfs}
\usepackage{mathtools}
\usepackage{pifont}
\usepackage[misc]{ifsym}
\usepackage{amssymb, amsmath, amsfonts, bm}

\usepackage{enumitem, bm, multirow, array, pifont, 
            xspace, colortbl, booktabs ,tabularx, placeins, soul, makecell, multicol, wrapfig,
            nicefrac, scalerel, pgfplots, marginnote, subcaption, outlines, capt-of, amsmath, hhline}
\pgfplotsset{compat=1.17} 

\usepackage[labelformat=simple]{subcaption}
\usepackage[linesnumbered,ruled,vlined]{algorithm2e}    
\SetKwInput{KwData}{Data} 
\SetKwInput{KwInput}{Input} 
\SetKwInput{KwOutput}{Output} 
\definecolor{comment-color}{rgb}{0.203921568627451,0.541176470588235,0.841176470588235}

\SetCommentSty{mycommfont}
\SetAlFnt{\small}
\SetAlCapFnt{\small}
\SetAlCapNameFnt{\small}
\usepackage{siunitx}
\usepackage{color}
\usepackage{alltt}
\usepackage{bbm}
\usepackage{graphicx}
\usepackage{nicefrac}
\usepackage{microtype}
\usepackage{lipsum}
\usepackage{tcolorbox}
\usepackage{xurl}

\usepackage{tikz}
\usepackage{pifont}
\newcommand{\cmark}{\checkmark}
\newcommand{\xmark}{\ding{55}}
\usepackage{xspace}
\newcommand*{\ie}{i.e.\@\xspace}
\newcommand*{\eg}{e.g.\@\xspace}

\usepackage{booktabs,tabularx,makecell,array}
\newcolumntype{Y}{>{\raggedright\arraybackslash}X}
\usepackage{calc}   
\usepackage{lipsum}

\definecolor{blue}{rgb}{0.2472,0.24,0.6}
\definecolor{lightblue}{RGB}{65,105,225} 
\usepackage[pagebackref,breaklinks,colorlinks,linkcolor=red,citecolor=lightblue,urlcolor=lightblue]{hyperref}

\usepackage{fontawesome5}

\usepackage{tcolorbox}
\tcbset{
  mycallout/.style={
    colback=gray!5,
    colframe=black!70,
    boxrule=0.8pt,
    arc=2pt,
    left=8pt,right=8pt,top=6pt,bottom=6pt,
  }
}

\DeclareRobustCommand{\loongrightarrow}{%
  \DOTSB\relbar\joinrel\relbar\joinrel\rightarrow
}

\usepackage[capitalize]{cleveref} 

\title{Improving Black-Box Generative Attacks via Generator Semantic Consistency}

\author{%
    Jongoh Jeong$^{1}$,
    Hunmin Yang$^{1,2}$,
    Jaeseok Jeong$^{1}$, and
    Kuk-Jin Yoon$^{1}$, \\
    $^{1}$Visual Intelligence Lab., 
    Korea Advanced Institute of Science and Technology (KAIST) \\
    $^{2}$Agency for Defense Development (ADD) \\
    \texttt{\{jeong2, hmyang, jason.jeong, kjyoon\}@kaist.ac.kr}
    \hfill
    \href{https://andyj1.github.io/scga}{\textcolor{lightblue}{\small \faGithub\;Project Page}}
}

\iclrfinalcopy 
\begin{document}

\maketitle
\begin{abstract}
Transfer attacks optimize on a surrogate and deploy to a black-box target. While iterative optimization attacks in this paradigm are limited by their per-input cost limits efficiency and scalability due to multistep gradient updates for each input, generative attacks alleviate these by producing adversarial examples in a single forward pass at test time. However, current generative attacks still adhere to optimizing surrogate losses (e.g., feature divergence) and overlook the generator’s internal dynamics, underexploring how the generator’s internal representations shape transferable perturbations. To address this, we enforce semantic consistency by aligning the early generator’s intermediate features to an 
exponential moving average (EMA)
teacher, stabilizing object-aligned representations and improving black-box transfer without inference-time overhead. To ground the mechanism, we quantify semantic stability as the standard deviation of foreground IoU between cluster-derived activation masks and foreground masks across generator blocks, and observe reduced semantic drift under our method. For more reliable evaluation, we also introduce Accidental Correction Rate (ACR) to separate inadvertent corrections from intended misclassifications, complementing the inherent blind spots in traditional Attack Success Rate (ASR), Fooling Rate (FR), and Accuracy metrics. Across architectures, domains, and tasks, our approach can be seamlessly integrated into existing generative attacks with consistent improvements in black-box transfer, while maintaining test-time efficiency. 
Code at \textcolor{lightblue}{\href{https://github.com/andyj1/scga}{https://github.com/andyj1/scga}}.


\end{abstract}

\section{Introduction}
\label{sec:introduction}

Deep neural networks have driven advances in computer vision, natural language processing, and medical diagnosis by learning rich hierarchical representations. At the same time, they remain vulnerable to small human-imperceptible perturbations known as adversarial examples (AE)~\cite{szegedy2013intriguing}, which can induce confident misclassification and raise safety concerns in real deployments. The risk is amplified in black-box settings, where an attacker has no access to the parameters or architecture of a model. In these scenarios, transfer-based attacks craft perturbations on a surrogate and deploy them against unseen targets, enabling a single perturbation strategy to threaten diverse safety-critical systems such as self-driving and biometrics.

Early white-box iterative attacks (for example, FGSM and its multistep variants~\cite{zhang2021fgsm,dong2018mifgsm,DI,TI}) rely on direct gradient access. Transfer-based attacks extend this idea by seeking perturbations that generalize across models, often using iterative optimization in the white-box regime~\cite{madry2017pgd,carlini2019c_and_w,zhang2021fgsm}. While effective, they require per-example iterative optimization, whereas generative attacks amortize this cost by producing perturbations in a single forward pass.

Generative transfer attacks train a feedforward perturbation generator against a surrogate and then produce adversarial noise with one forward pass at the test time~\cite{xiao2018generatingadvgan,wang2018perceptual,baluja2017adversarial,baluja2018learning,poursaeed2018generative,naseer2019cross,salzmann2021learning,zhang2022beyond,aich2022gama,yang2024facl,yang2024pdcl,nakka2025nat}. This design yields fast inference and strong scalability. However, current generative attacks are centered around optimizing the surrogate-level objectives and treat the generator merely as a tool to generate adversarial examples given the adversarial objective, overlooking the progressive AE synthesis process in which perturbations are incrementally formed block by block within the generator. This oversight leaves potential for improving transferability, as the intermediate blocks of the generator are where semantic structure, such as the contour of the object and the coarse shape, is preserved or degraded during synthesis~\cite{zhang2022beyond}. As a result, perturbations may disperse onto object-irrelevant regions that are relatively less
victim model
-agnostic, weakening adversarial transferability.
This critically raises the following questions: 

\noindent
\tcbox[
  mycallout,
  on line=false,         
  nobeforeafter,
  width=\linewidth,      
  boxsep=0pt             
]{
  \parbox{\dimexpr\linewidth-5\fboxsep\relax}{%
  Q1. \textit{At what stage of perturbation synthesis do semantic cues deteriorate?}\\
  Q2. \textit{Which generator blocks most influence transferability?}
  }%
}

To investigate the perturbation synthesis in detail, we partition the six intermediate blocks in the generator into three split blocks--early, mid, and late--and find that the early blocks better preserve object-aligned structure than later ones. We substantiate this claim with a diagnostic analysis of the stability of object-aligned perturbation semantics within the generator intermediate blocks  
As in Fig.~\ref{fig:motivation}, lower cross-block variability, and thus higher consistency of object semantics, is associated with higher transferability of AEs.


Guided by this observation, we propose a semantically consistent generative attack (SCGA) that explicitly targets semantic consistency during perturbation synthesis within the generator. Concretely, we use a Mean Teacher pathway in which an Exponential Moving Average (EMA)-updated teacher provides temporally smoothed reference features, and a self-feature consistency loss aligns the student’s early generator block activations with these references while keeping the adversarial objective on the surrogate features unchanged, as shown in Fig.~\ref{fig:overview}. This guidance operates only during training without additional test-time cost, and integrates with existing generative attacks. 

Finally, we broaden the evaluation beyond misclassification-based metrics (ASR, FR) and a correction-based metric (Accuracy) to include our proposed Accidental Correction Rate (ACR). For reliable evaluations, ACR complements these conventional metrics by identifying cases that are inherently likely to be overlooked, such as unintended corrections of initially wrong benign predictions. In a comprehensive evaluation setting, we demonstrate that the internal dynamics within the generator play a critical role in enhancing adversarial transferability between domains, models, and even tasks. 
We summarize our main contributions as follows:
\vspace{-2mm}
\begin{itemize}[leftmargin=*] \setlength\itemsep{-0.1mm}
    \item \textbf{Generator–internal evidence for perturbation semantics.} To investigate perturbation semantics within the generator, we partition the generator into early/mid/late blocks and quantify object-aligned semantics per block. Our analysis reveals that methods with lower variability in the foreground IoU across the intermediate blocks exhibit higher adversarial transfer. (\S\ref{sec:motivation})
    \item \textbf{Generator–level semantic consistency guidance.} 
      By enforcing training-only semantic consistency at the generator's \emph{early} intermediates, we achieve improved adversarial transfer while keeping the adversarial objective on the surrogate unchanged. The guidance can be seamlessly integrated into existing generative attacks without altering the test pipeline at no additional inference cost. (\S\ref{sec:semantic_structure_aware_generative_attack})
    \item \textbf{Comprehensive evaluation with an added reliability measure.} 
      We conduct a comprehensive transferability evaluation spanning classification (CLS) across architectures, domains, and dense prediction tasks (SS, OD). We also complement conventional Accuracy, ASR, and FR metrics by introducing a novel ACR metric to assess the attack reliability, measured by inadvertent corrections from intended misclassifications.
      (§\ref{sec:evaluation_metrics}).

\end{itemize}

\section{Background and Motivation} \label{sec:preliminaries}

\subsection{Preliminaries}
Given a pre-trained victim model $\mathcal{F}^{t}(\cdot)$ evaluated on a test distribution $\mathcal{D}_{\text{test}}$, the objective is to synthesize human-imperceptible perturbations that transfer across models, domains, and tasks, using only a source domain $\mathcal{D}_{\text{src}}$ and its pre-trained models as substitutes. Generative attack framework employs a generator $\mathcal{G}_{\theta}(\cdot)$ that maps a benign input $x$ to an unconstrained adversarial candidate $\tilde{x}^{\text{adv}}$, followed by a projector $\mathcal{P}(\cdot)$ that enforces the $\ell_{\infty}$-budget, i.e., $\lVert \mathcal{P}(\tilde{x}^{\text{adv}})-x\rVert_{\infty}\le \epsilon$.
Training of $\mathcal{G}_{\theta}(\cdot)$ is supervised in white-box fashion by a surrogate model $\mathcal{F}^{s}(\cdot)$ trained on $\mathcal{D}_{\text{src}}$, enabling gradient-based updates via backpropagation. The adversarial loss leverages surrogate logits or intermediate features of $\mathcal{F}^{s}(\cdot)$, \eg at layer $k$, to capture model-shared characteristics known to enhance black-box transferability \cite{naseer2019cross, salzmann2021learning, zhang2022beyond, aich2022gama, yang2024facl, yang2024pdcl}. Formally, $\mathcal{G}_{\theta}(\cdot)$ is optimized to generate AEs that maximize evaluation metrics against victim models $\mathcal{F}^{t}(\cdot)$ and/or relative to ground-truth labels $y$ with:
\vspace{-1mm}
\begin{equation}
    \mathtt{Metric}\Big(x, x_{adv}, \mathcal{F}^{t}(\cdot),y \Big),\;\;\hfill \textrm{with}\;\;\Vert x_{adv}-x \Vert_{\infty} \leq \epsilon, \qquad \footnotesize{\text{(See \S\ref{sec:evaluation_metrics} for metric details.)}}
    \vspace{-1.5mm}
\end{equation}
\noindent where $\epsilon$ denotes the maximum perturbation budget that guarantees a minimal change in $x$. Here, \texttt{Metric} refers to ASR, FR, Acc., and ACR for classification (CLS); mIoU and mAP50 for semantic segmentation (SS) and object detection (OD), respectively.

\subsection{Perturbation semantics in generator-internal dynamics} \label{sec:motivation}

\begin{table*}[!ht]
\begin{minipage}[t]{0.42\linewidth}
    \centering
    \vspace{0pt}
    \includegraphics[width=\linewidth]{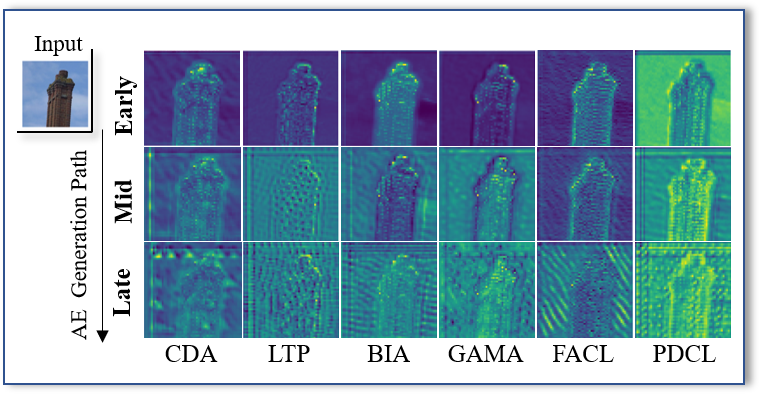}
    \vspace{-7mm} \caption*{(a)}
\end{minipage}
\begin{minipage}[t]{0.18\linewidth}
    \centering
    \vspace{0pt}
    \includegraphics[width=1.05\linewidth]{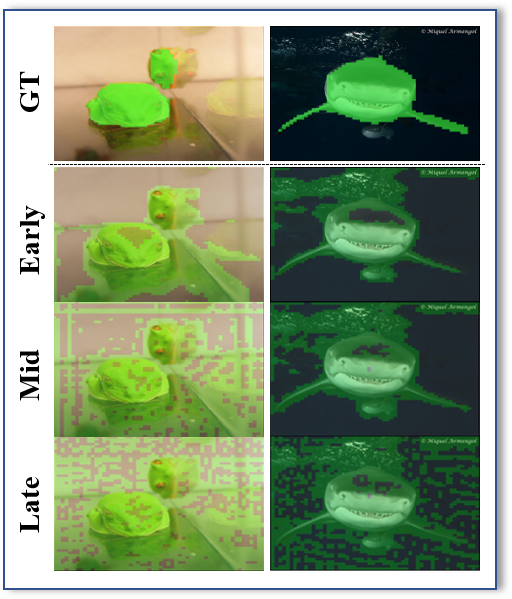}
    \vspace{-7mm} \caption*{(b)}
\end{minipage}
\hfill
\begin{minipage}[t]{0.35\linewidth}
    \centering
    \vspace{3pt}
    \setlength{\tabcolsep}{1pt}
    \renewcommand{\arraystretch}{1}
    \renewcommand{\aboverulesep}{4pt}
    \renewcommand{\belowrulesep}{4pt}
      
      
      
      
      
    \resizebox{\linewidth}{!}{%
        \begin{tabular}{lccccccc}
      \toprule
          \multicolumn{4}{l}{\texttt{Feature Cluster}} \\
          \shortstack[l]{\texttt{Foreground}\\\texttt{IoU}} & \multicolumn{3}{c}{\textbf{Intermediate block}} & \;\;\; & \multicolumn{3}{c}{\shortstack[c]{\textbf{Std.Dev.}\\(Variability)} \textcolor{ForestGreen}{$\downarrow$}} \\ \cmidrule{2-4} \cmidrule{6-8}
          \textbf{Method} & Early & Mid & Late && \multicolumn{2}{c}{Baseline} & $\rightarrow$ w/ Ours\\
      \midrule
          CDA & 37.72 & 36.74 & 32.14  && 2.77 & \multirow{6}{*}{\rotatebox{-90}{\textcolor{ForestGreen}{$\loongrightarrow$}}} & \textcolor{ForestGreen}{\textbf{1.51}} \\
          LTP & 32.48 & 28.16 & 28.16  && 2.59 && \textcolor{Magenta!50}{2.98}  \\
          BIA & 36.17 & 33.79 & 30.20  && 2.82 && \textcolor{ForestGreen}{\textbf{2.06}}  \\
          GAMA & 36.55 & 35.95 & 31.57 && 2.46 && \textcolor{ForestGreen}{\textbf{1.41}}  \\
          FACL & 36.48 & 34.38 & 31.76  && 2.19 && \textcolor{ForestGreen}{\textbf{1.17}} \\
          PDCL & 35.31 & 33.59 & 31.00  && 2.08 && \textcolor{ForestGreen}{\textbf{0.71}}  \\
      \bottomrule
        \end{tabular}
      }
      \vspace{-2mm} \caption*{(c)}
\end{minipage}
\vspace{-3mm}
\captionsetup{type=figure} \caption{Our observation on the semantic variability within the perturbation generator. (a) Generator intermediate feature maps for each block partition, (b) predicted masks from intermediate feature clusters on ImageNet-S~\cite{gao2022imagenet-S} from the baseline~\cite{zhang2022beyond}, and (c) quantified variability in foreground IoU.}
\label{fig:motivation}
\end{table*}

We observe that intermediate features progressively lose semantic recognizability across residual blocks. Figure~\ref{fig:motivation} shows that early maps preserve object contours, while mid and late maps blur them. Using k-means clustering to separate the foreground and background, we also find that stronger attacks preserve the coarse shape earlier and more consistently in stages. To better quantify how much semantic information is retained throughout the intermediates, we define semantic variability as the cross-block standard deviation of foreground IoU between clustered activation masks and foreground masks along perturbation trajectories, where advanced attacks achieve lower variability, suggesting more stable overlap with foreground. These findings are consistent with the well-established premise that the majority of noise being synthesized in the intermediate stage~\cite{zhang2022beyond, naseer2019cross, salzmann2021learning, zhang2022beyond, aich2022gama, yang2024facl, yang2024pdcl}. Based on this evidence, we apply a lightweight EMA teacher to early blocks, leaving inference unchanged, so that subsequent blocks concentrate perturbations on salient regions and black-box transfer improves. Further analysis is provided in \textit{Supp.}~\S D.

Crucially, these findings motivate our design to enforce semantic consistency in the intermediate stages of the generator, using an EMA teacher applied in the early blocks to curb semantic drift while leaving the inference path unchanged. By anchoring perturbations to early, semantically consistent features, the later blocks naturally concentrate the generated perturbations in salient object regions, thereby improving black-box transfer between models while preserving internal semantics.


\begin{figure*}[!t]
    \centering
    \includegraphics[width=\linewidth,trim={0 .3cm 0 0},clip]{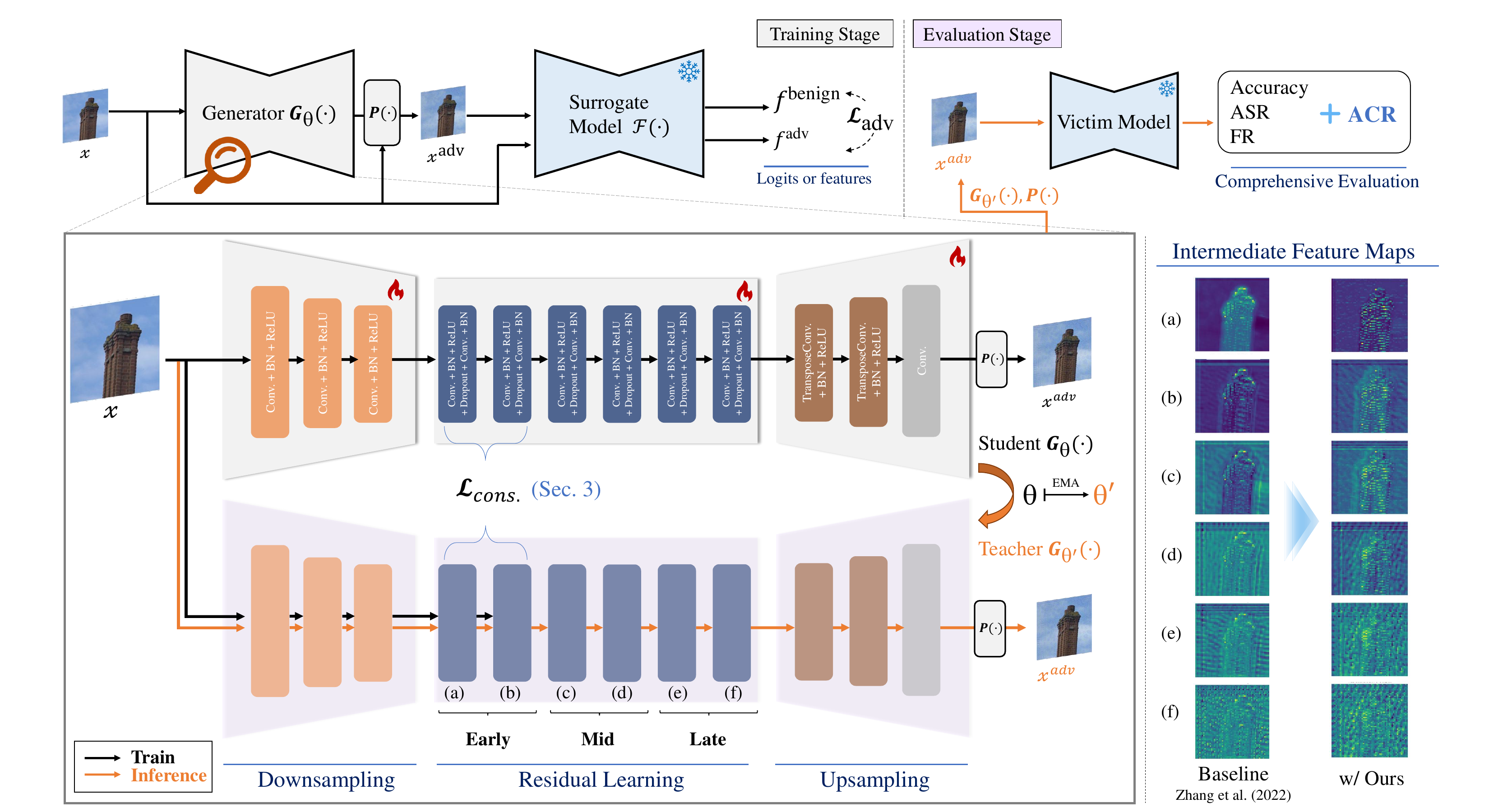}
    \vspace{-4mm}
    \caption{\textbf{Overview of our proposed SCGA framework.} Given a benign input image, a perturbation generator produces an adversarial output under the supervision of a Mean Teacher (\texttt{MT}) structure. The student and teacher share the generator architecture, with the teacher updated via EMA. Semantic consistency is enforced by aligning their intermediate features, selectively applied to the \textit{early} blocks to effectively preserve structural information from the benign input across the residual blocks. The adversarial example is then evaluated against victim models according to the four evaluation metrics. This \texttt{MT}-based design further promotes semantic alignment, combining consistency and integrity, thereby enhancing adversarial transferability across diverse victims.}
    \label{fig:overview}
\end{figure*}

\section{Semantically Consistent Generative Attack} \label{sec:semantic_structure_aware_generative_attack}
Our semantically consistent generative attack, as described in Alg.~\ref{alg:pseudocode_ours}, augments a standard generative adversarial attack with two key components: a Mean Teacher-based feature smoothing and a self‐feature consistency loss that enforces semantic preservation across the intermediate layers of the generator. We base our approach on the baseline work BIA~\cite{zhang2022beyond} as all subsequent works GAMA~\cite{aich2022gama}, FACL~\cite{yang2024facl}, PDCL~\cite{yang2024pdcl} base their losses on its feature similarity-based adversarial loss, and thus it is adequate to serve as a solid baseline. See \textit{Supp}.~\S C for the method distinction.

\paragraph{Role of Mean Teacher.}
The Mean Teacher (\texttt{MT}) framework~\cite{tarvainen2017meanteacher, deng2021unbiased, li2022cross, zhao2022enhanced, cao2023contrastive, dobler2023robust} has consistently demonstrated robustness in tasks characterized by significant domain shifts between training and testing. Its core mechanism of updating the teacher’s parameters with EMA of the student’s parameters provides a form of temporal ensemble that naturally suppresses instance-specific noise. Intuitively, this EMA update smooths out high-frequency perturbation artifacts, enriching the semantic consistency and stability of the teacher’s intermediate feature maps. As a result, these smoothed features serve as a reliable reference for the student, helping to preserve object contours and shapes throughout adversarial synthesis. To integrate \texttt{MT}, we maintain two generators: a \emph{student} $G_{\theta}(\cdot)$ that is trained via gradient descent, and a \emph{teacher} $G_{\theta'}(\cdot)$. We set these mean teacher features as a reference for our self-feature consistency matching. We update $\theta'$, per training step $t$, as follows: 
\begin{equation}
    \theta'_{t} \leftarrow \eta \,\theta'_{t-1} \;+\;(1-\eta)\,\theta_{t},
    \label{eq:ema_update}
\end{equation}
\noindent where $\eta\in[0,1]$ is a smoothing coefficient hyperparameter.

\begin{algorithm}[!t]
    \DontPrintSemicolon
    \caption{Pseudo-code of SCGA}
    \label{alg:pseudocode_ours}
    \KwData{Training dataset $\mathcal{D}_{src}$}
    \KwInput{Generator $\mathcal{G}_{\theta}(\cdot)$, a surrogate model trained on source data $\mathcal{F}^{s}(\cdot)$, projector $\mathcal{P}(\cdot)$, perturbation budget $\epsilon$}
    \KwOutput{Optimized teacher perturbation generator $\mathcal{G}_{\theta'}(\cdot)$}
    \smallskip
    Initialize generators: student $\mathcal{G}_{\theta}(\cdot) \gets$ random init., 
    teacher $\mathcal{G}_{\theta'}(\cdot) \gets \mathcal{G}_{\theta}(\cdot)$ \;
    \Repeat{$\mathcal{G}_{\theta}(\cdot)$ converges}{
        Randomly sample a mini-batch $x_{i}$ from $\mathcal{D}_{\textrm{src}}$\;
        Acquire student generator intermediate features: \quad\quad  $\mathbf{g}_{i} \gets \mathcal{G}_{\theta}^{\textrm{enc}}(x_{i})$\;
        Acquire teacher generator intermediate features: \quad\quad  $\mathbf{g}_{i}' \gets \mathcal{G}_{\theta'}^{\textrm{enc}}(x_{i})$\;
        Generate unbounded adversarial examples from student generator intermediate features: \quad\quad $\tilde{x}_{i}^{\textrm{adv}} \gets \mathcal{G}_{\theta}^{\textrm{dec}}(\mathbf{g}_{i})$\;
        Bound (project) $\tilde{x}_{i}^{\textrm{adv}}$ using $\mathcal{P}$  within the perturbation budget such that $||\mathcal{P}(\tilde{x}_{i}^{\textrm{adv}}) - x_{i}||_{\infty} \le \epsilon
        $ to obtain $x_{i}^{\textrm{adv}}$\;
        Forward pass $x_{i}$ and $x_{i}^{\textrm{adv}}$ through the surrogate model, $\mathcal{F}^{s}(\cdot)$ at layer $k$, to acquire $f_{i}^{\textrm{benign}}, f_{i}^{\textrm{adv}}$\;
        Compute loss using $f_{i}^{\textrm{benign}}, f_{i}^{\textrm{adv}}$, $\mathbf{g}_{i}, \mathbf{g}_{i}'$: \quad\quad $\mathcal{L} = \mathcal{L}_{\textrm{adv}} + \lambda_{\textrm{cons.}}\cdot\mathcal{L}_{\textrm{cons.}}$ \hfill\tcp{Eq.\ref{eq:loss}}
        Update student generator parameters via backpropagation\;
        EMA update teacher weights with student weights: \quad\quad  $\theta \mapsto \theta'$\hfill\tcp{Eq.~\ref{eq:ema_update}}
    }
    
\end{algorithm}

\paragraph{Self-feature consistency.}
Object-salient intermediate representations have been shown to be critical for adversarial transfer in black-box settings~\cite{wu2020boosting,byun2022improving,kim2022diverse, zhang2022beyond}, and recent work has explored manipulating input or surrogate-level features to this end~\cite{huang2019enhancing,li2023ilpd_attack,nakka2025nat}. In our generative framework, however, a na\"ive generator progressively loses semantic integrity in its intermediate layers (Fig.~\ref{fig:motivation}), scattering perturbations away from object-salient regions. To preserve these crucial object cues, we introduce a self-feature consistency mechanism grounded in the \texttt{MT} paradigm~\cite{grill2020bootstrap,caron2021emerging,lee2023exploring}.
Concretely, we treat the EMA teacher as the source of temporally smoothed, semantically rich features. At each training iteration, we extract early block activations from both the student and the teacher and enforce semantic consistency via a hinge‐based feature consistency loss as follows:
\begin{align}
\mathcal{L}_{\textrm{cons.}}
&= 
   \sum_{\ell=1}^{L_{\text{early}}}
    \mathcal{W}_{\textrm{cons.}} \cdot
   \Bigl[\,
     \tau
     - \frac{\langle \mathbf g^{\ell}_s,\mathbf g^{\ell}_t \rangle}
            {\|\mathbf g^{\ell}_s\|_2\,\|\mathbf g^{\ell}_t\|_2}
   \,\Bigr]_+,
\label{eq:consistency}
\end{align}

\noindent where $[\cdot]_+ \coloneqq \max(0,\cdot)$ and $\tau$ is the similarity threshold. This loss anchors the student’s edges and shape prior to the smoothed semantics of the teacher, ensuring that subsequent perturbations focus on object-centric regions. $\mathcal{W}_{\textrm{cons.}}\in\mathbb{R}^{|L|}$ denotes the softmax output of a learnable parameter for intermediate block-wise loss weighting. When combined with the adversarial objective, these semantically consistent perturbations that are highly transferable and tightly aligned with the core structure of the image.  
For fair comparisons with state-of-the-art methods, we adopt adversarial loss in the surrogate feature space as practiced in the baseline, \eg 
BIA
~\cite{zhang2022beyond}:
\begin{equation}
    \mathcal{L}_{\textrm{adv}}=\texttt{cos}(\mathcal{F}_{k}(x), \mathcal{F}_{k}(x^{adv})),
\end{equation}
\vspace{-1mm}
\noindent where \texttt{cos}($\cdot,\cdot$) denotes cosine similarity.

\paragraph{Final loss objective.}
Putting the proposed and baseline losses together on the \texttt{MT} framework, we formulate the final loss objective with $\lambda_{\textrm{cons.}}$ as a weight term for $\mathcal{L}_{\textrm{cons.}}$, as follows:
\begin{equation}
    \mathcal{L} = \mathcal{L}_{\textrm{adv}} + \lambda_{\textrm{cons.}} \cdot \mathcal{L}_{\textrm{cons.}}.
    \label{eq:loss}
\end{equation}

\section{Experiments}
\label{sec:experiments}

    We refer to \textit{Supp.} \S E.3 for training implementations and computational complexity. 
    For evaluation (\textit{Supp.} \S E.1--2), we conduct cross-setting tests under two black-box protocols. 
    In the cross-model setting, perturbations are crafted on surrogate models trained with the same data distribution (\ie, ImageNet-1K~\cite{imagenet}) and then tested on unseen target model architectures. 
    In the cross-domain/task settings, adversarial examples are to generalize across domain/task shifts without access to any target-distribution samples.
           
    
\subsection{Limitations in existing evaluation protocol} \label{sec:limitation_evalution}
    \begin{wraptable}{r}{0.6\textwidth}
        \vspace{-4mm}
    \centering
    \setlength{\tabcolsep}{5pt}
    \renewcommand{\arraystretch}{1}
    \renewcommand{\aboverulesep}{10pt}
    \renewcommand{\belowrulesep}{10pt}
    \caption{\textbf{Examples of real-world impacts on predictions}\\with different evaluation metrics and attack reliability concerns.}
    \label{tab:metric_impacts}
    \fontsize{32}{32}\selectfont
    \vspace{-3mm}
    \resizebox{\linewidth}{!}{
    \begin{tabular}{lcccccc}
    \toprule
    \rowcolor{Goldenrod!20} \multicolumn{7}{l}{\textbf{Real-world examples:}} \\\midrule
        Scenario \# &\textbf{GT Label }
        && \textbf{Benign pred.} 
        & \textbf{Adv. pred.} 
        & \textbf{Impact} 
        & \textbf{Captured by} \\
    \midrule
        1& \texttt{cat} && \texttt{cat} \textcolor{ForestGreen}{\cmark} & \texttt{cat}\textcolor{ForestGreen}{\cmark} & Correct $\rightarrow$ Correct & Acc. only \\
         2& \texttt{cat} && \texttt{cat} \textcolor{ForestGreen}{\cmark} & \texttt{dog} \textcolor{red}{\xmark} & Correct $\rightarrow$ Incorrect & ASR, FR \\
         3&\texttt{van} && \texttt{truck} \textcolor{red}{\xmark} & \texttt{bus} \textcolor{red}{\xmark} & Incorrect $\rightarrow$ Other incorrect & FR only \\
         4&\shortstack[c]{\texttt{pelagic}\\\texttt{cormorant}} && \texttt{albatross} \textcolor{red}{\xmark} & \shortstack[c]{\texttt{pelagic}\\\texttt{cormorant}} \textcolor{ForestGreen}{\cmark} & Incorrect $\rightarrow$ Correct & ACR, FR. Acc. \\
    \midrule\midrule
    \rowcolor{Goldenrod!20} \multicolumn{7}{l}{\textbf{Reliable attack example:}} \\\midrule
        \multicolumn{2}{c}{\textbf{Cross-Setting}}
        && \textbf{GT Label}
        & \textbf{Benign pred. }
        & \textbf{Intended Attack} 
        & \textbf{Unreliable Attack} \\
    \midrule
        \multicolumn{2}{c}{ImageNet\;$\rightarrow$\;FGVC Aircraft} && \shortstack[c]{\texttt{F-22}\\\texttt{Raptor}} & \shortstack[c]{\texttt{F-22}\\\texttt{Raptor}} \textcolor{red}{\xmark} & \shortstack[c]{\texttt{F-18}\\\texttt{Hornet}}
         \textcolor{red}{\xmark} & \shortstack[c]{\texttt{F-22}\\\texttt{Raptor}} \textcolor{ForestGreen}{\cmark} \\
    \bottomrule
    \end{tabular}}
        \vspace{-2mm}
    \end{wraptable}    
    Although developing an effective attack mechanism is crucial, it must be validated by fair and comprehensive evaluations. The current evaluation protocols adopted by previous works 
    GAP~\cite{poursaeed2018generative}, CDA~\cite{naseer2019cross}, LTP~\cite{salzmann2021learning}, BIA~\cite{zhang2022beyond}, GAMA~\cite{aich2022gama}, FACL~\cite{yang2024facl}, PDCL~\cite{yang2024pdcl}
    exhibit three key limitations. 
    (L1) Most studies report only one primary metric (either ASR, FR, or Acc.), 
    offering only a one-dimensional view of attack robustness and neglecting other aspects such as unintended corrections in predictions. 
    (L2) Data sets and sample sizes are often arbitrarily or limited to a single scale, preventing a fair comparison between attacks and undermining statistical significance. 
    (L3) Evaluations in previous work commonly target a narrow set of victim architectures (e.g., mostly CNN-based), lacking the diversity of modern model families, including vision transformers (ViT) and state-space models (SSM), and thus overstating robustness. 
    Although conventional work frames the success of attacks as \textit{fooling} the target classifier, we contend that evaluation facets should be expanded for a reliable assessment of attacks. 
    
    To address these shortcomings, we introduce, in \S \ref{sec:evaluation_metrics} (L1), \textit{Accidental Correction Rate} (ACR) as a complementary metric that captures the proportion of AEs that \textit{inadvertently} restore correct predictions, enriching the evaluation of attack efficacy alongside conventional measures (\ie ASR, FR, Acc.) as demonstrated with practical examples in Table~\ref{tab:metric_impacts}. ACR measures a nuanced model behavior often missed by ASR and FR, which is crucial for a complete understanding of robustness in safety-critical systems where any unreliable response to perturbation may pose a risk. 
    We also evaluate AEs on the \textit{entire} validation set in \S \ref{sec:exp_results} (L2, L3), instead of arbitrary subsets, and cover a \textit{wide range} of victim models for the classification task. We provide further details in \textit{Supp.} \S E. 

\subsection{Evaluation Metrics} \label{sec:evaluation_metrics}

    We tested the effectiveness and transferability of adversarial attacks across model architectures and domain shifts using four key metrics. For notational convenience here, let $f(x)$ denote the predicted label for input $x$, $f(x + \delta)$ the prediction after applying adversarial perturbation $\delta$, and $y$ the ground-truth label. The evaluation set is indicated by $\mathcal{D}$, with $\mathcal{C} = \{x \in \mathcal{D} \mid f(x) = y\}$ representing correctly classified samples, and $\mathcal{I} = \{x \in \mathcal{D} \mid f(x) \ne y\}$ denoting misclassified samples under clean inference. We formally define our evaluation metrics (\%) as follows:

\vspace{-6mm}
\begin{align}
    \text{Acc.}\;
    &= |\{x \in \mathcal{D} \;|\; f(x+\delta) = y \}|\;/\;|\mathcal{D}|,
    &\text{ASR}\;
    = |\{x \in \mathcal{C} \;|\; f(x) = y \wedge f(x+\delta) \ne y \}|\;/\;|\mathcal{C}|, \nonumber\\[-2pt]
    \text{FR}\;
    &= |\{x \in \mathcal{D} \;|\; f(x) \ne f(x+\delta)\}|\;/\;|\mathcal{D}|, 
    &\text{ACR}\;
    = |\{x \in \mathcal{I} \;|\; f(x) \ne y \wedge f(x+\delta) = y \}|\;/\;|\mathcal{I}|, 
\end{align}
\noindent where \emph{Top-1 Accuracy}~\cite{zhang2022beyond, yang2024facl, yang2024pdcl} measures the overall proportion of correctly classified samples under clean or adversarial conditions. It serves as a global performance indicator to assess degraded performance after the attack, orthogonal to FR, ASR, and ACR.
\emph{Attack Success Rate (ASR)}~\cite{poursaeed2018generative, naseer2019cross} is a subset of FR, which measures the proportion of samples originally correctly classified that are misclassified by adversarial attack. It directly reflects the targeted misclassification.
\emph{Fooling Rate (FR)}~\cite{salzmann2021learning, nakka2025nat} quantifies the proportion of adversarial examples that cause a change in the model’s prediction, regardless of correctness. It reflects how often the attack disrupts the original decision and is used as a transferability measure.
\emph{ACR}, also a subset of FR, is a novel metric that quantifies how often misclassified samples are “accidentally” corrected by adversarial perturbations. This unintended side effect provides insight into the nuanced model uncertainty and behavior at the decision boundaries.
For SS and OD, we use the standard mIoU and mAP50 metrics, respectively.

\subsection{Experimental Results} \label{sec:exp_results}

\begin{table*}[!t]
\centering 
\caption{\textbf{Quantitative cross-model transferability results}. We report the improvements ($\Delta$ \%p) of our method relative to each baseline, with better results marked in a darker color. `Avg.' corresponds to black-box average.}
\setlength{\tabcolsep}{5pt}
\renewcommand{\arraystretch}{1.1}
\renewcommand{\aboverulesep}{10pt}
\renewcommand{\belowrulesep}{10pt}
\label{tab:cross_model}
\vspace{-0.25cm}
{
\fontsize{38}{42}\selectfont
\resizebox{\linewidth}{!}{%
\begin{tabular}{ccccccccccccccccccccccccc}
\toprule
        \multicolumn{2}{c}{\texttt{Cross-model}}
        & \multicolumn{11}{c}{\bf CNN} 
        & \multicolumn{6}{c}{\bf Transformer} 
        & \multicolumn{2}{c}{\bf Mixer} 
        & \multicolumn{2}{c}{\bf Mamba} 
        & \multirow{2}{*}{Avg.}  \\ 
        \cmidrule[1pt](lr){3-13} 
        \cmidrule[1pt](lr){14-19} 
        \cmidrule[1pt](lr){20-21} 
        \cmidrule[1pt](lr){22-23}
\textbf{Method} 
        & \textbf{Metric}  
        & (a) & (b) & (c) & (d) & (e) & (f) & (g) & (h) & (i) & (j) & (k) 
        & (l) & (m) & (n) & (o) & (p) & (q) 
        & (r) & (s)
        & (t) & (v)
        & \\
\midrule
    \rowcolor{Goldenrod!20}
    Benign & Acc. (\%) $\downarrow$
        & 74.60 & 77.33 & 74.22 & 75.74 & 76.19 & 77.95 & 66.50 & 55.91 & 79.12 & 81.49 & 75.42
        & 80.67	& 79.28 &	81.19	& 80.48	& 79.10	& 57.91	
        & 69.90	& 66.53
        & 66.53 &	73.21
        & 73.77
        \\
\midrule

\multirow{4}{*}{\shortstack{CDA\\w/ Ours}}
     & Acc. ($\Delta$\%p) $\downarrow$
       & -15.93 & -8.39 & -12.93 & -12.70 & -8.41 & -11.21 & -5.09 & -6.48 & -10.17 & -35.91 & -19.74
       & -0.12 & -0.14 & \textcolor{gray!90}{+0.72} & \textcolor{gray!90}{+0.03} & \textcolor{gray!90}{+0.06} & \textcolor{gray!90}{+0.09}
       & \textcolor{gray!90}{+0.53} & \textcolor{gray!90}{+0.73}
       & \textcolor{gray!90}{+0.06} & \textcolor{gray!90}{+0.29}
       & \textcolor{ForestGreen!75}{\textbf{-6.89}} \\
    & ASR ($\Delta$\%p) $\uparrow$
       & +20.13 & +10.35 & +16.52 & +15.96 & +10.37 & +13.75 & +6.92 & +10.35 & +12.19 & +42.13 & +24.37
       & +0.09 & +0.10 & \textcolor{gray!90}{-0.95} & \textcolor{gray!90}{-0.05} & \textcolor{gray!90}{-0.05} & \textcolor{gray!90}{-0.23}
       & \textcolor{gray!90}{-0.67} & \textcolor{gray!90}{-1.04}
       & \textcolor{gray!90}{-0.13} & \textcolor{gray!90}{-0.51}
       & \textcolor{ForestGreen!75}{\textbf{+8.55}} \\
    & FR ($\Delta$\%p) $\uparrow$
       & +17.39 & +9.29 & +14.24 & +13.80 & +8.96 & +11.91 & +5.87 & +7.86 & +11.49 & +38.92 & +21.57
       & +0.09 & +0.03 & \textcolor{gray!90}{-0.99} & \textcolor{gray!90}{-0.05} & \textcolor{gray!90}{-0.15} & \textcolor{gray!90}{-0.37}
       & \textcolor{gray!90}{-0.74} & \textcolor{gray!90}{-0.94}
       & \textcolor{gray!90}{-0.20} & \textcolor{gray!90}{-0.60}
       & \textcolor{ForestGreen!75}{\textbf{+7.49}} \\ \cmidrule{2-24}
    & ACR ($\Delta$\%p) $\downarrow$
       & -3.58 & -1.72 & -2.59 & -2.52 & -2.19 & -2.25 & -1.45 & -1.58 & -2.52 & -7.04 & -4.57
       & -0.24 & -0.31 & -0.28 & -0.05 & \textcolor{gray!90}{+0.12} & -0.12
       & \textcolor{gray!90}{+0.20} & \textcolor{gray!90}{+0.08}
       & -0.11 & -0.29
       & \textcolor{ForestGreen!75}{\textbf{-1.57}} \\
\midrule

\multirow{4}{*}{\shortstack{LTP\\w/ Ours}}
     & Acc. ($\Delta$\%p) $\downarrow$
       & -8.71 & -9.52 & -8.45 & -10.24 & -4.62 & -10.00 & -5.59 & -9.60 & -9.57 & -5.57 & -5.93
       & -1.23 & -1.77 & -7.17 & -1.74 & -3.57 & -5.86
       & -5.87 & -9.05
       & -3.13 & -5.93
       & \textcolor{ForestGreen!75}{\textbf{-6.34}} \\
    & ASR ($\Delta$\%p) $\uparrow$
       & +11.11 & +11.92 & +10.87 & +12.92 & +5.90 & +12.27 & +8.04 & +15.83 & +11.89 & +6.55 & +7.50
       & +1.66 & +2.37 & +8.63 & +2.38 & +4.71 & +9.70
       & +8.22 & +13.03
       & +4.44 & +7.97
       & \textcolor{ForestGreen!75}{\textbf{+8.47}} \\
    & FR ($\Delta$\%p) $\uparrow$
       & +9.53 & +10.58 & +9.35 & +11.30 & +5.32 & +10.70 & +6.93 & +12.25 & +11.65 & +6.15 & +6.54
       & +2.19 & +2.82 & +8.71 & +2.79 & +5.10 & +8.95
       & +7.63 & +11.10
       & +3.87 & +7.27
       & \textcolor{ForestGreen!75}{\textbf{+7.65}} \\\cmidrule{2-24}
    & ACR ($\Delta$\%p) $\downarrow$
       & -1.65 & -1.31 & -1.48 & -1.85 & -0.53 & -2.01 & -0.72 & -1.71 & -0.76 & -1.01 & -0.77
       & \textcolor{gray!90}{+0.52} & \textcolor{gray!90}{+0.53} & -0.79 & \textcolor{gray!90}{+0.91} & \textcolor{gray!90}{+0.84} & -0.56
       & -0.44 & -1.04
       & -0.53 & -0.36
       & \textcolor{ForestGreen!75}{\textbf{-0.70}} \\
\midrule

\multirow{4}{*}{\shortstack{BIA\\w/ Ours}}
     & Acc. ($\Delta$\%p) $\downarrow$
       & -2.23 & -1.99 & -1.29 & \textcolor{gray!90}{+0.01} & -3.72 & -1.59 & -3.29 & -2.85 & -0.39 & \textcolor{gray!90}{+3.12} & -0.63
       & -0.80 & -0.32 & -0.78 & -0.56 & -1.18 & -1.08
       & -1.48 & -0.45
       & -0.38 & \textcolor{gray!90}{+0.03}
       & \textcolor{ForestGreen!75}{\textbf{-1.04}} \\
    & ASR ($\Delta$\%p) $\uparrow$
       & +2.83 & +2.46 & +1.64 & \textcolor{gray!90}{-0.05} & +4.55 & +1.96 & +4.68 & +4.80 & +0.56 & \textcolor{gray!90}{-3.32} & +1.21
       & +0.92 & +0.33 & +0.89 & +0.59 & +1.40 & +1.91
       & +2.05 & +0.88
       & +0.56 & +0.03
       & \textcolor{ForestGreen!75}{\textbf{+1.47}} \\
    & FR ($\Delta$\%p) $\uparrow$
       & +2.57 & +2.28 & +1.48 & \textcolor{gray!90}{-0.06} & +4.20 & +1.73 & +3.78 & +3.69 & +0.56 & \textcolor{gray!90}{-3.10} & +0.69
       & +1.01 & +0.44 & +1.00 & +0.57 & +1.44 & +1.62
       & +1.82 & +0.81
       & +0.35 & +0.04
       & \textcolor{ForestGreen!75}{\textbf{+1.28}} \\\cmidrule{2-24}
    & ACR ($\Delta$\%p) $\downarrow$
       & -0.45 & -0.36 & -0.28 & -0.09 & -1.09 & -0.27 & -0.53 & -0.38 & \textcolor{gray!90}{+0.24} & \textcolor{gray!90}{+0.33} & -0.37
       & -0.33 & -0.27 & -0.29 & -0.45 & -0.30 & \textcolor{gray!90}{+0.06}
       & -0.12 & \textcolor{gray!90}{+0.40}
       & -0.03 & \textcolor{gray!90}{+0.19}
       & \textcolor{ForestGreen!75}{\textbf{-0.21}} \\
\midrule

\multirow{4}{*}{\shortstack{GAMA\\w/ Ours}}
    & Acc. ($\Delta$\%p) $\downarrow$
      & -2.54 & -2.46 & -2.65 & -2.15 & -2.49 & -2.19 & -0.24 & -0.17 & -0.97 & -2.49 & -2.94
      & \textcolor{gray!90}{+0.07} & \textcolor{gray!90}{+0.03} & -0.24 & -0.01 & -0.15 & -0.51
      & \textcolor{gray!90}{+0.07} & -0.48
      & -0.59 & -0.41
      & \textcolor{ForestGreen!75}{\textbf{-1.12}} \\
    & ASR ($\Delta$\%p) $\uparrow$
      & +3.22 & +3.14 & +3.40 & +2.82 & +3.14 & +2.73 & +0.34 & +0.30 & +1.22 & +2.92 & +3.67
      & \textcolor{gray!90}{-0.04} & \textcolor{gray!90}{-0.05} & +0.30 & +0.13 & +0.12 & +0.89
      & \textcolor{gray!90}{-0.17} & +0.83
      & +0.75 & +0.52
      & \textcolor{ForestGreen!75}{\textbf{+1.44}} \\
    & FR ($\Delta$\%p) $\uparrow$
      & +2.81 & +2.87 & +2.91 & +2.56 & +2.76 & +2.53 & +0.24 & +0.21 & +1.20 & +2.67 & +3.23
      & +0.05 & +0.03 & +0.27 & +0.20 & +0.05 & +0.73
      & \textcolor{gray!90}{-0.21} & +0.73
      & +0.59 & +0.44
      & \textcolor{ForestGreen!75}{\textbf{+1.28}} \\\cmidrule{2-24}
    & ACR ($\Delta$\%p) $\downarrow$
      & -0.58 & -0.14 & -0.51 & -0.08 & -0.43 & -0.31 & -0.03 & \textcolor{gray!90}{0.00} & -0.02 & -0.49 & -0.56
      & \textcolor{gray!90}{+0.21} & -0.01 & -0.02 & \textcolor{gray!90}{+0.53} & -0.27 & \textcolor{gray!90}{+0.02}
      & -0.16 & \textcolor{gray!90}{+0.20}
      & -0.29 & -0.09
      & \textcolor{ForestGreen!75}{\textbf{-0.14}} \\
\midrule

\multirow{4}{*}{\shortstack{FACL\\w/ Ours}}
    & Acc. ($\Delta$\%p) $\downarrow$
      & \textcolor{gray!90}{+0.10} & -0.59 & -3.35 & -1.97 & -4.92 & -0.60 & -3.29 & -0.69 & -2.01 & -1.91 & -2.64
      & \textcolor{gray!90}{+0.11} & -0.33 & \textcolor{gray!90}{+0.21} & -0.51 & \textcolor{gray!90}{+0.56} & -0.18
      & -0.50 & \textcolor{gray!90}{+0.45}
      & -0.30 & -0.17
      & \textcolor{ForestGreen!75}{\textbf{-1.07}} \\ 
    & ASR ($\Delta$\%p) $\uparrow$
      & \textcolor{gray!90}{-0.20} & +0.74 & +4.30 & +2.46 & +6.15 & +0.75 & +4.68 & +1.25 & +2.40 & +2.23 & +3.15
      & \textcolor{gray!90}{-0.10} & +0.41 & \textcolor{gray!90}{-0.24} & +0.53 & \textcolor{gray!90}{-0.69} & +0.14
      & +0.68 & \textcolor{gray!90}{-0.72}
      & +0.34 & +0.15
      & \textcolor{ForestGreen!75}{\textbf{+1.35}} \\
    & FR ($\Delta$\%p) $\uparrow$
      & \textcolor{gray!90}{-0.20} & +0.64 & +3.75 & +2.27 & +5.37 & +0.74 & +3.97 & +0.96 & +2.24 & +2.05 & +2.78
      & \textcolor{gray!90}{-0.02} & +0.47 & \textcolor{gray!90}{-0.19} & +0.47 & \textcolor{gray!90}{-0.67} & +0.08
      & +0.72 & \textcolor{gray!90}{-0.64}
      & +0.25 & +0.14
      & \textcolor{ForestGreen!75}{\textbf{+1.20}} \\\cmidrule{2-24}
    & ACR ($\Delta$\%p) $\downarrow$
      & -0.23 & -0.09 & -0.61 & -0.46 & -0.97 & -0.08 & -0.54 & \textcolor{gray!90}{0.00} & -0.52 & -0.41 & -0.96
      & \textcolor{gray!90}{+0.16} & -0.02 & \textcolor{gray!90}{+0.05} & -0.46 & \textcolor{gray!90}{+0.09} & -0.23
      & -0.09 & -0.09
      & -0.24 & -0.24
      & \textcolor{ForestGreen!75}{\textbf{-0.28}} \\
\midrule

\multirow{4}{*}{\shortstack{PDCL\\w/ Ours}}
    & Acc. ($\Delta$\%p) $\downarrow$
      & \textcolor{gray!90}{+0.55} & -0.29 & \textcolor{gray!90}{+1.01} & -0.40 & -0.31 & -0.98 & -1.13 & -0.06 & -1.09 & -0.72 & \textcolor{gray!90}{+0.79}
      & -0.07 & \textcolor{gray!90}{+0.06} & -0.11 & \textcolor{gray!90}{+0.18} & -0.14 & \textcolor{gray!90}{+0.06}
      & \textcolor{gray!90}{+0.52} & \textcolor{gray!90}{+0.09}
      & \textcolor{gray!90}{+0.37} & \textcolor{gray!90}{+0.11}
      & \textcolor{ForestGreen!75}{\textbf{-0.07}} \\
    & ASR ($\Delta$\%p) $\uparrow$
      & \textcolor{gray!90}{-0.73} & +0.31 & \textcolor{gray!90}{-1.26} & +0.56 & +0.46 & +1.19 & +1.64 & +0.07 & +1.30 & +0.83 & \textcolor{gray!90}{-0.96}
      & +0.08 & \textcolor{gray!90}{-0.14} & +0.04 & \textcolor{gray!90}{-0.19} & +0.05 & \textcolor{gray!90}{-0.12}
      & \textcolor{gray!90}{-0.64} & \textcolor{gray!90}{-0.06}
      & \textcolor{gray!90}{-0.43} & \textcolor{gray!90}{-0.09}
      & \textcolor{ForestGreen!75}{\textbf{+0.09}}\\
    & FR ($\Delta$\%p) $\uparrow$
      & \textcolor{gray!90}{-0.68} & +0.27 & \textcolor{gray!90}{-1.08} & +0.36 & +0.45 & +1.09 & +1.42 & +0.13 & +1.23 & +0.81 & \textcolor{gray!90}{-0.88}
      & +0.22 & \textcolor{gray!90}{-0.10} & +0.09 & \textcolor{gray!90}{-0.03} & +0.09 & \textcolor{gray!90}{-0.22}
      & \textcolor{gray!90}{-0.44} & +0.05
      & \textcolor{gray!90}{-0.33} & \textcolor{gray!90}{-0.10}
      & \textcolor{ForestGreen!75}{\textbf{+0.11}}\\\cmidrule{2-24}
    & ACR ($\Delta$\%p) $\downarrow$
      & \textcolor{gray!90}{+0.03} & -0.18 & \textcolor{gray!90}{+0.29} & \textcolor{gray!90}{+0.09} & \textcolor{gray!90}{+0.18} & -0.22 & -0.12 & -0.05 & -0.33 & -0.22 & \textcolor{gray!90}{+0.21}
      & -0.06 & -0.23 & -0.42 & \textcolor{gray!90}{+0.08} & -0.47 & -0.04
      & \textcolor{gray!90}{+0.22} & \textcolor{gray!90}{+0.15}
      & \textcolor{gray!90}{+0.27} & \textcolor{gray!90}{+0.16}
      & \textcolor{ForestGreen!75}{\textbf{-0.03}}\\
\bottomrule
\end{tabular}%
}}
\end{table*}


\begin{table*}

\centering
\setlength{\tabcolsep}{1pt}
\renewcommand{\arraystretch}{1}
\renewcommand{\aboverulesep}{3pt}
\renewcommand{\belowrulesep}{3pt}
\caption{\textbf{Quantitative cross-domain/task transferability results}. We report the average improvement ($\Delta$ \%p) with ours added from each baseline for each domain. Better results in green \textbf{boldface}.}
\label{tab:cross_domain_task}
\vspace{-0.3cm}
\fontsize{24}{28}\selectfont
\resizebox{\linewidth}{!}{
\begin{tabular}{lcccccccccccccccc|ccccccc}
\toprule
    \multicolumn{17}{c}{\texttt{Cross-domain}} & \multicolumn{6}{c}{\texttt{Cross-task}}\\ \cmidrule{2-23}
    && \multicolumn{4}{c}{\textbf{CUB-200-2011}
    } 
    && \multicolumn{4}{c}{\textbf{Stanford Cars}
    } 
    && \multicolumn{4}{c}{\textbf{FGVC Aircraft}
    } 
    & \multirow{2}{*}{\shortstack{Avg.\\Acc.}} 
    & \multicolumn{2}{c}{\textbf{SemSeg (SS)}} 
    & \multirow{2}{*}{\shortstack{Avg.\\mIoU $\downarrow$}} 
    & \multicolumn{2}{c}{\textbf{ObjDet (OD)}} 
    & \multirow{2}{*}{\shortstack{Avg.\\mAP50 $\downarrow$}} 
    \\  \cmidrule{3-6}\cmidrule{8-11}\cmidrule{13-16} \cmidrule{18-19}  \cmidrule{21-22}
\textbf{Method \textbackslash\;Metric} 
    && Acc $\downarrow$ & ASR $\uparrow$ & FR $\uparrow$ & ACR $\downarrow$
    && Acc $\downarrow$ & ASR $\uparrow$ & FR $\uparrow$ & ACR $\downarrow$
    && Acc $\downarrow$ & ASR $\uparrow$ & FR $\uparrow$ & ACR $\downarrow$
    &
    & DeepLabV3+
    & SegFormer
    &
    & Faster R-CNN
    & DETR
    \\
\midrule
\rowcolor{Goldenrod!20} Benign  
    && 86.91 & \textcolor{gray}{\small N/A} & \textcolor{gray}{\small N/A} & \textcolor{gray}{\small N/A} 
    && 93.56 & \textcolor{gray}{\small N/A} & \textcolor{gray}{\small N/A} & \textcolor{gray}{\small N/A}
    && 92.07 & \textcolor{gray}{\small N/A} & \textcolor{gray}{\small N/A} & \textcolor{gray}{\small N/A} & 90.85 
    & 76.21 & 71.89 & 74.05
    & 61.01 & 62.36 & 61.69 \\
\midrule
CDA         
    && 67.73 & 21.48 & 14.16 & 26.66
    && 77.68 & 21.88 & 15.38 & 24.07
    && 64.42 & 27.51 & 14.55 & 31.13 & 69.94 & 25.63 & 20.16 & 22.90 & 32.78 & 26.29 & 29.54\\ 
w/ Ours ($\Delta$\%p)    
    && \textcolor{ForestGreen!75}{\textbf{-16.92}} & \textcolor{ForestGreen!75}{\textbf{+21.48}} & \textcolor{ForestGreen!75}{\textbf{+20.63}} & \textcolor{ForestGreen!75}{\textbf{-3.94}} 
    && \textcolor{ForestGreen!75}{\textbf{-5.86}} & \textcolor{ForestGreen!75}{\textbf{+2.38}} & \textcolor{ForestGreen!75}{\textbf{+2.35}} & \textcolor{ForestGreen!75}{\textbf{-0.24}} 
    && \textcolor{ForestGreen!75}{\textbf{-22.58}} & \textcolor{ForestGreen!75}{\textbf{+27.74}} & \textcolor{ForestGreen!75}{\textbf{+26.44}} & \textcolor{ForestGreen!75}{\textbf{-6.00}} & \textcolor{ForestGreen!75}{\textbf{-15.12}} 
    & \textcolor{ForestGreen!75}{\textbf{-0.47}} & +0.10 & \textcolor{ForestGreen!75}{\textbf{-0.18}} & \textcolor{ForestGreen!75}{\textbf{-0.80}} & \textcolor{ForestGreen!75}{\textbf{-0.61}} & \textcolor{ForestGreen!75}{\textbf{-0.71}} \\[2pt]
\midrule
LTP         
    && 48.74 & 45.32 & 8.75 & 49.31
    && 57.98 & 39.02 & 13.03 & 40.85
    && 43.01 & 54.15 & 8.35 & 56.48 & 49.91 & 23.71 & 26.97 & 25.34 & 29.39 & 22.41 & 25.90 \\  
w/ Ours ($\Delta$\%p)   
    && \textcolor{ForestGreen!75}{\textbf{-10.43}} & \textcolor{ForestGreen!75}{\textbf{+11.72}} & \textcolor{ForestGreen!75}{\textbf{+10.87}} & \textcolor{ForestGreen!75}{\textbf{-0.90}} 
    && \textcolor{ForestGreen!75}{\textbf{-10.62}} & \textcolor{ForestGreen!75}{\textbf{+10.98}} & \textcolor{ForestGreen!75}{\textbf{+10.68}} & \textcolor{ForestGreen!75}{\textbf{-2.16}} 
    && \textcolor{ForestGreen!75}{\textbf{-6.41}} & \textcolor{ForestGreen!75}{\textbf{+6.66}} & \textcolor{ForestGreen!75}{\textbf{+6.35}} & \textcolor{ForestGreen!75}{\textbf{-1.41}} & \textcolor{ForestGreen!75}{\textbf{-9.15}} 
    & \textcolor{ForestGreen!75}{\textbf{-1.44}} & \textcolor{ForestGreen!75}{\textbf{-0.29}} & \textcolor{ForestGreen!75}{\textbf{-0.86}} & \textcolor{ForestGreen!75}{\textbf{-2.54}} & \textcolor{ForestGreen!75}{\textbf{-0.23}} & \textcolor{ForestGreen!75}{\textbf{-1.39}} \\[2pt]
\midrule
BIA         
    && 47.92 & 46.13 & 8.54 & 50.26 
    && 59.89 & 37.22 & 13.96 & 38.97
    && 45.38 & 51.52 & 9.24 & 54.06 & 51.07 & 23.89 & 25.60 & 24.75 & 28.43 & 21.01 & 24.72\\  
w/ Ours ($\Delta$\%p) 
    && \textcolor{ForestGreen!75}{\textbf{-0.02}} & \textcolor{ForestGreen!75}{\textbf{+0.08}} & \textcolor{ForestGreen!75}{\textbf{+0.09}} & +0.49  
    && \textcolor{ForestGreen!75}{\textbf{-6.89}} & \textcolor{ForestGreen!75}{\textbf{+6.89}} & \textcolor{ForestGreen!75}{\textbf{+6.70}} & \textcolor{ForestGreen!75}{\textbf{-2.00}} 
    && \textcolor{ForestGreen!75}{\textbf{-4.98}} & \textcolor{ForestGreen!75}{\textbf{+5.25}} & \textcolor{ForestGreen!75}{\textbf{+4.89}} & \textcolor{ForestGreen!75}{\textbf{-1.24}} & \textcolor{ForestGreen!75}{\textbf{-3.96}} 
    & \textcolor{ForestGreen!75}{\textbf{-1.84}} & \textcolor{ForestGreen!75}{\textbf{-0.85}} & \textcolor{ForestGreen!75}{\textbf{-1.35}} & \textcolor{ForestGreen!75}{\textbf{-0.09}} & \textcolor{ForestGreen!75}{\textbf{-0.29}} & \textcolor{ForestGreen!75}{\textbf{-0.20}} \\[2pt]
\midrule
GAMA         
    && 48.72 & 45.41 & 9.51 & 49.67
    && 54.59 & 42.58 & 11.94 & 44.28
    && 42.37 & 54.46 & 7.49 & 56.77 & 48.56 & 23.67 & 25.95 & 24.81 & 28.01 & 20.71 & 24.36 \\  
w/ Ours ($\Delta$\%p)
    && \textcolor{ForestGreen!75}{\textbf{-2.41}} & \textcolor{ForestGreen!75}{\textbf{+2.59}} & \textcolor{ForestGreen!75}{\textbf{+2.30}} & \textcolor{ForestGreen!75}{\textbf{-0.67}}
    && \textcolor{ForestGreen!75}{\textbf{-2.33}} & \textcolor{ForestGreen!75}{\textbf{+2.24}} & \textcolor{ForestGreen!75}{\textbf{+2.10}} & \textcolor{ForestGreen!75}{\textbf{-0.60}} 
    && \textcolor{ForestGreen!75}{\textbf{-2.68}} & \textcolor{ForestGreen!75}{\textbf{+3.06}} & \textcolor{ForestGreen!75}{\textbf{+2.87}} & +0.14 & \textcolor{ForestGreen!75}{\textbf{-2.47}} 
    & \textcolor{ForestGreen!75}{\textbf{-0.43}} & \textcolor{ForestGreen!75}{\textbf{-1.58}} & \textcolor{ForestGreen!75}{\textbf{-1.01}} & \textcolor{ForestGreen!75}{\textbf{-0.41}} & +0.08 & \textcolor{ForestGreen!75}{\textbf{-0.16}} \\[2pt] 
\midrule
FACL         
    && 40.85 & 54.36 & 7.21 & 58.01
    && 51.23 & 48.23 & 12.9  & 49.71 
    && 40.08 & 59.35 & 7.34 & 61.39 & 44.05 & 23.75 & 26.40 & 25.08 & 27.94 & 20.91 & 24.43 \\  
w/ Ours ($\Delta$\%p)
    && +3.12 & -3.79 & -3.58 & +0.70
    && \textcolor{ForestGreen!75}{\textbf{-7.26}} & \textcolor{ForestGreen!75}{\textbf{+2.34}} & \textcolor{ForestGreen!75}{\textbf{+4.72}} & \textcolor{ForestGreen!75}{\textbf{-4.49}} 
    && \textcolor{ForestGreen!75}{\textbf{-2.68}} & \textcolor{ForestGreen!75}{\textbf{+0.60}} & \textcolor{ForestGreen!75}{\textbf{+0.66}} & \textcolor{ForestGreen!75}{\textbf{-0.24}} & \textcolor{ForestGreen!75}{\textbf{-2.27}} 
    & \textcolor{ForestGreen!75}{\textbf{-0.37}} & \textcolor{ForestGreen!75}{\textbf{-1.39}} & \textcolor{ForestGreen!75}{\textbf{-0.88}} & \textcolor{ForestGreen!75}{\textbf{-0.30}} & \textcolor{ForestGreen!75}{\textbf{-0.62}} & \textcolor{ForestGreen!75}{\textbf{-0.46}} \\[2pt]
\midrule
PDCL         
    && 42.36 & 52.32 & 7.48 & 55.93 
    && 50.41 & 46.85 & 12.31 & 48.46 
    && 38.96 & 58.23 & 6.86 & 60.34 & 43.91 & 24.42 & 26.05 & 25.24 & 28.48 & 21.38 & 24.93\\   
w/ Ours ($\Delta$\%p)
    && \textcolor{ForestGreen!75}{\textbf{-0.46}} & \textcolor{ForestGreen!75}{\textbf{+0.61}} & \textcolor{ForestGreen!75}{\textbf{+0.66}} & +0.40  
    && \textcolor{ForestGreen!75}{\textbf{-0.71}} & \textcolor{ForestGreen!75}{\textbf{+0.75}} & \textcolor{ForestGreen!75}{\textbf{+0.69}} & \textcolor{ForestGreen!75}{\textbf{-0.32}} 
    && \textcolor{ForestGreen!75}{\textbf{-1.38}} & \textcolor{ForestGreen!75}{\textbf{+1.52}} & \textcolor{ForestGreen!75}{\textbf{+1.42}} & +0.14 & \textcolor{ForestGreen!75}{\textbf{-0.85}} 
    & \textcolor{ForestGreen!75}{\textbf{-1.91}} & \textcolor{ForestGreen!75}{\textbf{-0.17}} & \textcolor{ForestGreen!75}{\textbf{-1.04}} & \textcolor{ForestGreen!75}{\textbf{-0.82}} & \textcolor{ForestGreen!75}{\textbf{-0.65}} & \textcolor{ForestGreen!75}{\textbf{-0.73}} \\
\bottomrule
\end{tabular}
\vspace{-8mm}
}

\end{table*}

We demonstrate enhanced cross-model attacks in Table~\ref{tab:cross_model}, wherein augmenting each baseline generative attack with our method yields consistent improvements across various architectures. Although these results confirm the orthogonality and efficacy of our framework, we observe that CLIP-based approaches with objectives similar to ours, \eg PDCL~\cite{yang2024pdcl}, yield only marginal improvements when combined with our method. 
We conjecture that optimizing for divergence in CLIP’s high-dimensional semantic embedding space may override or dilute the local structural consistency enforced by our early block semantic consistency, thus attenuating incremental gains from preserving fine-grained object contours and textures (see \textit{Supp}~\S D
for detailed explanation).

Table~\ref{tab:cross_domain_task} presents the black-box cross-domain transferability results. In both cross-domain and task, the transferability enhancements become more pronounced. Incorporating \texttt{MT} smoothing and early block consistency steadily enhances the attack performance across unseen domains, architectures, and tasks, demonstrating the broad applicability beyond the source data distribution and task.

\begin{figure*}[!t]
    \centering
    \includegraphics[width=.99\linewidth,trim={0 3cm 11cm 0},clip]{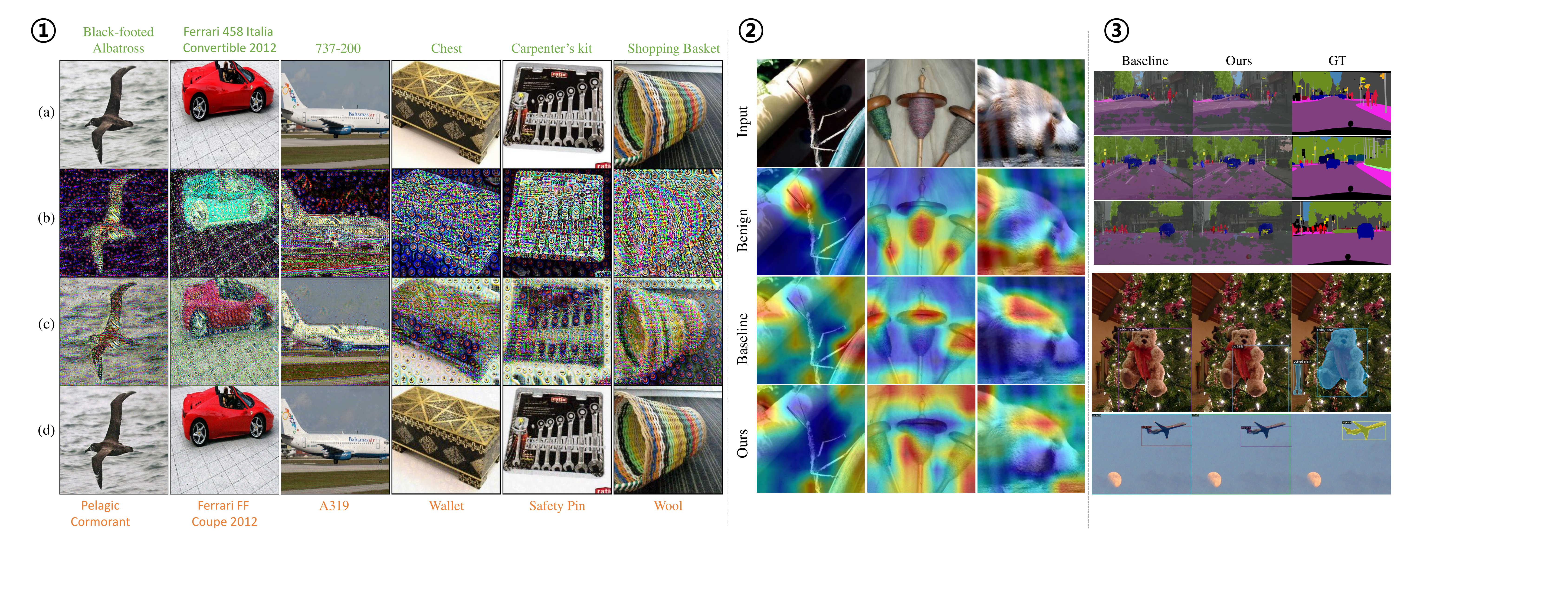}
    \vspace{-5mm}
    \caption{\textbf{Qualitative results.} Our semantically consistent generative attack successfully guides the generator to focus perturbations particularly on the semantically salient regions, effectively fooling the victim classifier. \textit{\textcircled{\raisebox{-0.9pt}{1}}:} (a) benign input image, (b) generated perturbation (normalized for visual purposes only), (c) unbounded adversarial image, and (d) bounded adversarial image across CUB-200-2011~\cite{cub}, Stanford Cars~\cite{car}, FGVC Aircraft~\cite{air}, and ImageNet-1K~\cite{imagenet}. 
    The label on top (\textcolor{ForestGreen}{green}) and bottom (\textcolor{orange}{orange}) denotes the correct label and prediction after the attack, respectively.
    \textit{\textcircled{\raisebox{-0.9pt}{2}}:} 
    We highlight that our method induces Grad-CAM~\cite{selvaraju2017gradcam} to focus on \textit{drastically different regions} in our adversarial examples compared to both the benign image and the adversarial examples crafted by the baseline~\cite{zhang2022beyond}. Moreover, our approach \textit{noticeably spreads and reduces the high activation regions} observed in the benign and baseline cases, enhancing the transferability of our adversarial perturbations.
    \textit{\textcircled{\raisebox{-0.9pt}{3}}:} Cross-task prediction results (SS on top, OD on bottom). Our approach further disrupts the victim models by triggering higher false positive rates and wrong class label predictions. See \textit{Supp.} \S E.4 for additional visualizations.
    }
    \vspace{-5mm}
    \label{fig:qualitative}
\end{figure*}

With measurable gains in attack accuracy, we visually verify whether our method actually induces the generator to pay more attention to the object-salient regions in Fig.~\ref{fig:qualitative}. Through Grad-CAM~\cite{selvaraju2017gradcam} comparisons against the baseline, ours either reinforces confusion or flips the correctly attending regions (similar to those of benign). Across unseen tasks, we also observe fewer pixels and instances with the correct classifications. 
We attribute this cross‐task generalization to our label‐agnostic training pipeline and further validate that our method can be integrated with alternative generator architectures beyond ResNet in \textit{Supp.}~\S B.


\begin{wraptable}{r}{0.5\textwidth}
\vspace{-4mm}
\centering
\caption{\textbf{Superior attack success with our method against robustly trained models} including Adversarially trained (AT) models and robust input pre-processing methods. Better averaged results in \textcolor{ForestGreen}{green} \textbf{boldface}.}
\label{tab:defense}
\setlength{\tabcolsep}{2pt}
\renewcommand{\arraystretch}{1}
\renewcommand{\aboverulesep}{5pt}
\renewcommand{\belowrulesep}{5pt}
\fontsize{24}{24}\selectfont
\vspace{-3mm}
\resizebox{\linewidth}{!}{
\begin{tabular}{ccccccccc}
    \toprule
        \textbf{Method} &  \textbf{Metric} & \textbf{Adv.IncV3} & \textbf{Adv.ViT} & \textbf{Adv.ConvNeXt} & \textbf{JPEG} & \textbf{BDR}& \textbf{R\&P} & \textbf{Avg.} \\
    \midrule
         \rowcolor{Goldenrod!20} Benign & Acc. (\%) $\downarrow$ & 76.33 & 48.82 & 58.44 & 74.68 & 74.68 & 76.58 &  68.26 \\
    \midrule
        \multirow{4}{*}{\makecell[c]{Baseline\\ \cite{zhang2022beyond}}} & 
        Acc. (\%)  $\downarrow$       & 68.54 & 45.64 & 53.88 & 63.49 & 47.82 & 44.78 & 54.03 \\
        
        & ASR (\%) $\uparrow$                & 14.95 & 11.72 & 10.26 & 20.24 & 40.76 & 44.59 &  23.75 \\
       
        & FR (\%) $\uparrow$  & 24.02 & 25.48 & 19.40 & 28.09 & 48.06 & 51.60 &  32.78 \\
    \cmidrule{2-9}
        & ACR (\%) $\downarrow$          & 15.30 &  4.96 &  3.46 & 11.45 & 11.30 & 10.56 &   9.51 \\
    \midrule
        \multirow{4}{*}{w/ Ours}
        & Acc. (\%) $\downarrow$         & 67.92 & 45.33 & 53.62 & 60.83 & 44.07 & 39.01 & \textcolor{ForestGreen}{\textbf{51.80}} \\
        
        & ASR (\%) $\uparrow$                & 15.75 & 11.95 & 10.65 & 23.74 & 45.37 & 51.63 & \textcolor{ForestGreen}{\textbf{26.52}} \\
        
        & FR (\%) $\uparrow$ & 24.83 & 25.31 & 19.60 & 31.61 & 52.22 & 57.86 & \textcolor{ForestGreen}{\textbf{35.28}} \\
    \cmidrule{2-9}
        & ACR (\%) $\downarrow$          & 15.23 &  4.57 &  3.38 & 11.48 & 10.29 &  9.08 &  \textcolor{ForestGreen}{\textbf{9.01}} \\
    \bottomrule
\end{tabular}
}

\vspace{-2mm}
\end{wraptable}
Against robust training (\ie adversarially trained 
IncV3~\cite{kurakin2018adversarial},ViT~\cite{dosovitskiy2020vit}, ConvNeXt~\cite{singh2023convstem}, and input pre-processing JPEG~\cite{JPEG} BDR~\cite{xu2017bitreduction}, R\&P~\cite{ xie2017randomization})
techniques, our methods demonstrate superior attacks compared to the baseline as shown in Table~\ref{tab:defense},
reinforcing our hypothesis that enforcing semantic consistency in early generator blocks not only boosts transferability in standard black-box settings but also produces perturbations capable of further enhancing attacks against defense mechanisms.
By anchoring structural cues in the early stages, our self-feature consistency loss
yields more potent and robust attacks against adversarially trained models and input pre-processing defenses alike.


\paragraph{
Interplay with the baselines.}
The pattern of gains across baselines in Table~\ref{tab:cross_domain_task} is largely determined by the level at which each method probes the surrogate (logits, frequency domain, or intermediate features). By enforcing early-block semantic anchoring, our generator produces locally structured, object-aware perturbations. These perturbations move energy away from degenerate high-frequency noise and toward low- and mid-frequency components that align with objects and boundaries. This structural regularization couples most strongly with CNN-centric objectives. When combined with CDA, whose relativistic loss is defined directly on CNN logits, and with frequency- or CNN-prior-based baselines such as FACL and PDCL, our semantics-enhanced perturbations yield the largest improvements on CNN victims. ViT victims, whose global attention patterns and feature geometry differ more from the CNN surrogate, tend to show smaller or more localized changes.

\begin{wraptable}{r}{0.58\textwidth}
    \vspace{-4mm}

\centering
\setlength{\tabcolsep}{1pt}
\renewcommand{\arraystretch}{1}
\renewcommand{\aboverulesep}{1pt}
\renewcommand{\belowrulesep}{1pt}
\caption{
\textbf{Ablation study on the targeted generator intermediate block and the proposed components.} Our self-feature consistency strategy on the early intermediate block outperforms matching other block features (a), and the generator trained with all of our components together performs best (b).}
\label{tab:ablation}
\vspace{-0.3cm}
\resizebox{\linewidth}{!}{
\begin{tabular}{lcc|ccccc|ccccccc}
\toprule
     & &
     &\multirow{4}{*}{\rotatebox{90}{\makecell[c]{\textbf{Block}}}} 
     & \multirow{4}{*}{\rotatebox{90}{Early}} 
     & \multirow{4}{*}{\rotatebox{90}{Mid}} 
     & \multirow{4}{*}{\rotatebox{90}{Late}} 
     & \multirow{4}{*}{\rotatebox{90}{All}} 
     & $\mathbf{\mathcal{L}_{adv}}$ & \cmark & \cmark & \cmark & \cmark & \cmark \\
     &&&&&&&& \texttt{PT} & {\textcolor{gray!20}{\footnotesize{\xmark}}} & {\textcolor{gray!20}{\footnotesize{\xmark}}} & \cmark & \cmark & {\textcolor{gray!20}{\footnotesize{\xmark}}} \\
    &&&&&&&& \texttt{MT} & \cmark & \cmark & {\textcolor{gray!20}{\footnotesize{\xmark}}} & {\textcolor{gray!20}{\footnotesize{\xmark}}} & {\textcolor{gray!20}{\footnotesize{\xmark}}}\\
    \multirow{-4}{*}{\rotatebox[origin=c]{90}{Cross-}} & Task & Metric 
    &&&&&& $\mathbf{\mathcal{L}_{\text{cons.}}}$ & \cmark & {\textcolor{gray!20}{\footnotesize{\xmark}}} & \cmark & {\textcolor{gray!20}{\footnotesize{\xmark}}}  & {\textcolor{gray!20}{\footnotesize{\xmark}}}\\

\midrule

    \multirow{4}{*}{\rotatebox[origin=c]{90}{Model}}   & \multirow{4}{*}{CLS}  & Acc. (\%) $\downarrow$  & \cellcolor{Goldenrod!20}  & \textbf{44.13} & 45.76 & 45.79 & 51.13 & \cellcolor{Goldenrod!20} & \textbf{44.13} & 48.23 & 45.11 & 46.49 & 45.17\\ 
    &                                               & ASR (\%) $\uparrow$ & \cellcolor{Goldenrod!20}  & \textbf{44.02} & 41.85 & 41.87 & 41.67 & \cellcolor{Goldenrod!20} & \textbf{44.02} & 33.37  & 42.80 & 41.02 & 42.55\\ 
    &                                                & FR (\%) $\uparrow$ & \cellcolor{Goldenrod!20}  & \textbf{50.66} & 48.71 & 48.75 & 48.57 & \cellcolor{Goldenrod!20} & \textbf{50.66} & 44.08  & 49.47 & 46.99 & 49.38\\ 
\cmidrule{3-3}\cmidrule{5-8}\cmidrule{10-14}
    &                                                & ACR (\%) $\downarrow$ & \cellcolor{Goldenrod!20}  & \textbf{8.32} & 8.59 & 8.66 & 8.60 & \cellcolor{Goldenrod!20} & \textbf{8.32} & 8.71 & 8.43 & 8.68 & 8.53\\ 
\cmidrule{2-3} \cmidrule{5-8} \cmidrule{10-14} 

    \multirow{4}{*}{\rotatebox[origin=c]{90}{Domain}}   & \multirow{4}{*}{CLS} & Acc. (\%) $\downarrow$   &\cellcolor{Goldenrod!20}& \textbf{47.10} & 50.95 & 49.03 & 51.13 &  \cellcolor{Goldenrod!20}  & \textbf{47.10} & 48.46  & 49.57 & 51.63  & 51.07 \\ 
    &                                               & ASR (\%) $\uparrow$  & \cellcolor{Goldenrod!20}& \textbf{49.02} & 44.91 & 47.02 & 44.72& \cellcolor{Goldenrod!20}  & \textbf{49.02} & 47.60  & 46.35 & 44.17 & 44.96 \\ 
    &                                               & FR  (\%) $\uparrow$       & \cellcolor{Goldenrod!20}& \textbf{51.66} & 47.67 & 49.75 & 47.50 &  \cellcolor{Goldenrod!20}  & \textbf{51.66} & 50.30  & 49.10 & 47.02 & 47.76 \\ 
\cmidrule{3-3}\cmidrule{5-8}\cmidrule{10-14}
    &                                               & ACR (\%) $\downarrow$   &  \cellcolor{Goldenrod!20}& \textbf{9.66} & 10.36 & 10.57 & 10.36 &  \cellcolor{Goldenrod!20}  & \textbf{9.66} & 9.99  & 9.89 & 10.73 & 10.58 \\ 
\cmidrule{2-3} \cmidrule{5-8} \cmidrule{10-14}

    \multirow{2}{*}{\rotatebox[origin=c]{90}{Task}} & SS  & mIoU $\downarrow$ &\cellcolor{Goldenrod!20}& 23.40 & 24.10 & \textbf{22.82} & 23.92 & \cellcolor{Goldenrod!20}\cellcolor{Goldenrod!20}                         & \textbf{23.40} & 23.96  & 24.83 & 23.73 & 24.75 \\
\cmidrule{2-3} \cmidrule{5-8} \cmidrule{10-14}

     & OD   & mAP50 $\downarrow$ & \cellcolor{Goldenrod!20}\multirow{-10}{*}{(a)}& \textbf{24.52} & 24.53 & 24.69 & \textbf{24.52} &\cellcolor{Goldenrod!20}\multirow{-10}{*}{(b)}      & 24.52 & 24.73  & 24.55 & \textbf{24.41} & 24.72 \\
    \bottomrule
\end{tabular}
}

    \vspace{-5mm}
\end{wraptable}
In contrast, mid-layer feature-based attacks such as LTP, BIA, and GAMA rely on intermediate surrogate features that transfer more readily across architectures. These methods benefit more uniformly. Our generator-side semantics act as a complementary regularizer that sharpens feature-space separability on both CNN and ViT targets, with broadly positive or neutral effects. On image classification, the additional gains when combining with PDCL are modest. 
This behavior is consistent with a saturation regime in which the strong CLIP-space objective already induces powerful global semantic shifts and dominates the joint gradient. Even in this setting, our anchor still rebalances the perturbation spectrum. For localization-oriented downstream tasks such as detection and segmentation, the same local structural consistency produces noticeably larger cross-task improvements. This behavior suggests that Ours refines the global CLIP-driven semantic direction rather than competing with it.


\paragraph{Ablation studies.}
We conducted ablation studies on the intermediate block and our proposed components in Table~\ref{tab:ablation}. Across all cross-settings, we observe the highest gains with self-feature consistency applied to the \textit{early} block compared to those at other and all locations, insinuating the early block matching triggers generator features to place stricter constraint such that perturbations are progressively focused on or around the object. 

We also observe performance gains with each component: $\mathcal{L}_{\textrm{adv}}$, \texttt{MT}, and $\mathcal{L}_{\textrm{cons.}}$, wherein our consistency of self-features on the intermediate features of the generator serves to widen the transferability gap even further. We attribute this improvement to explicit semantic alignment in the early blocks which complements the effect of implicit smoothing with \texttt{MT}. We also compare against the plain student-copy teacher, as indicated by plain teacher (\texttt{PT}), with and without $\mathcal{L}_{\textrm{cons.}}$, which under-performs our \texttt{MT} configuration. These results validate our hypothesis that anchoring perturbation synthesis on the early intermediate blocks consistently preserves the object semantics the most, and thus guides later blocks to concentrate noise on object-centric regions, maximizing transferability.



\subsection{Generator intermediate block-level analysis}

    \paragraph{Feature difference.}
    \begin{figure*}[!h]
        \centering
        \includegraphics[width=.99\linewidth]{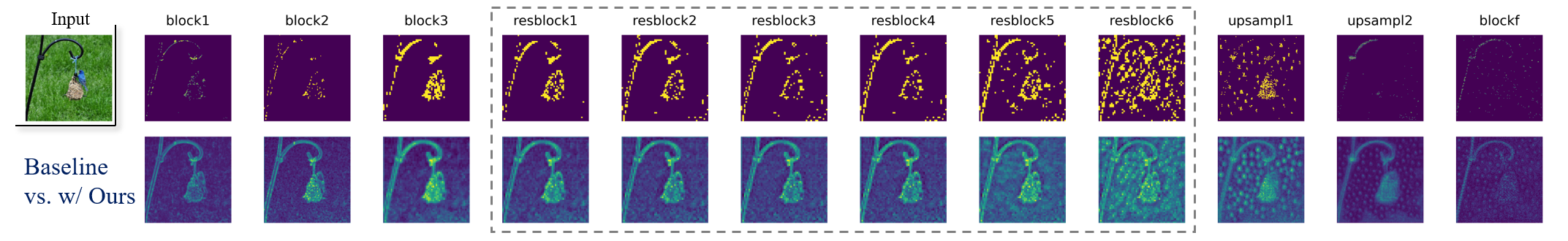}
        \vspace{-4mm}
        \caption{Visualization of generator intermediate block-level differences with the baseline~\cite{zhang2022beyond}: raw feature differences on bottom, and thresholded on top (normalized for illustration purposes only).  With our generator-internal semantic consistency mechanism, we progressively guide adversarial perturbation to focus on the salient object regions initially and gradually disperse to surrounding background regions. See \textit{Supp.} Fig.S10 for other baseline comparisons.}
        \label{fig:main_diffmap}
    \end{figure*}

    

    Following~\cite{zhang2022beyond,yang2024facl} but generalizing the procedure to all generator layers as follows, for each layer $l$:
    \vspace{-2mm}
    \begin{equation}
    \resizebox{\linewidth}{!}{$
        \mathtt{Diff}(\mathbf{g}^{l,pooled}_{\mathrm{baseline}}, \mathbf{g}^{l,pooled}_{\mathrm{ours}}) = 
    \nonumber
        \begin{cases}
            1, & \mathbf{g}^{l,pooled}_{\mathrm{ours}} - \mathbf{g}^{l,pooled}_{\mathrm{baseline}} > \tau_{\mathtt{diff}},\\
            0, & else,
        \end{cases}
        \qquad \text{with} \quad \mathbf{g}^{l,pooled}_{(\cdot)} = \frac{1}{C} \left| \displaystyle\sum_{C} G_{\theta_{(\cdot)}}^{l}(\mathbf{x}) \right|,
    $}
    \label{eq:main_diffmap}
    \vspace{-1mm}
    \end{equation}
    we visualized generator block-wise feature difference maps between each baseline with and without our method. 
    At each block, we computed the difference by applying cross-channel average pooling to the activation tensor and then thresholding the resulting map to qualitatively emphasize the added perturbations. As shown in Fig.~\ref{fig:main_diffmap}, the thresholded mask (row 1) and the raw feature difference map (row 2) jointly illustrate that, specifically within the targeted \texttt{resblocks} layers in our design, the adversarial signal concentrates on object-salient regions extracted by the preceding downsampling stages. Gradually into the later blocks, the generator learns to craft perturbation not only on object-salient regions but also regions closer to the background, generating more transferable noise. Compared to each baseline alone, our approach more strongly induces perturbations to better align with the semantic characteristics primarily in the intermediate residual blocks.
    
    \paragraph{Spectral energy comparisons.}

    \begin{wraptable}{r}{0.48\textwidth}
        \vspace{-6mm}
        \centering
        \caption{Spectral energy by band (Baseline$\rightarrow$w/ Ours).}
        \vspace{-3mm}
        \setlength{\tabcolsep}{5pt}
        \renewcommand{\arraystretch}{1}
        \renewcommand{\aboverulesep}{3pt}
        \renewcommand{\belowrulesep}{3pt}
        \fontsize{6}{6}\selectfont
        \footnotesize
        \resizebox{\linewidth}{!}{
        \begin{tabular}{lcccc}
        \toprule
        & \textbf{Band} & \textbf{Early}  & \textbf{Mid}  & \textbf{Late}\\
        \midrule
        \multirow{2}{*}{\makecell[l]{CDA\\$\rightarrow$w/ Ours}} & Low ($\uparrow$) & 0.82$\to$\textbf{0.91}  & 0.75$\to$\textbf{0.97}   & 0.77$\to$\textbf{0.96}  \\
        & High ($\downarrow$) & 0.18$\to$\textbf{0.09}  & 0.25$\to$\textbf{0.03}  & 0.23$\to$\textbf{0.04}  \\
        \midrule
        \multirow{2}{*}{\makecell[l]{LTP\\$\rightarrow$w/ Ours}} & Low ($\uparrow$)  & 0.73$\to$0.72  & 0.78$\to$\textbf{0.79}  & 0.95$\to$0.75   \\
        & High ($\downarrow$) & 0.27$\to$0.28  &0.22$\to$\textbf{0.21}  & 0.05$\to$0.25  \\
        \midrule
        \multirow{2}{*}{\makecell[l]{BIA\\$\rightarrow$w/ Ours}} & Low ($\uparrow$)  & 0.56$\to$0.56   & 0.53$\to$\textbf{0.54}  & 0.53$\to$\textbf{0.58}  \\
        & High ($\downarrow$) & 0.44$\to$0.44   & 0.47$\to$\textbf{0.45}  & 0.47$\to$\textbf{0.42}  \\
        \midrule
        \multirow{2}{*}{\makecell[l]{GAMA\\$\rightarrow$w/ Ours}} & Low ($\uparrow$)  & 0.57$\to$\textbf{0.79}  & 0.54$\to$\textbf{0.60}  & 0.56$\to$\textbf{0.59}  \\
        & High ($\downarrow$) & 0.43$\to$\textbf{0.21}  &0.46$\to$\textbf{0.40}  &0.44$\to$\textbf{0.41}  \\
        \midrule
        \multirow{2}{*}{\makecell[l]{FACL\\$\rightarrow$w/ Ours}} & Low ($\uparrow$)  & 0.57$\to$\textbf{0.73}  & 0.52$\to$\textbf{0.61}  & 0.54$\to$\textbf{0.59} \\
        & High ($\downarrow$) & 0.43$\to$\textbf{0.27}  &0.48$\to$\textbf{0.39}  &0.46$\to$\textbf{0.45}  \\
        \midrule
        \multirow{2}{*}{\makecell[l]{PDCL\\$\rightarrow$w/ Ours}} & Low ($\uparrow$)  & 0.54$\to$\textbf{0.62}  & 0.51$\to$\textbf{0.59}  & 0.58$\to$\textbf{0.59}  \\
        & High ($\downarrow$) & 0.46$\to$\textbf{0.38}  &0.49$\to$\textbf{0.41}  &0.42$\to$\textbf{0.41}  \\
        
        \bottomrule
        \end{tabular}}
        \label{fig:spectral_energy_ratio}
    \vspace{-4.5mm}
    \end{wraptable}
    To validate the early-block semantic anchoring hypothesis, we conducted a frequency-domain energy analysis of intermediate feature activations in Table~\ref{fig:spectral_energy_ratio}, exploiting the link between spectral content and visual structure: low-frequency (LF) components encode coarse shapes and layouts, whereas high frequencies (HF) capture fine texture. By tracking the normalized low-band energy in every block before and after our method, we obtained a quantitative measure of how strongly each block preserves the coarse structure. Anchoring on the early blocks, rather than mid or late, consistently raises low-frequency energy and suppresses superfluous high-frequency noise downstream, confirming that our method targeting semantic consistency in the early intermediates more effectively propagates the same semantic scaffold through later blocks, yielding higher adversarial transferability.
    
    The pattern reveals how anchoring affects generator's frequency bias. For band-wise relatively balanced models such as GAMA, the early-block anchor sharply increases low-frequency energy (0.57 → 0.79 ↑), giving later blocks a clearer structural blueprint. When a baseline already over-emphasizes low frequencies, as in LTP whose late-block LF reaches 0.95, our method lowers that value to 0.75 ↓, restoring HF detail. 
    This spectral analysis thus reveals that 
    anchoring on the early intermediate features results in perturbations that remain coarse semantic structures aligned and intact within the generator, thereby enhancing transfer effectively across unseen domains and architectures.

\paragraph{Hyperparameter sensitivity.}
We vary the EMA coefficient ($\eta$) and the consistency weight $\lambda_{\textrm{cons.}}$ and report the cross-setting transfer performance in Table~\ref{tab:hyperparameter}. We observe a trade-off between optimizing classification and cross-task scores for both hyperparameters, as no single combination uniformly outperforms the rest. However, maintaining relatively high values for both tends to yield better performance, indicating that each module sufficiently contributes to the overall self-consistency mechanism. Based on this observation, we select $\lambda_{\textrm{cons.}} = 0.7$ and $\eta = 0.999$ as our default configuration, which provides the best overall balance across all cross-setting scenarios.

We define the “early”, “mid”, and “late” stages of the generator intermediates by grouping two consecutive residual blocks based on the observation that perturbations undergo the most noticeable qualitative changes over every two blocks. As illustrated in Fig.~\ref{fig:overview}, the first two blocks (rows (a)–(b)) still closely track the benign image: coarse object shape, foreground–background separation, and large-scale texture are clearly preserved. The next two blocks (rows (c)–(d)) begin to introduce more pronounced distortions and fine-grained variations, while the final two blocks (rows (e)–(f)) predominantly add high-frequency details and noise-like patterns that are no longer easily interpretable as object-level structure. This makes the first two blocks a natural choice for enforcing semantic consistency: they are structurally well-formed and dominantly encode benign scene semantics, before most of the perturbation mass emerges in later stages.

\begin{wraptable}{r}{0.55\textwidth}
\vspace{-4.5mm}
\caption{\textbf{
Hyperparameter ($\lambda_{\text{cons}}$, $\eta$) sensitivity and early-block selection. (Domain (Acc.), Model (Acc.), SS (mIoU), OD (mAP50)).}}
\label{tab:hyperparameter}
\vspace{-2.5mm}
\centering
\centering
\setlength{\tabcolsep}{12pt}
\renewcommand{\arraystretch}{1}
\renewcommand{\aboverulesep}{1pt}
\renewcommand{\belowrulesep}{1pt}
\label{tab:sensitivity_lambda_cons}
\resizebox{\linewidth}{!}{%
\begin{tabular}{lccccc}
\toprule
$\lambda_{\text{cons}}$ & 0.1 & 0.3 & 0.5 & \cellcolor{Goldenrod!20} 0.7 & 0.9 \\
\midrule
Domain       & 49.55 & 48.29 & 48.49 & \cellcolor{Goldenrod!20}47.10 & 50.08 \\
Model         & 44.84 & 44.80 & 44.68 & \cellcolor{Goldenrod!20}44.13 & 45.89 \\
Task (SS)      & 22.59 & 23.63 & 23.82 & \cellcolor{Goldenrod!20}23.40 & 22.79 \\
Task (OD)     & 23.96 & 24.36 & 24.78 & \cellcolor{Goldenrod!20}24.52 & 24.19 \\
\bottomrule
\end{tabular}
}
\vspace{1pt}
\setlength{\tabcolsep}{15pt}
\renewcommand{\arraystretch}{1}
\renewcommand{\aboverulesep}{1pt}
\renewcommand{\belowrulesep}{1pt}
\label{tab:sensitivity_eta}
\resizebox{\linewidth}{!}{%
\begin{tabular}{lcccccc}
\toprule
$\eta$ & 0.9 & 0.95 & 0.98 & 0.99 & 0.995 & \cellcolor{Goldenrod!20}  0.999 \\
\midrule
Domain        & 48.72 & 50.47 & 51.17 & 47.97 & 48.09 & \cellcolor{Goldenrod!20}47.10 \\
Model          & 45.04 & 45.56 & 45.84 & 44.79 & 44.70 & \cellcolor{Goldenrod!20}44.13 \\
Task (SS)      & 24.89 & 23.92 & 24.23 & 23.39 & 24.34 & \cellcolor{Goldenrod!20}23.40 \\
Task (OD)     & 24.83 & 24.51 & 24.73 & 24.26 & 24.48 & \cellcolor{Goldenrod!20}24.52 \\
\bottomrule
\end{tabular}
}
\vspace{1pt}
\centering
\setlength{\tabcolsep}{20pt}
\renewcommand{\arraystretch}{1}
\renewcommand{\aboverulesep}{2pt}
\renewcommand{\belowrulesep}{2pt}
\label{tab:early_block_selection}
\resizebox{\linewidth}{!}{%
\begin{tabular}{lccc}
\toprule
\texttt{Block} & 1 & 2 & \cellcolor{Goldenrod!20}  1 \& 2 \\
\midrule
Domain        & 49.13 & 49.88 & \cellcolor{Goldenrod!20}47.10 \\
Model         & 47.62 & 48.82 & \cellcolor{Goldenrod!20}44.13 \\
Task (SS)      & 23.78 & 22.57 & \cellcolor{Goldenrod!20}23.40 \\
Task (OD)     & 24.47 & 24.15 & \cellcolor{Goldenrod!20}24.52 \\
\bottomrule
\end{tabular}
}   

\vspace{-3mm}
\end{wraptable}
Applying the temporal self-consistency loss only to block~1 or only to block~2 yields some benefits, but using both early blocks jointly provides a better balance across domains, models, and tasks. This pattern aligns with our intuition: anchoring both early blocks preserves coarse semantics at the onset of perturbation generation, which in turn biases later blocks to place perturbations along near-object regions rather than injecting unconstrained noise. As a consequence, the resulting perturbations align more closely with shared, object-level structure across architectures and datasets, thereby enhancing model- and data-agnostic black-box transfer. See Supp.\S E
for ablations of other components.


%

\section{Conclusion}
\label{sec:conclusion}

\begin{wrapfigure}{r}{0.5\textwidth}
    \vspace{-22mm}
    \centering
    \includegraphics[width=\linewidth,trim={0 1.1cm 0 0},clip]{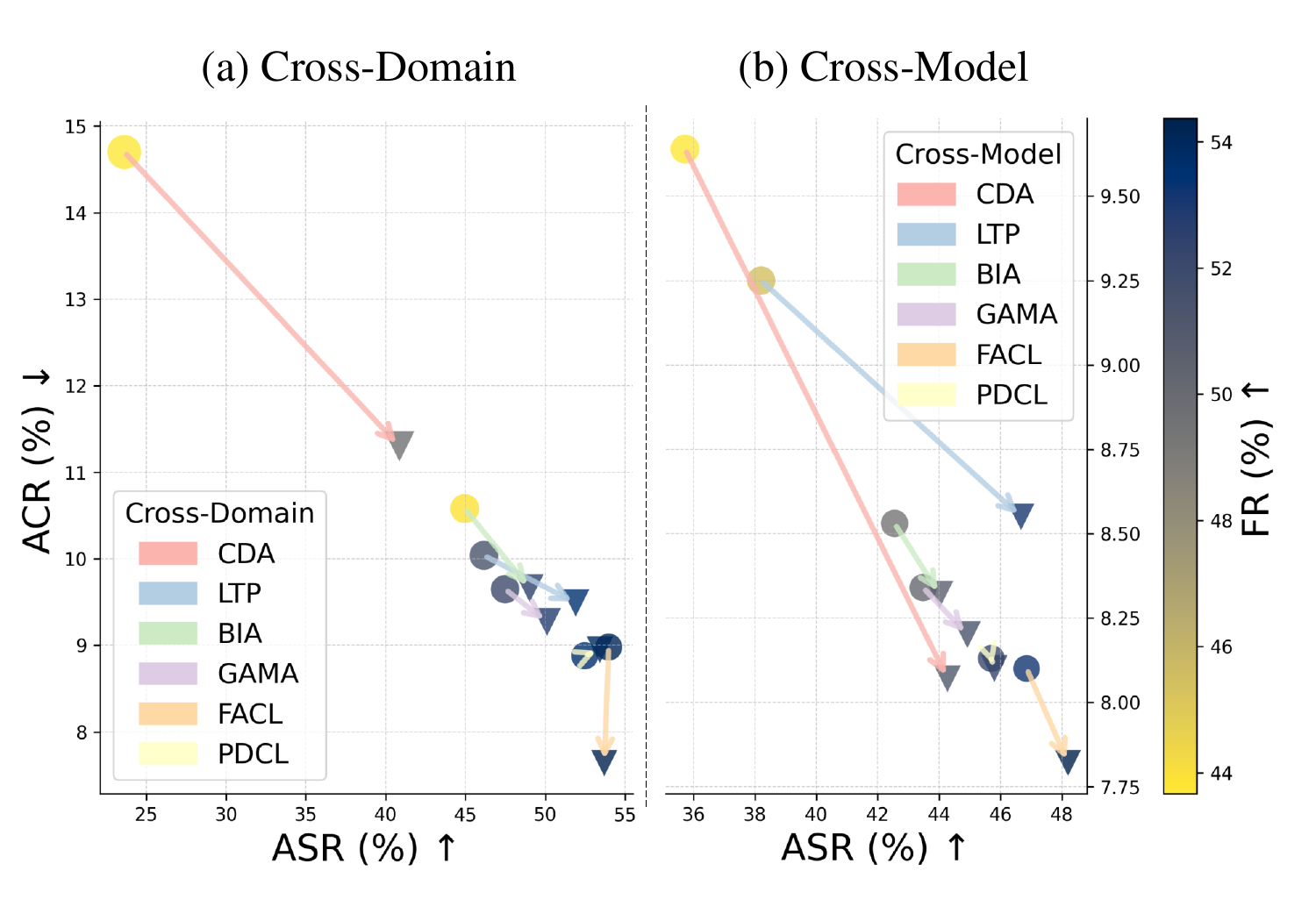}
    \vspace{-5.8mm}
    \caption{Our semantically consistent generative attack effectively exploits the generator intermediates to craft adversarial examples to enhance transferability from the baselines (${\tiny{\fullcirc}}\rightarrow\blacktriangledown$) across domains (a) and models (b).}
    \label{fig:teaser}
    \vspace{-5mm}
\end{wrapfigure}
\vspace{-5mm}
In this paper, we introduce a semantically consistent generative attack leveraging the Mean Teacher and early‐block semantic consistency to preserve the object semantics during perturbation generation, thus guiding it towards object‐salient regions to markedly improve black‐box transferability as in Fig.~\ref{fig:teaser}. 
With comprehensive evaluations across various models, domains, and tasks, we demonstrate object salient regions play a crucial role within the generator.
See \textit{Supp.} \S E.6 for limitations and broader societal impact.


\section*{Acknowledgments}
This work was supported by the Technology Innovation Program (2410013617,20024355, Development of autonomous driving connectivity technology based on sensor-infrastructure cooperation) funded By the Ministry of Trade, Industry \& Energy (MOTIE, Korea) and the Institute of Information \& communications Technology Planning \& Evaluation (IITP) grant funded by the Korea government (MSIT) (No. RS-2024-00457882, AI Research Hub Project).

{
    \bibliography{main}

\begin{thebibliography}{122}
\providecommand{\natexlab}[1]{#1}
\providecommand{\url}[1]{\texttt{#1}}
\expandafter\ifx\csname urlstyle\endcsname\relax
  \providecommand{\doi}[1]{doi: #1}\else
  \providecommand{\doi}{doi: \begingroup \urlstyle{rm}\Url}\fi

\bibitem[Aich et~al.(2022)Aich, Ta, Gupta, Song, Krishnamurthy, Asif, and Roy-Chowdhury]{aich2022gama}
Abhishek Aich, Calvin-Khang Ta, Akash Gupta, Chengyu Song, Srikanth Krishnamurthy, Salman Asif, and Amit Roy-Chowdhury.
\newblock Gama: Generative adversarial multi-object scene attacks.
\newblock \emph{Advances in Neural Information Processing Systems}, 35:\penalty0 36914--36930, 2022.

\bibitem[Baluja \& Fischer(2017)Baluja and Fischer]{baluja2017adversarial}
Shumeet Baluja and Ian Fischer.
\newblock Adversarial transformation networks: Learning to generate adversarial examples.
\newblock \emph{arXiv preprint arXiv:1703.09387}, 2017.

\bibitem[Baluja \& Fischer(2018)Baluja and Fischer]{baluja2018learning}
Shumeet Baluja and Ian Fischer.
\newblock Learning to attack: Adversarial transformation networks.
\newblock In \emph{Proceedings of the AAAI Conference on Artificial Intelligence}, volume~32, 2018.

\bibitem[Bao et~al.(2021)Bao, Dong, Piao, and Wei]{bao2021beit}
Hangbo Bao, Li~Dong, Songhao Piao, and Furu Wei.
\newblock Beit: Bert pre-training of image transformers.
\newblock \emph{arXiv preprint arXiv:2106.08254}, 2021.

\bibitem[Byun et~al.(2022)Byun, Cho, Kwon, Kim, and Kim]{byun2022improving}
Junyoung Byun, Seungju Cho, Myung-Joon Kwon, Hee-Seon Kim, and Changick Kim.
\newblock Improving the transferability of targeted adversarial examples through object-based diverse input.
\newblock In \emph{Proceedings of the IEEE/CVF Conference on Computer Vision and Pattern Recognition}, pp.\  15244--15253, 2022.

\bibitem[Cai et~al.(2022)Cai, Li, Hu, Gan, and Han]{cai2022efficientvit}
Han Cai, Junyan Li, Muyan Hu, Chuang Gan, and Song Han.
\newblock Efficientvit: Multi-scale linear attention for high-resolution dense prediction.
\newblock \emph{arXiv preprint arXiv:2205.14756}, 2022.

\bibitem[Cao et~al.(2023)Cao, Joshi, Gui, and Wang]{cao2023contrastive}
Shengcao Cao, Dhiraj Joshi, Liang-Yan Gui, and Yu-Xiong Wang.
\newblock Contrastive mean teacher for domain adaptive object detectors.
\newblock In \emph{Proceedings of the IEEE/CVF conference on computer vision and pattern recognition}, pp.\  23839--23848, 2023.

\bibitem[Carion et~al.(2020)Carion, Massa, Synnaeve, Usunier, Kirillov, and Zagoruyko]{carion2020detr}
Nicolas Carion, Francisco Massa, Gabriel Synnaeve, Nicolas Usunier, Alexander Kirillov, and Sergey Zagoruyko.
\newblock End-to-end object detection with transformers.
\newblock In \emph{European conference on computer vision}, pp.\  213--229. Springer, 2020.

\bibitem[Carlini et~al.(2019)Carlini, Athalye, Papernot, Brendel, Rauber, Tsipras, Goodfellow, Madry, and Kurakin]{carlini2019c_and_w}
Nicholas Carlini, Anish Athalye, Nicolas Papernot, Wieland Brendel, Jonas Rauber, Dimitris Tsipras, Ian Goodfellow, Aleksander Madry, and Alexey Kurakin.
\newblock On evaluating adversarial robustness.
\newblock \emph{arXiv preprint arXiv:1902.06705}, 2019.

\bibitem[Caron et~al.(2021)Caron, Touvron, Misra, J{\'e}gou, Mairal, Bojanowski, and Joulin]{caron2021emerging}
Mathilde Caron, Hugo Touvron, Ishan Misra, Herv{\'e} J{\'e}gou, Julien Mairal, Piotr Bojanowski, and Armand Joulin.
\newblock Emerging properties in self-supervised vision transformers.
\newblock In \emph{Proceedings of the IEEE/CVF international conference on computer vision}, pp.\  9650--9660, 2021.

\bibitem[Chen et~al.(2024)Chen, Chen, Chen, Zhang, Zou, and Shi]{chen2024diffusion}
Jianqi Chen, Hao Chen, Keyan Chen, Yilan Zhang, Zhengxia Zou, and Zhenwei Shi.
\newblock Diffusion models for imperceptible and transferable adversarial attack.
\newblock \emph{IEEE Transactions on Pattern Analysis and Machine Intelligence}, 2024.

\bibitem[Chen et~al.(2018)Chen, Zhu, Papandreou, Schroff, and Adam]{chen2018deeplabv3plus}
Liang-Chieh Chen, Yukun Zhu, George Papandreou, Florian Schroff, and Hartwig Adam.
\newblock Encoder-decoder with atrous separable convolution for semantic image segmentation.
\newblock In \emph{Proceedings of the European conference on computer vision (ECCV)}, pp.\  801--818, 2018.

\bibitem[Cordts et~al.(2016)Cordts, Omran, Ramos, Rehfeld, Enzweiler, Benenson, Franke, Roth, and Schiele]{cordts2016cityscapes}
Marius Cordts, Mohamed Omran, Sebastian Ramos, Timo Rehfeld, Markus Enzweiler, Rodrigo Benenson, Uwe Franke, Stefan Roth, and Bernt Schiele.
\newblock The cityscapes dataset for semantic urban scene understanding.
\newblock In \emph{Proceedings of the IEEE conference on computer vision and pattern recognition}, pp.\  3213--3223, 2016.

\bibitem[Dai et~al.(2024)Dai, Liang, and Xiao]{dai2024advdiff}
Xuelong Dai, Kaisheng Liang, and Bin Xiao.
\newblock Advdiff: Generating unrestricted adversarial examples using diffusion models.
\newblock In \emph{European Conference on Computer Vision}, pp.\  93--109. Springer, 2024.

\bibitem[D'Alfonso(2011)]{d2011quantifying}
Simon D'Alfonso.
\newblock On quantifying semantic information.
\newblock \emph{Information}, 2\penalty0 (1):\penalty0 61--101, 2011.

\bibitem[Deng et~al.(2021)Deng, Li, Chen, and Duan]{deng2021unbiased}
Jinhong Deng, Wen Li, Yuhua Chen, and Lixin Duan.
\newblock Unbiased mean teacher for cross-domain object detection.
\newblock In \emph{Proceedings of the IEEE/CVF conference on computer vision and pattern recognition}, pp.\  4091--4101, 2021.

\bibitem[D{\"o}bler et~al.(2023)D{\"o}bler, Marsden, and Yang]{dobler2023robust}
Mario D{\"o}bler, Robert~A Marsden, and Bin Yang.
\newblock Robust mean teacher for continual and gradual test-time adaptation.
\newblock In \emph{Proceedings of the IEEE/CVF Conference on Computer Vision and Pattern Recognition}, pp.\  7704--7714, 2023.

\bibitem[Dong et~al.(2018)Dong, Liao, Pang, Su, Zhu, Hu, and Li]{dong2018mifgsm}
Yinpeng Dong, Fangzhou Liao, Tianyu Pang, Hang Su, Jun Zhu, Xiaolin Hu, and Jianguo Li.
\newblock Boosting adversarial attacks with momentum.
\newblock In \emph{Proceedings of the IEEE conference on computer vision and pattern recognition}, pp.\  9185--9193, 2018.

\bibitem[Dong et~al.(2019)Dong, Pang, Su, and Zhu]{TI}
Yinpeng Dong, Tianyu Pang, Hang Su, and Jun Zhu.
\newblock Evading defenses to transferable adversarial examples by translation-invariant attacks.
\newblock In \emph{Proceedings of the IEEE/CVF Conference on Computer Vision and Pattern Recognition}, pp.\  4312--4321, 2019.

\bibitem[Dosovitskiy et~al.(2021)Dosovitskiy, Beyer, Kolesnikov, Weissenborn, Zhai, Unterthiner, Dehghani, Minderer, Heigold, Gelly, Uszkoreit, and Houlsby]{dosovitskiy2020vit}
Alexey Dosovitskiy, Lucas Beyer, Alexander Kolesnikov, Dirk Weissenborn, Xiaohua Zhai, Thomas Unterthiner, Mostafa Dehghani, Matthias Minderer, Georg Heigold, Sylvain Gelly, Jakob Uszkoreit, and Neil Houlsby.
\newblock An image is worth 16x16 words: Transformers for image recognition at scale.
\newblock \emph{ICLR}, 2021.

\bibitem[Fang et~al.(2024{\natexlab{a}})Fang, Kong, Chen, Dai, Wu, and Xia]{fang2024cgnc}
Hao Fang, Jiawei Kong, Bin Chen, Tao Dai, Hao Wu, and Shu-Tao Xia.
\newblock Clip-guided generative networks for transferable targeted adversarial attacks.
\newblock In \emph{European Conference on Computer Vision}, pp.\  1--19. Springer, 2024{\natexlab{a}}.

\bibitem[Fang et~al.(2024{\natexlab{b}})Fang, Kong, Chen, Dai, Wu, and Xia]{fang2024clip}
Hao Fang, Jiawei Kong, Bin Chen, Tao Dai, Hao Wu, and Shu-Tao Xia.
\newblock Clip-guided generative networks for transferable targeted adversarial attacks.
\newblock In \emph{European Conference on Computer Vision}, pp.\  1--19. Springer, 2024{\natexlab{b}}.

\bibitem[Floridi(2002)]{floridi2002philosophy}
Luciano Floridi.
\newblock What is the philosophy of information?
\newblock \emph{Metaphilosophy}, 33\penalty0 (1-2):\penalty0 123--145, 2002.

\bibitem[Furlanello et~al.(2018)Furlanello, Lipton, Tschannen, Itti, and Anandkumar]{furlanello2018bornagain}
Tommaso Furlanello, Zachary Lipton, Michael Tschannen, Laurent Itti, and Anima Anandkumar.
\newblock Born again neural networks.
\newblock In \emph{International conference on machine learning}, pp.\  1607--1616. PMLR, 2018.

\bibitem[Gao et~al.(2022)Gao, Li, Yang, Cheng, Han, and Torr]{gao2022imagenet-S}
Shanghua Gao, Zhong-Yu Li, Ming-Hsuan Yang, Ming-Ming Cheng, Junwei Han, and Philip Torr.
\newblock Large-scale unsupervised semantic segmentation.
\newblock 2022.

\bibitem[Girshick(2015)]{girshick2015fasterrcnn}
Ross Girshick.
\newblock Fast r-cnn.
\newblock In \emph{Proceedings of the IEEE international conference on computer vision}, pp.\  1440--1448, 2015.

\bibitem[Grill et~al.(2020)Grill, Strub, Altch{\'e}, Tallec, Richemond, Buchatskaya, Doersch, Avila~Pires, Guo, Gheshlaghi~Azar, et~al.]{grill2020bootstrap}
Jean-Bastien Grill, Florian Strub, Florent Altch{\'e}, Corentin Tallec, Pierre Richemond, Elena Buchatskaya, Carl Doersch, Bernardo Avila~Pires, Zhaohan Guo, Mohammad Gheshlaghi~Azar, et~al.
\newblock Bootstrap your own latent-a new approach to self-supervised learning.
\newblock \emph{Advances in neural information processing systems}, 33:\penalty0 21271--21284, 2020.

\bibitem[Guo et~al.(2017{\natexlab{a}})Guo, Rana, Cisse, and Van Der~Maaten]{JPEG}
Chuan Guo, Mayank Rana, Moustapha Cisse, and Laurens Van Der~Maaten.
\newblock Countering adversarial images using input transformations.
\newblock \emph{arXiv preprint arXiv:1711.00117}, 2017{\natexlab{a}}.

\bibitem[Guo et~al.(2017{\natexlab{b}})Guo, Rana, Cisse, and Van Der~Maaten]{guo2017countering}
Chuan Guo, Mayank Rana, Moustapha Cisse, and Laurens Van Der~Maaten.
\newblock Countering adversarial images using input transformations.
\newblock \emph{arXiv preprint arXiv:1711.00117}, 2017{\natexlab{b}}.

\bibitem[Hatamizadeh \& Kautz(2025)Hatamizadeh and Kautz]{hatamizadeh2025mambavision}
Ali Hatamizadeh and Jan Kautz.
\newblock Mambavision: A hybrid mamba-transformer vision backbone.
\newblock In \emph{Proceedings of the Computer Vision and Pattern Recognition Conference}, pp.\  25261--25270, 2025.

\bibitem[He et~al.(2016)He, Zhang, Ren, and Sun]{res152}
Kaiming He, Xiangyu Zhang, Shaoqing Ren, and Jian Sun.
\newblock Deep residual learning for image recognition.
\newblock In \emph{CVPR}, 2016.

\bibitem[Hu et~al.(2018)Hu, Shen, and Sun]{senet}
Jie Hu, Li~Shen, and Gang Sun.
\newblock Squeeze-and-excitation networks.
\newblock In \emph{Proceedings of the IEEE conference on computer vision and pattern recognition}, pp.\  7132--7141, 2018.

\bibitem[Huang \& Shen(2025)Huang and Shen]{huang2025huang}
Chihan Huang and Xiaobo Shen.
\newblock Huang: A robust diffusion model-based targeted adversarial attack against deep hashing retrieval.
\newblock In \emph{Proceedings of the AAAI Conference on Artificial Intelligence}, volume~39, pp.\  3626--3634, 2025.

\bibitem[Huang et~al.(2017)Huang, Liu, van~der Maaten, and Weinberger]{densenet}
Gao Huang, Zhuang Liu, Laurens van~der Maaten, and Kilian~Q. Weinberger.
\newblock Densely connected convolutional networks.
\newblock In \emph{CVPR}, 2017.

\bibitem[Huang et~al.(2019)Huang, Katsman, He, Gu, Belongie, and Lim]{huang2019enhancing}
Qian Huang, Isay Katsman, Horace He, Zeqi Gu, Serge Belongie, and Ser-Nam Lim.
\newblock Enhancing adversarial example transferability with an intermediate level attack.
\newblock In \emph{Proceedings of the IEEE/CVF international conference on computer vision}, pp.\  4733--4742, 2019.

\bibitem[Iandola et~al.(2016)Iandola, Han, Moskewicz, Ashraf, Dally, and Keutzer]{iandola2016squeezenet}
Forrest~N Iandola, Song Han, Matthew~W Moskewicz, Khalid Ashraf, William~J Dally, and Kurt Keutzer.
\newblock Squeezenet: Alexnet-level accuracy with 50x fewer parameters and< 0.5 mb model size.
\newblock \emph{arXiv preprint arXiv:1602.07360}, 2016.

\bibitem[Kim et~al.(2021)Kim, Ji, Yoon, and Hwang]{kim2021_pskd_selfkd}
Kyungyul Kim, ByeongMoon Ji, Doyoung Yoon, and Sangheum Hwang.
\newblock Self-knowledge distillation with progressive refinement of targets.
\newblock In \emph{Proceedings of the IEEE/CVF international conference on computer vision}, pp.\  6567--6576, 2021.

\bibitem[Kim et~al.(2022)Kim, Hong, and Yoon]{kim2022diverse}
Woo~Jae Kim, Seunghoon Hong, and Sung-Eui Yoon.
\newblock Diverse generative perturbations on attention space for transferable adversarial attacks.
\newblock In \emph{2022 IEEE international conference on image processing (ICIP)}, pp.\  281--285. IEEE, 2022.

\bibitem[Kingma \& Ba(2015)Kingma and Ba]{adam}
Diederik~P. Kingma and Jimmy Ba.
\newblock Adam: {A} method for stochastic optimization.
\newblock In \emph{ICLR}, 2015.

\bibitem[Kirillov et~al.(2023)Kirillov, Mintun, Ravi, Mao, Rolland, Gustafson, Xiao, Whitehead, Berg, Lo, et~al.]{kirillov2023sam}
Alexander Kirillov, Eric Mintun, Nikhila Ravi, Hanzi Mao, Chloe Rolland, Laura Gustafson, Tete Xiao, Spencer Whitehead, Alexander~C Berg, Wan-Yen Lo, et~al.
\newblock Segment anything.
\newblock In \emph{Proceedings of the IEEE/CVF international conference on computer vision}, pp.\  4015--4026, 2023.

\bibitem[Krause et~al.(2013)Krause, Stark, Deng, and Fei-Fei]{car}
Jonathan Krause, Michael Stark, Jia Deng, and Li~Fei-Fei.
\newblock 3d object representations for fine-grained categorization.
\newblock In \emph{2013 IEEE International Conference on Computer Vision Workshops}, pp.\  554--561, 2013.
\newblock \doi{10.1109/ICCVW.2013.77}.

\bibitem[Kurakin et~al.(2016)Kurakin, Goodfellow, and Bengio]{kurakin2016bim}
Alexey Kurakin, Ian Goodfellow, and Samy Bengio.
\newblock Adversarial machine learning at scale.
\newblock \emph{arXiv preprint arXiv:1611.01236}, 2016.

\bibitem[Kurakin et~al.(2018)Kurakin, Goodfellow, and Bengio]{kurakin2018adversarial}
Alexey Kurakin, Ian~J Goodfellow, and Samy Bengio.
\newblock Adversarial examples in the physical world.
\newblock In \emph{Artificial intelligence safety and security}, pp.\  99--112. Chapman and Hall/CRC, 2018.

\bibitem[Lee et~al.(2023)Lee, Willette, Kim, Lee, and Hwang]{lee2023exploring}
Youngwan Lee, Jeffrey~Ryan Willette, Jonghee Kim, Juho Lee, and Sung~Ju Hwang.
\newblock Exploring the role of mean teachers in self-supervised masked auto-encoders.
\newblock In \emph{The Eleventh International Conference on Learning Representations}, 2023.
\newblock URL \url{https://openreview.net/forum?id=7sn6Vxp92xV}.

\bibitem[Lei et~al.(2025)Lei, Guo, Lee, Duong, and Lau]{lei2025towards}
Chun~Tong Lei, Zhongliang Guo, Hon~Chung Lee, Minh~Quoc Duong, and Chun~Pong Lau.
\newblock Towards more transferable adversarial attack in black-box manner.
\newblock \emph{arXiv preprint arXiv:2505.18097}, 2025.

\bibitem[Li et~al.(2020{\natexlab{a}})Li, Deng, Li, Yan, Gao, and Huang]{li2020towards}
Maosen Li, Cheng Deng, Tengjiao Li, Junchi Yan, Xinbo Gao, and Heng Huang.
\newblock Towards transferable targeted attack.
\newblock In \emph{Proceedings of the IEEE/CVF conference on computer vision and pattern recognition}, pp.\  641--649, 2020{\natexlab{a}}.

\bibitem[Li et~al.(2020{\natexlab{b}})Li, Guo, and Chen]{li2020ilaplusplus}
Qizhang Li, Yiwen Guo, and Hao Chen.
\newblock Yet another intermediate-level attack.
\newblock In \emph{European Conference on Computer Vision}, pp.\  241--257. Springer, 2020{\natexlab{b}}.

\bibitem[Li et~al.(2023)Li, Guo, Zuo, and Chen]{li2023ilpd_attack}
Qizhang Li, Yiwen Guo, Wangmeng Zuo, and Hao Chen.
\newblock Improving adversarial transferability via intermediate-level perturbation decay.
\newblock \emph{Advances in Neural Information Processing Systems}, 36:\penalty0 32900--32912, 2023.

\bibitem[Li et~al.(2022)Li, Dai, Ma, Liu, Chen, Wu, He, Kitani, and Vajda]{li2022cross}
Yu-Jhe Li, Xiaoliang Dai, Chih-Yao Ma, Yen-Cheng Liu, Kan Chen, Bichen Wu, Zijian He, Kris Kitani, and Peter Vajda.
\newblock Cross-domain adaptive teacher for object detection.
\newblock In \emph{Proceedings of the IEEE/CVF conference on computer vision and pattern recognition}, pp.\  7581--7590, 2022.

\bibitem[Li et~al.(2024)Li, Li, Yang, Song, Yang, and Pan]{li2024dual_selfkd}
Zheng Li, Xiang Li, Lingfeng Yang, Renjie Song, Jian Yang, and Zhigeng Pan.
\newblock Dual teachers for self-knowledge distillation.
\newblock \emph{Pattern Recognition}, 151:\penalty0 110422, 2024.

\bibitem[Liang et~al.(2023)Liang, Wu, Dai, Li, Zhao, Zhang, Zhang, Vajda, and Marculescu]{liang2023open}
Feng Liang, Bichen Wu, Xiaoliang Dai, Kunpeng Li, Yinan Zhao, Hang Zhang, Peizhao Zhang, Peter Vajda, and Diana Marculescu.
\newblock Open-vocabulary semantic segmentation with mask-adapted clip.
\newblock In \emph{Proceedings of the IEEE/CVF conference on computer vision and pattern recognition}, pp.\  7061--7070, 2023.

\bibitem[Lin et~al.(2019)Lin, Song, He, Wang, and Hopcroft]{SI}
Jiadong Lin, Chuanbiao Song, Kun He, Liwei Wang, and John~E Hopcroft.
\newblock Nesterov accelerated gradient and scale invariance for adversarial attacks.
\newblock \emph{arXiv preprint arXiv:1908.06281}, 2019.

\bibitem[Lin et~al.(2014)Lin, Maire, Belongie, Hays, Perona, Ramanan, Doll{\'a}r, and Zitnick]{lin2014mscoco}
Tsung-Yi Lin, Michael Maire, Serge Belongie, James Hays, Pietro Perona, Deva Ramanan, Piotr Doll{\'a}r, and C~Lawrence Zitnick.
\newblock Microsoft coco: Common objects in context.
\newblock In \emph{Computer vision--ECCV 2014: 13th European conference, zurich, Switzerland, September 6-12, 2014, proceedings, part v 13}, pp.\  740--755. Springer, 2014.

\bibitem[Liu et~al.(2023{\natexlab{a}})Liu, Li, Wu, and Lee]{liu2023llava}
Haotian Liu, Chunyuan Li, Qingyang Wu, and Yong~Jae Lee.
\newblock Visual instruction tuning, 2023{\natexlab{a}}.

\bibitem[Liu et~al.(2023{\natexlab{b}})Liu, Zhu, Liang, Zhang, Fang, Zhang, and Chang]{liu2023improving}
Jiayang Liu, Siyu Zhu, Siyuan Liang, Jie Zhang, Han Fang, Weiming Zhang, and Ee-Chien Chang.
\newblock Improving adversarial transferability by stable diffusion.
\newblock \emph{arXiv preprint arXiv:2311.11017}, 2023{\natexlab{b}}.

\bibitem[Liu et~al.(2024)Liu, Zhou, Zhang, Chen, Zhao, and Lam]{liu2024boosting}
Renyang Liu, Wei Zhou, Tianwei Zhang, Kangjie Chen, Jun Zhao, and Kwok-Yan Lam.
\newblock Boosting black-box attack to deep neural networks with conditional diffusion models.
\newblock \emph{IEEE Transactions on Information Forensics and Security}, 19:\penalty0 5207--5219, 2024.

\bibitem[Liu et~al.(2021)Liu, Lin, Cao, Hu, Wei, Zhang, Lin, and Guo]{liu2021swin}
Ze~Liu, Yutong Lin, Yue Cao, Han Hu, Yixuan Wei, Zheng Zhang, Stephen Lin, and Baining Guo.
\newblock Swin transformer: Hierarchical vision transformer using shifted windows.
\newblock In \emph{Proceedings of the IEEE/CVF international conference on computer vision}, pp.\  10012--10022, 2021.

\bibitem[Liu et~al.(2022)Liu, Mao, Wu, Feichtenhofer, Darrell, and Xie]{liu2022convnext}
Zhuang Liu, Hanzi Mao, Chao-Yuan Wu, Christoph Feichtenhofer, Trevor Darrell, and Saining Xie.
\newblock A convnet for the 2020s.
\newblock In \emph{Proceedings of the IEEE/CVF conference on computer vision and pattern recognition}, pp.\  11976--11986, 2022.

\bibitem[Ma et~al.(2024)Ma, Li, Xiao, Cao, Zhang, Ye, and Zhao]{ma2024jailbreaking}
Jiachen Ma, Yijiang Li, Zhiqing Xiao, Anda Cao, Jie Zhang, Chao Ye, and Junbo Zhao.
\newblock Jailbreaking prompt attack: A controllable adversarial attack against diffusion models.
\newblock \emph{arXiv preprint arXiv:2404.02928}, 2024.

\bibitem[Maaten \& Hinton(2008)Maaten and Hinton]{maaten2008visualizingtsne}
Laurens van~der Maaten and Geoffrey Hinton.
\newblock Visualizing data using t-sne.
\newblock \emph{Journal of machine learning research}, 9\penalty0 (Nov):\penalty0 2579--2605, 2008.

\bibitem[Madry et~al.(2017)Madry, Makelov, Schmidt, Tsipras, and Vladu]{madry2017pgd}
Aleksander Madry, Aleksandar Makelov, Ludwig Schmidt, Dimitris Tsipras, and Adrian Vladu.
\newblock Towards deep learning models resistant to adversarial attacks.
\newblock \emph{arXiv preprint arXiv:1706.06083}, 2017.

\bibitem[Maji et~al.(2013)Maji, Rahtu, Kannala, Blaschko, and Vedaldi]{air}
Subhransu Maji, Esa Rahtu, Juho Kannala, Matthew~B. Blaschko, and Andrea Vedaldi.
\newblock Fine-grained visual classification of aircraft.
\newblock \emph{ArXiv}, abs/1306.5151, 2013.
\newblock URL \url{https://api.semanticscholar.org/CorpusID:2118703}.

\bibitem[Marcel \& Rodriguez(2010)Marcel and Rodriguez]{marcel2010torchvision}
S{\'e}bastien Marcel and Yann Rodriguez.
\newblock Torchvision the machine-vision package of torch.
\newblock In \emph{Proceedings of the 18th ACM international conference on Multimedia}, pp.\  1485--1488, 2010.

\bibitem[Minderer et~al.(2022)Minderer, Gritsenko, Stone, Neumann, Weissenborn, Dosovitskiy, Mahendran, Arnab, Dehghani, Shen, et~al.]{minderer2022simple}
Matthias Minderer, Alexey Gritsenko, Austin Stone, Maxim Neumann, Dirk Weissenborn, Alexey Dosovitskiy, Aravindh Mahendran, Anurag Arnab, Mostafa Dehghani, Zhuoran Shen, et~al.
\newblock Simple open-vocabulary object detection.
\newblock In \emph{European conference on computer vision}, pp.\  728--755. Springer, 2022.

\bibitem[Moosavi-Dezfooli et~al.(2017)Moosavi-Dezfooli, Fawzi, Fawzi, and Frossard]{moosavi2017universal}
Seyed-Mohsen Moosavi-Dezfooli, Alhussein Fawzi, Omar Fawzi, and Pascal Frossard.
\newblock Universal adversarial perturbations.
\newblock In \emph{Proceedings of the IEEE conference on computer vision and pattern recognition}, pp.\  1765--1773, 2017.

\bibitem[Nakka \& Salzmann(2021)Nakka and Salzmann]{salzmann2021learning}
Krishna Nakka and Mathieu Salzmann.
\newblock Learning transferable adversarial perturbations.
\newblock In A.~Beygelzimer, Y.~Dauphin, P.~Liang, and J.~Wortman Vaughan (eds.), \emph{Advances in Neural Information Processing Systems}, 2021.
\newblock URL \url{https://openreview.net/forum?id=sIDvIyR5I1R}.

\bibitem[Nakka \& Alahi(2025)Nakka and Alahi]{nakka2025nat}
Krishna~Kanth Nakka and Alexandre Alahi.
\newblock Nat: Learning to attack neurons for enhanced adversarial transferability.
\newblock In \emph{2025 IEEE/CVF Winter Conference on Applications of Computer Vision (WACV)}, pp.\  7593--7604. IEEE, 2025.

\bibitem[Naseer et~al.(2019)Naseer, Khan, Khan, Shahbaz~Khan, and Porikli]{naseer2019cross}
Muhammad~Muzammal Naseer, Salman~H Khan, Muhammad~Haris Khan, Fahad Shahbaz~Khan, and Fatih Porikli.
\newblock Cross-domain transferability of adversarial perturbations.
\newblock \emph{Advances in Neural Information Processing Systems}, 32, 2019.

\bibitem[Naseer et~al.(2020)Naseer, Khan, Hayat, Khan, and Porikli]{NRP}
Muzammal Naseer, Salman Khan, Munawar Hayat, Fahad~Shahbaz Khan, and Fatih Porikli.
\newblock A self-supervised approach for adversarial robustness.
\newblock In \emph{Proceedings of the IEEE/CVF Conference on Computer Vision and Pattern Recognition}, pp.\  262--271, 2020.

\bibitem[Oquab et~al.(2023)Oquab, Darcet, Moutakanni, Vo, Szafraniec, Khalidov, Fernandez, Haziza, Massa, El-Nouby, et~al.]{oquab2023dinov2}
Maxime Oquab, Timoth{\'e}e Darcet, Th{\'e}o Moutakanni, Huy Vo, Marc Szafraniec, Vasil Khalidov, Pierre Fernandez, Daniel Haziza, Francisco Massa, Alaaeldin El-Nouby, et~al.
\newblock Dinov2: Learning robust visual features without supervision.
\newblock \emph{arXiv preprint arXiv:2304.07193}, 2023.

\bibitem[Peng et~al.(2025)Peng, Tao, Wang, Wang, and Wang]{peng2025boosting}
Jinjia Peng, Zeze Tao, Huibing Wang, Meng Wang, and Yang Wang.
\newblock Boosting adversarial transferability via residual perturbation attack.
\newblock In \emph{Proceedings of the IEEE/CVF International Conference on Computer Vision}, pp.\  1261--1270, 2025.

\bibitem[Poursaeed et~al.(2018)Poursaeed, Katsman, Gao, and Belongie]{poursaeed2018generative}
Omid Poursaeed, Isay Katsman, Bicheng Gao, and Serge Belongie.
\newblock Generative adversarial perturbations.
\newblock In \emph{Proceedings of the IEEE conference on computer vision and pattern recognition}, pp.\  4422--4431, 2018.

\bibitem[Prakash et~al.(2018)Prakash, Moran, Garber, DiLillo, and Storer]{prakash2018deflecting}
Aaditya Prakash, Nick Moran, Solomon Garber, Antonella DiLillo, and James Storer.
\newblock Deflecting adversarial attacks with pixel deflection.
\newblock In \emph{Proceedings of the IEEE conference on computer vision and pattern recognition}, pp.\  8571--8580, 2018.

\bibitem[Radford et~al.(2021)Radford, Kim, Hallacy, Ramesh, Goh, Agarwal, Sastry, Askell, Mishkin, Clark, et~al.]{CLIP}
Alec Radford, Jong~Wook Kim, Chris Hallacy, Aditya Ramesh, Gabriel Goh, Sandhini Agarwal, Girish Sastry, Amanda Askell, Pamela Mishkin, Jack Clark, et~al.
\newblock Learning transferable visual models from natural language supervision.
\newblock In \emph{International conference on machine learning}, pp.\  8748--8763. PMLR, 2021.

\bibitem[Radosavovic et~al.(2020)Radosavovic, Kosaraju, Girshick, He, and Doll{\'a}r]{radosavovic2020regnet}
Ilija Radosavovic, Raj~Prateek Kosaraju, Ross Girshick, Kaiming He, and Piotr Doll{\'a}r.
\newblock Designing network design spaces.
\newblock In \emph{Proceedings of the IEEE/CVF conference on computer vision and pattern recognition}, pp.\  10428--10436, 2020.

\bibitem[Rao et~al.(2022)Rao, Zhao, Chen, Tang, Zhu, Huang, Zhou, and Lu]{rao2022denseclip}
Yongming Rao, Wenliang Zhao, Guangyi Chen, Yansong Tang, Zheng Zhu, Guan Huang, Jie Zhou, and Jiwen Lu.
\newblock Denseclip: Language-guided dense prediction with context-aware prompting.
\newblock In \emph{Proceedings of the IEEE/CVF conference on computer vision and pattern recognition}, pp.\  18082--18091, 2022.

\bibitem[Ronneberger et~al.(2015)Ronneberger, Fischer, and Brox]{ronneberger2015unet}
Olaf Ronneberger, Philipp Fischer, and Thomas Brox.
\newblock U-net: Convolutional networks for biomedical image segmentation.
\newblock In \emph{Medical image computing and computer-assisted intervention--MICCAI 2015: 18th international conference, Munich, Germany, October 5-9, 2015, proceedings, part III 18}, pp.\  234--241. Springer, 2015.

\bibitem[Rudin et~al.(1992)Rudin, Osher, and Fatemi]{rudin1992nonlinear}
Leonid~I Rudin, Stanley Osher, and Emad Fatemi.
\newblock Nonlinear total variation based noise removal algorithms.
\newblock \emph{Physica D: nonlinear phenomena}, 60\penalty0 (1-4):\penalty0 259--268, 1992.

\bibitem[Russakovsky et~al.(2015)Russakovsky, Deng, Su, Krause, Satheesh, Ma, Huang, Karpathy, Khosla, Bernstein, Berg, and Li]{imagenet}
Olga Russakovsky, Jia Deng, Hao Su, Jonathan Krause, Sanjeev Satheesh, Sean Ma, Zhiheng Huang, Andrej Karpathy, Aditya Khosla, Michael~S. Bernstein, Alexander~C. Berg, and Fei{-}Fei Li.
\newblock Imagenet large scale visual recognition challenge.
\newblock \emph{IJCV}, 2015.

\bibitem[Selvaraju et~al.(2017)Selvaraju, Cogswell, Das, Vedantam, Parikh, and Batra]{selvaraju2017gradcam}
Ramprasaath~R Selvaraju, Michael Cogswell, Abhishek Das, Ramakrishna Vedantam, Devi Parikh, and Dhruv Batra.
\newblock Grad-cam: Visual explanations from deep networks via gradient-based localization.
\newblock In \emph{Proceedings of the IEEE international conference on computer vision}, pp.\  618--626, 2017.

\bibitem[Shin et~al.(2022)Shin, Albanie, and Xie]{shin2022unsupervised}
Gyungin Shin, Samuel Albanie, and Weidi Xie.
\newblock Unsupervised salient object detection with spectral cluster voting.
\newblock In \emph{Proceedings of the IEEE/CVF Conference on Computer Vision and Pattern Recognition}, pp.\  3971--3980, 2022.

\bibitem[Singh et~al.(2023)Singh, Croce, and Hein]{singh2023convstem}
Naman~D Singh, Francesco Croce, and Matthias Hein.
\newblock Revisiting adversarial training for imagenet: Architectures, training and generalization across threat models.
\newblock In \emph{NeurIPS}, 2023.

\bibitem[Szegedy et~al.(2013)Szegedy, Zaremba, Sutskever, Bruna, Erhan, Goodfellow, and Fergus]{szegedy2013intriguing}
Christian Szegedy, Wojciech Zaremba, Ilya Sutskever, Joan Bruna, Dumitru Erhan, Ian Goodfellow, and Rob Fergus.
\newblock Intriguing properties of neural networks.
\newblock \emph{arXiv preprint arXiv:1312.6199}, 2013.

\bibitem[Szegedy et~al.(2016)Szegedy, Vanhoucke, Ioffe, Shlens, and Wojna]{inc-v3}
Christian Szegedy, Vincent Vanhoucke, Sergey Ioffe, Jonathon Shlens, and Zbigniew Wojna.
\newblock Rethinking the inception architecture for computer vision.
\newblock In \emph{CVPR}, 2016.

\bibitem[Tan \& Le()Tan and Le]{tan2104efficientnetv2}
Mingxing Tan and QV~Le.
\newblock Efficientnetv2: Smaller models and faster training. arxiv 2021.
\newblock \emph{arXiv preprint arXiv:2104.00298}.

\bibitem[Tan et~al.(2019)Tan, Chen, Pang, Vasudevan, Sandler, Howard, and Le]{tan2019mnasnet}
Mingxing Tan, Bo~Chen, Ruoming Pang, Vijay Vasudevan, Mark Sandler, Andrew Howard, and Quoc~V Le.
\newblock Mnasnet: Platform-aware neural architecture search for mobile.
\newblock In \emph{Proceedings of the IEEE/CVF conference on computer vision and pattern recognition}, pp.\  2820--2828, 2019.

\bibitem[Tarvainen \& Valpola(2017)Tarvainen and Valpola]{tarvainen2017meanteacher}
Antti Tarvainen and Harri Valpola.
\newblock Mean teachers are better role models: Weight-averaged consistency targets improve semi-supervised deep learning results.
\newblock \emph{Advances in neural information processing systems}, 30, 2017.

\bibitem[Tolstikhin et~al.(2021)Tolstikhin, Houlsby, Kolesnikov, Beyer, Zhai, Unterthiner, Yung, Steiner, Keysers, Uszkoreit, et~al.]{tolstikhin2021mlpmixer}
Ilya~O Tolstikhin, Neil Houlsby, Alexander Kolesnikov, Lucas Beyer, Xiaohua Zhai, Thomas Unterthiner, Jessica Yung, Andreas Steiner, Daniel Keysers, Jakob Uszkoreit, et~al.
\newblock Mlp-mixer: An all-mlp architecture for vision.
\newblock \emph{Advances in neural information processing systems}, 34:\penalty0 24261--24272, 2021.

\bibitem[Touvron et~al.(2021)Touvron, Cord, Douze, Massa, Sablayrolles, and J{\'e}gou]{touvron2021deit}
Hugo Touvron, Matthieu Cord, Matthijs Douze, Francisco Massa, Alexandre Sablayrolles, and Herv{\'e} J{\'e}gou.
\newblock Training data-efficient image transformers \& distillation through attention.
\newblock In \emph{International conference on machine learning}, pp.\  10347--10357. PMLR, 2021.

\bibitem[Trockman \& Kolter(2022)Trockman and Kolter]{trockman2022convmixer}
Asher Trockman and J~Zico Kolter.
\newblock Patches are all you need?
\newblock \emph{arXiv preprint arXiv:2201.09792}, 2022.

\bibitem[Tu et~al.(2022)Tu, Talebi, Zhang, Yang, Milanfar, Bovik, and Li]{tu2022maxvit}
Zhengzhong Tu, Hossein Talebi, Han Zhang, Feng Yang, Peyman Milanfar, Alan Bovik, and Yinxiao Li.
\newblock Maxvit: Multi-axis vision transformer.
\newblock In \emph{European conference on computer vision}, pp.\  459--479. Springer, 2022.

\bibitem[Wah et~al.(2011)Wah, Branson, Welinder, Perona, and Belongie]{cub}
C.~Wah, S.~Branson, P.~Welinder, P.~Perona, and S.~Belongie.
\newblock {The Caltech-UCSD Birds-200-2011 Dataset}.
\newblock Technical report, California Institute of Technology, 2011.

\bibitem[Wang et~al.(2018)Wang, Xu, Wang, and Tao]{wang2018perceptual}
Chaoyue Wang, Chang Xu, Chaohui Wang, and Dacheng Tao.
\newblock Perceptual adversarial networks for image-to-image transformation.
\newblock \emph{IEEE Transactions on Image Processing}, 27\penalty0 (8):\penalty0 4066--4079, 2018.

\bibitem[Wang et~al.(2022)Wang, Wu, Weng, Chen, Qi, and Jiang]{wang2022cross}
Rui Wang, Zuxuan Wu, Zejia Weng, Jingjing Chen, Guo-Jun Qi, and Yu-Gang Jiang.
\newblock Cross-domain contrastive learning for unsupervised domain adaptation.
\newblock \emph{IEEE Transactions on Multimedia}, 2022.

\bibitem[Wang et~al.(2021)Wang, He, Wang, and He]{Admix}
Xiaosen Wang, Xuanran He, Jingdong Wang, and Kun He.
\newblock Admix: Enhancing the transferability of adversarial attacks.
\newblock In \emph{Proceedings of the IEEE/CVF International Conference on Computer Vision}, pp.\  16158--16167, 2021.

\bibitem[Wang et~al.(2023)Wang, Yang, Feng, Sun, Guo, Zhang, and Ren]{wang2023towards}
Zhibo Wang, Hongshan Yang, Yunhe Feng, Peng Sun, Hengchang Guo, Zhifei Zhang, and Kui Ren.
\newblock Towards transferable targeted adversarial examples.
\newblock In \emph{Proceedings of the IEEE/CVF conference on computer vision and pattern recognition}, pp.\  20534--20543, 2023.

\bibitem[Wightman()]{Wightman_PyTorch_Image_Models}
Ross Wightman.
\newblock {PyTorch Image Models}.
\newblock URL \url{https://github.com/huggingface/pytorch-image-models}.

\bibitem[Wu et~al.(2025)Wu, Tan, Ma, Ma, Zhu, and Li]{wu2025boosting}
Shangbo Wu, Yu-an Tan, Ruinan Ma, Wencong Ma, Dehua Zhu, and Yuanzhang Li.
\newblock Boosting generative adversarial transferability with self-supervised vision transformer features.
\newblock \emph{arXiv preprint arXiv:2506.21046}, 2025.

\bibitem[Wu et~al.(2020)Wu, Su, Chen, Zhao, King, Lyu, and Tai]{wu2020boosting}
Weibin Wu, Yuxin Su, Xixian Chen, Shenglin Zhao, Irwin King, Michael~R Lyu, and Yu-Wing Tai.
\newblock Boosting the transferability of adversarial samples via attention.
\newblock In \emph{Proceedings of the IEEE/CVF Conference on Computer Vision and Pattern Recognition}, pp.\  1161--1170, 2020.

\bibitem[Xiao et~al.(2018)Xiao, Li, Zhu, He, Liu, and Song]{xiao2018generatingadvgan}
Chaowei Xiao, Bo~Li, Jun-Yan Zhu, Warren He, Mingyan Liu, and Dawn Song.
\newblock Generating adversarial examples with adversarial networks.
\newblock \emph{arXiv preprint arXiv:1801.02610}, 2018.

\bibitem[Xie et~al.(2018)Xie, Wang, Zhang, Ren, and Yuille]{xie2017randomization}
Cihang Xie, Jianyu Wang, Zhishuai Zhang, Zhou Ren, and Alan Yuille.
\newblock Mitigating adversarial effects through randomization.
\newblock In \emph{International Conference on Learning Representations}, 2018.
\newblock URL \url{https://openreview.net/forum?id=Sk9yuql0Z}.

\bibitem[Xie et~al.(2019)Xie, Zhang, Zhou, Bai, Wang, Ren, and Yuille]{DI}
Cihang Xie, Zhishuai Zhang, Yuyin Zhou, Song Bai, Jianyu Wang, Zhou Ren, and Alan~L Yuille.
\newblock Improving transferability of adversarial examples with input diversity.
\newblock In \emph{Proceedings of the IEEE/CVF conference on computer vision and pattern recognition}, pp.\  2730--2739, 2019.

\bibitem[Xie et~al.(2021)Xie, Wang, Yu, Anandkumar, Alvarez, and Luo]{xie2021segformer}
Enze Xie, Wenhai Wang, Zhiding Yu, Anima Anandkumar, Jose~M Alvarez, and Ping Luo.
\newblock Segformer: Simple and efficient design for semantic segmentation with transformers.
\newblock \emph{Advances in neural information processing systems}, 34:\penalty0 12077--12090, 2021.

\bibitem[Xie et~al.(2017)Xie, Girshick, Doll{\'a}r, Tu, and He]{xie2017resnext}
Saining Xie, Ross Girshick, Piotr Doll{\'a}r, Zhuowen Tu, and Kaiming He.
\newblock Aggregated residual transformations for deep neural networks.
\newblock In \emph{Proceedings of the IEEE conference on computer vision and pattern recognition}, pp.\  1492--1500, 2017.

\bibitem[Xu et~al.(2018)Xu, Evans, and Qi]{xu2017bitreduction}
Weilin Xu, David Evans, and Yanjun Qi.
\newblock Feature squeezing: Detecting adversarial examples in deep neural networks.
\newblock In \emph{NDSS}, 2018.
\newblock \doi{10.14722/ndss.2018.23295}.
\newblock URL \url{https://www.ndss-symposium.org/ndss-paper/feature-squeezing-detecting-adversarial-examples-in-deep-neural-networks/}.

\bibitem[Xue \& Chen(2024)Xue and Chen]{xue2024rethinking}
Haotian Xue and Yongxin Chen.
\newblock Rethinking adversarial attacks as protection against diffusion-based mimicry.
\newblock In \emph{Proceedings of the NeurIPS 2024 Workshop on Safe Generative AI}, 2024.
\newblock URL \url{https://neurips.cc/virtual/2024/106308}.
\newblock Poster.

\bibitem[Yang et~al.(2024{\natexlab{a}})Yang, Jeong, and Yoon]{yang2024facl}
Hunmin Yang, Jongoh Jeong, and Kuk-Jin Yoon.
\newblock Facl-attack: Frequency-aware contrastive learning for transferable adversarial attacks.
\newblock In \emph{Proceedings of the AAAI Conference on Artificial Intelligence}, volume~38, pp.\  6494--6502, 2024{\natexlab{a}}.

\bibitem[Yang et~al.(2024{\natexlab{b}})Yang, Jeong, and Yoon]{yang2024pdcl}
Hunmin Yang, Jongoh Jeong, and Kuk-Jin Yoon.
\newblock Prompt-driven contrastive learning for transferable adversarial attacks.
\newblock In \emph{European Conference on Computer Vision}, pp.\  36--53. Springer, 2024{\natexlab{b}}.

\bibitem[Yun et~al.(2020)Yun, Park, Lee, and Shin]{yun2020cskd_selfkd}
Sukmin Yun, Jongjin Park, Kimin Lee, and Jinwoo Shin.
\newblock Regularizing class-wise predictions via self-knowledge distillation.
\newblock In \emph{The IEEE/CVF Conference on Computer Vision and Pattern Recognition (CVPR)}, June 2020.

\bibitem[Zhang et~al.(2022{\natexlab{a}})Zhang, Tian, Tang, Chu, Wei, Shen, et~al.]{zhang2022segvit}
Bowen Zhang, Zhi Tian, Quan Tang, Xiangxiang Chu, Xiaolin Wei, Chunhua Shen, et~al.
\newblock Segvit: Semantic segmentation with plain vision transformers.
\newblock \emph{Advances in Neural Information Processing Systems}, 35:\penalty0 4971--4982, 2022{\natexlab{a}}.

\bibitem[Zhang et~al.(2021)Zhang, Karjauv, Benz, Ham, Cho, Youn, and Kweon]{zhang2021fgsm}
Chaoning Zhang, Adil Karjauv, Philipp Benz, Soomin Ham, Gyusang Cho, Chan-Hyun Youn, and In~So Kweon.
\newblock Is fgsm optimal or necessary for l$\infty$ adversarial attack?
\newblock In \emph{Workshop on Adversarial Machine Learning in Real-World Computer Vision Systems and Online Challenges (AML-CV)}. Computer Vision Foundation (CVF), IEEE Computer Society, 2021.

\bibitem[Zhang et~al.(2022{\natexlab{b}})Zhang, Li, Chen, Song, Gao, He, and Xue]{zhang2022beyond}
Qilong Zhang, Xiaodan Li, Yuefeng Chen, Jingkuan Song, Lianli Gao, Yuan He, and Hui Xue.
\newblock Beyond imagenet attack: Towards crafting adversarial examples for black-box domains.
\newblock In \emph{International Conference on Learning Representations}, 2022{\natexlab{b}}.

\bibitem[Zhang et~al.(2018)Zhang, Xiang, Hospedales, and Lu]{zhang2018dml}
Ying Zhang, Tao Xiang, Timothy~M Hospedales, and Huchuan Lu.
\newblock Deep mutual learning.
\newblock In \emph{Proceedings of the IEEE conference on computer vision and pattern recognition}, pp.\  4320--4328, 2018.

\bibitem[Zhao et~al.(2023{\natexlab{a}})Zhao, Chu, Liu, Li, Li, and Duan]{zhao2023m3d}
Anqi Zhao, Tong Chu, Yahao Liu, Wen Li, Jingjing Li, and Lixin Duan.
\newblock Minimizing maximum model discrepancy for transferable black-box targeted attacks.
\newblock In \emph{Proceedings of the IEEE/CVF conference on computer vision and pattern recognition}, pp.\  8153--8162, 2023{\natexlab{a}}.

\bibitem[Zhao et~al.(2023{\natexlab{b}})Zhao, Chu, Liu, Li, Li, and Duan]{zhao2023minimizing}
Anqi Zhao, Tong Chu, Yahao Liu, Wen Li, Jingjing Li, and Lixin Duan.
\newblock Minimizing maximum model discrepancy for transferable black-box targeted attacks.
\newblock In \emph{Proceedings of the IEEE/CVF conference on computer vision and pattern recognition}, pp.\  8153--8162, 2023{\natexlab{b}}.

\bibitem[Zhao et~al.(2022)Zhao, Yu, Sun, Zhang, and Wei]{zhao2022enhanced}
Shiji Zhao, Jie Yu, Zhenlong Sun, Bo~Zhang, and Xingxing Wei.
\newblock Enhanced accuracy and robustness via multi-teacher adversarial distillation.
\newblock In \emph{European Conference on Computer Vision}, pp.\  585--602. Springer, 2022.

\bibitem[Zhao et~al.(2025)Zhao, Zhang, Li, Sicre, Amsaleg, Backes, Li, Wang, and Shen]{zhao2025revisiting}
Zhengyu Zhao, Hanwei Zhang, Renjue Li, Ronan Sicre, Laurent Amsaleg, Michael Backes, Qi~Li, Qian Wang, and Chao Shen.
\newblock Revisiting transferable adversarial images: Systemization, evaluation, and new insights.
\newblock \emph{IEEE Transactions on Pattern Analysis and Machine Intelligence}, 2025.

\bibitem[Zhou et~al.(2022)Zhou, Wei, Wang, Shen, Xie, Yuille, and Kong]{zhou2021ibot}
Jinghao Zhou, Chen Wei, Huiyu Wang, Wei Shen, Cihang Xie, Alan Yuille, and Tao Kong.
\newblock ibot: Image bert pre-training with online tokenizer.
\newblock \emph{International Conference on Learning Representations (ICLR)}, 2022.

\bibitem[Zhou et~al.(2018)Zhou, Hou, Chen, Tang, Huang, Gan, and Yang]{zhou2018transferable}
Wen Zhou, Xin Hou, Yongjun Chen, Mengyun Tang, Xiangqi Huang, Xiang Gan, and Yong Yang.
\newblock Transferable adversarial perturbations.
\newblock In \emph{Proceedings of the European conference on computer vision (ECCV)}, pp.\  452--467, 2018.

\bibitem[Zhu et~al.(2024)Zhu, Liao, Zhang, Wang, Liu, and Wang]{zhu2024visionmamba}
Lianghui Zhu, Bencheng Liao, Qian Zhang, Xinlong Wang, Wenyu Liu, and Xinggang Wang.
\newblock Vision mamba: Efficient visual representation learning with bidirectional state space model.
\newblock In \emph{Forty-first International Conference on Machine Learning}, 2024.
\newblock URL \url{https://openreview.net/forum?id=YbHCqn4qF4}.

\bibitem[Zhu et~al.(2007)Zhu, Mumford, et~al.]{zhu2007stochastic}
Song-Chun Zhu, David Mumford, et~al.
\newblock A stochastic grammar of images.
\newblock \emph{Foundations and Trends{\textregistered} in Computer Graphics and Vision}, 2\penalty0 (4):\penalty0 259--362, 2007.

\bibitem[Zou et~al.(2023)Zou, Yang, Zhang, Li, Li, Wang, Wang, Gao, and Lee]{zou2023segment}
Xueyan Zou, Jianwei Yang, Hao Zhang, Feng Li, Linjie Li, Jianfeng Wang, Lijuan Wang, Jianfeng Gao, and Yong~Jae Lee.
\newblock Segment everything everywhere all at once.
\newblock \emph{Advances in neural information processing systems}, 36:\penalty0 19769--19782, 2023.

\end{thebibliography}
    \bibliographystyle{iclr2026_conference}
}

\newpage


\appendix

\section*{Supplementary Material}
In this supplementary material, we provide comprehensive insights and detailed resources that complement our main manuscript. 
First, we provide an in‐depth review of related work in Sec.~\ref{sec:relatedwork} and additional materials in Sec.~\ref{sec:additional_relatedwork}. 
We then highlight the distinctions of our method that clearly distinguish it from the concurrent works in Sec.~\ref{sec:supp_method_distinction}, 
describe the reason for enhanced adversarial transferability in Sec.~\ref{sec:supp_reasons_for_transferability}, 
and provide additional experimental details and results in Sec.~\ref{sec:additional_experiments} aimed at supplementing our manuscript for a better understanding of our approach.

\setcounter{figure}{0}
\setcounter{table}{0}
\setcounter{equation}{0}
\renewcommand{\thefigure}{S\arabic{figure}}
\renewcommand{\thetable}{S\arabic{table}}
\renewcommand{\theequation}{S\arabic{equation}}

\section{Related Work}
\label{sec:relatedwork}

\subsection{Transfer‐based Adversarial Attacks}  
Transfer‐based attacks exploit the empirical finding that adversarial perturbations crafted on one model often remain effective against others, even when architectures or training data differ. Early methods relied on iterative gradient‐based strategies—momentum‐integrated attacks (DI~\cite{DI}, TI~\cite{TI}), input‐diversity techniques (SI~\cite{SI}, Admix~\cite{Admix}), and strong baselines such as BIM~\cite{kurakin2016bim}, PGD~\cite{madry2017pgd}, and C\&W~\cite{carlini2019c_and_w}. These approaches enhance transferability via gradient smoothing, input transformations, and ensemble gradients, but incur heavy per-example optimization costs and often struggle against architectures that diverge significantly from the surrogate.  

More recent work has introduced generative frameworks that train feed-forward generators to synthesize perturbations in a single pass. GAP~\cite{poursaeed2018generative}, CDA~\cite{naseer2019cross}, and LTP~\cite{salzmann2021learning} demonstrated orders-of-magnitude speedups with comparable transfer rates. Subsequent advances BIA~\cite{zhang2022beyond}, GAMA~\cite{aich2022gama}, FACL-Attack~\cite{yang2024facl}, PDCL-Attack~\cite{yang2024pdcl}, and NAT~\cite{nakka2025nat} have further improved robustness by integrating logit or mid-level layer feature divergence, frequency-domain constraints, text prompt-driven, and neuron-targeted losses. 
In a similar lineage, targeted attacks~\cite{li2020towards, wang2023towards, zhao2023minimizing, fang2024clip, peng2025boosting} aim to steer the classifer into mispredicting as the target class, fooling the decision boundary towards that targeted class.
Rather than focusing solely on end-to-end optimization or domain-level constraints, we analyze the generator’s intermediate feature hierarchy and preserve semantic fidelity in its early blocks to steer perturbations onto object-centric regions, thereby enhancing cross-model transfer effectiveness.

\subsection{Generative Model-based Attacks}  
Generative attacks recast adversarial synthesis as a learning problem, training an image-to-image network (e.g.\ GAN or encoder–decoder) to produce perturbations in one pass. GAP~\cite{poursaeed2018generative} pioneered a framework in which the generator outputs adversarial noise that is then added to the input. CDA~\cite{wang2022cross} extends this by training a transformation network that directly outputs adversarial examples. Subsequent works incorporate perceptual losses based on surrogate logits~\cite{salzmann2021learning} and mid-level surrogate features~\cite{zhang2022beyond,nakka2025nat}. Building on the feature-similarity loss of Zhang et al.~\cite{zhang2022beyond}, more recent approaches leverage foundation models such as CLIP~\cite{CLIP}~\cite{aich2022gama,yang2024pdcl} and apply frequency-domain manipulations to surrogate features~\cite{yang2024facl}, further boosting transferability. 
Another concurrent work, dSVA~\cite{wu2025boosting}, innovates surrogate level manipulation by exploiting a dual self-supervised ViT ensemble features, which shows a different attack behavior as the previous works targeting a CNN surrogate.
While prior frameworks prioritize perturbation realism or frequency characteristics, we explicitly target the generator’s internal semantics by combining Mean Teacher-based smoothing with self-feature consistency on early blocks, preserving object contours and textures and concentrating adversarial perturbations in the most transferable regions. 

\subsection{Self-Knowledge Distillation}  
Self-knowledge distillation (Self-KD) aims to train a model to refine its own representations without an external teacher. Pioneering works in this field, Born-Again Networks~\cite{furlanello2018bornagain} and Deep Mutual Learning~\cite{zhang2018dml}, demonstrated that iterative self- and peer-distillation can improve generalization and robustness. Recent works~\cite{li2024dual_selfkd, yun2020cskd_selfkd} incorporate self-kd by aligning logits or intermediate features within its own network, or progressively updating the network~\cite{kim2021_pskd_selfkd}.
In this paradigm of using a student-teacher framework, the Mean Teacher framework~\cite{tarvainen2017meanteacher}, originally developed for semi-supervised learning, aims to maintain a teacher as the exponential moving average of the student’s weights, implicitly enforcing temporal consistency in predictions or feature maps. This EMA-based smoothing has been shown to reduce overfitting, stabilize training, and enhance domain invariance—properties that are directly relevant to generating perturbations that transfer across black-box models. Departing from classification-centric distillation, we integrate the Mean Teacher paradigm into a generative attack pipeline, using EMA to smooth intermediate features and enforcing hinge-based feature consistency on early blocks to preserve semantic integrity critical for cross-setting transferability.

\section{Additional Related Work} \label{sec:additional_relatedwork}
    \paragraph{Iterative optimization-based attacks.}
    For years, iterative gradient-based attacks have become a cornerstone of adversarial research. Methods such as Projected Gradient Descent (PGD)~\cite{madry2017pgd} extend the Fast Gradient Sign Method by applying multiple small, $\ell_{\infty}$-bounded steps; Momentum Iterative FGSM (MI-FGSM)~\cite{dong2018mifgsm} further stabilizes updates via accumulated momentum; Diverse Input FGSM (DI-FGSM)~\cite{DI} injects random resizing and padding at each iteration; and Translation-Invariant FGSM (TI-FGSM)~\cite{TI} averages gradients over shifted inputs to enhance spatial robustness. More advanced variants even incorporate feature-space objectives to target intermediate representations~\cite{zhang2021fgsm}. 

    \paragraph{Generative model-based attacks.}
    In parallel, more efficient generative model-based attacks train a feed-forward image-to-image transformation network to synthesize perturbations in a single pass: Universal Adversarial Perturbations (UAP)~\cite{moosavi2017universal} learn a single image-agnostic noise vector, Generative Adversarial Perturbations (GAP)~\cite{poursaeed2018generative} use a GAN framework to produce highly transferable noise maps (added to the input images), and AdvGAN~\cite{xiao2018generatingadvgan} leverages GANs for image-dependent attacks that balance stealth and speed. Together, these two paradigms offer complementary trade-offs between precision, transferability, and inference efficiency.

    In this vein, recent generative model-based untargeted attack methods~\cite{naseer2019cross, salzmann2021learning, zhang2022beyond, aich2022gama, yang2024facl, yang2024pdcl, nakka2025nat} have further added techniques to enhance the transferability of the crafted adversarial examples by incorporating surrogate model's output logit-level and mid-level feature-level separation, frequency domain manipulation, vision-language model guidance, and heuristic selection of one effective neuron-level generator among a pool of multiple generators. However, none of these works have dealt with directly manipulating the generative feature space to improve the transferability of AEs. To address this, we uncover the correlation between generative features and adversarial transferability of the output AEs.

    \noindent\textbf{Note:} Our semantically consistent generative attack does not redesign attack pipelines in a label-free or label-required setting. Rather, it operates orthogonally regardless of label availability: by regulating generator features, it can be plugged into any generative framework because its own objective is independent of label availability. This contrasts with prior work that mainly changes the adversarial loss, for example by moving from logit- to feature-based objectives.

    
    \paragraph{U-Net-based generator.}\label{sec:supp_unet}
    Along with ResNet~\cite{res152}, U-Net~\cite{ronneberger2015unet} is another effective network architecture comprising a symmetric encoder–decoder with skip connections, fusing low- and high-level features to preserve fine-grained details, which are ideal when perturbations must tightly follow object boundaries. By contrast, a ResNet generator stacks residual blocks with identity shortcuts, thus building deep hierarchical representations that emphasize global context. Although U-Net decoders add computational overhead, they can produce sharper, pixel-accurate noise, while ResNet backbones scale more efficiently and excel at generating broadly distributed perturbations. Ultimately, the choice of generator architecture hinges on the desired trade-off between pixel-level fidelity, attack transferability, and inference speed.

    In the context of generative adversarial attack, GAP~\cite{poursaeed2018generative} first demonstrated that U-Net can serve as a perturbation generator with a lower inference time cost than that with ResNet. However, the authors of \cite{poursaeed2018generative} also stated that ResNet, in general, outperforms U-Net in attack transferability. In this work, we demonstrate that our method of anchoring perturbation generation on early-intermediate features can also be applied to a different generator architecture than ResNet, namely U-Net. 

    \begin{wraptable}{r}{0.6\textwidth}
        \vspace{-4mm}

\centering
\caption{\textbf{Quantitative cross-task transferability results}. We report the average improvement ($\Delta$\%p) for our components applied to different generator architectures and evaluated against semantic segmentation (mIoU $\downarrow$) and object detection (mAP50 $\downarrow$) models. \texttt{MT} denotes mean teacher, and better results in \textbf{boldface}.}
\setlength{\tabcolsep}{2pt}
\renewcommand{\arraystretch}{1}
\renewcommand{\aboverulesep}{0.5pt}
\renewcommand{\belowrulesep}{0.5pt}
\vspace{-0.3cm}
\resizebox{\linewidth}{!}{
\begin{tabular}{cccccc|ccc}
    \toprule
        \multicolumn{2}{c}{\texttt{Cross-task}} 
        && \multicolumn{6}{c}{Generator Arch.} \\ \cmidrule{4-9}
        &&& \multicolumn{3}{c|}{\textbf{U-Net}} & \multicolumn{3}{c}{\textbf{ResNet}}\\ 
    \cmidrule{4-9}
         & \textbf{Method} 
         & \cellcolor{Goldenrod!25}{Benign}
         & \multicolumn{1}{c|}{Baseline} & \multicolumn{1}{c|}{+\texttt{MT}} & \multicolumn{1}{c|}{+\texttt{MT}+$\mathcal{L}_{\textrm{cons.}}$} 
         & \multicolumn{1}{c|}{Baseline} & \multicolumn{1}{c|}{+\texttt{MT}} & \multicolumn{1}{c|}{+\texttt{MT}+$\mathcal{L}_{\textrm{cons.}}$} \\
    \midrule
        \multirow{3}{*}{\makecell{\rotatebox{90}{SS}}} & DeepLabV3+~\cite{chen2018deeplabv3plus} & \cellcolor{Goldenrod!25}{76.21}
            & 24.22 & +0.76 & \textbf{-2.92}  
            & 23.89 & \textbf{-0.79} & \textbf{-1.84} \\ 
        & SegFormer~\cite{xie2021segformer} & \cellcolor{Goldenrod!25}{71.89} 
            & 29.34 & \textbf{-3.31} & \textbf{-3.81}
            & 25.60 & \textbf{-0.78} & \textbf{-0.85} \\
    \cmidrule{3-9}
        & \textbf{Avg.} & \cellcolor{Goldenrod!25}{74.05} 
            & 26.78 & \textbf{-1.27} & \textbf{-3.36} 
            & 24.75 & \textbf{-0.79} & \textbf{-1.35}\\
    \midrule
        \multirow{3}{*}{\makecell{\rotatebox{90}{OD}}} & FRCNN~\cite{girshick2015fasterrcnn} & \cellcolor{Goldenrod!25}{61.01}
            & 27.51 & \textbf{-0.05} & \textbf{-0.12} 
            & 28.43 & +0.03 & \textbf{-0.09} \\
        & DETR~\cite{carion2020detr} & \cellcolor{Goldenrod!25}{62.36} 
            & 23.92 & \textbf{-3.18} & \textbf{-3.54} 
            & 21.01 & \textbf{-0.02} & \textbf{-0.29} \\
    \cmidrule{3-9}
        & \textbf{Avg.} & \cellcolor{Goldenrod!25}{61.69} 
            & 25.72 & \textbf{-1.62} & \textbf{-1.83} 
            & 24.72 & +0.01 & \textbf{-0.20} \\
    \bottomrule
\end{tabular}
}
\label{tab:cross_task_unet}

        \vspace{-4mm}
    \end{wraptable}
    Specifically, we note the differences between the U-Net and ResNet generators in detail. Due to the symmetric encoder-decoder design in the U-Net, there is only one feature block at the \texttt{bottleneck} that we can employ as the intermediate block feature, as opposed to six in ResNet. Given this architecture, we applied our semantic consistency mechanism on this \texttt{bottleneck} feature for the cross-task experiments in Table~\ref{tab:cross_task_unet}, where we observe consistent improvements in attack transferability with the addition of each of our components. 
    While 
    Given this observation, we remark that while U-Net can still be leveraged as a generative model for transfer-based attacks, further research on boosting U-Net-based attack transferability may be necessary. 
    
    
    

    \paragraph{Diffusion-based generative attacks.}    
    Diffusion attacks generate imperceptible perturbations via iterative denoising~\cite{chen2024diffusion}. Later studies extend this idea to conditional or Stable Diffusion for black-box transfer~\cite{ liu2024boosting, liu2023improving, lei2025towards}, treat diffusion mimicry as a defense target~\cite{xue2024rethinking}, and push attacks to prompts, retrieval, or fully synthesized images~\cite{ma2024jailbreaking, huang2025huang, dai2024advdiff}. All rely on high sampling steps, often with extra classifier or CLIP guidance, so inference is markedly slower and heavier than single-shot GAN or gradient generators. Our framework avoids that cost with a fast, lightweight alternative suited to real-time or large-scale threats.
    While diffusion-based approaches are certainly relevant in the literature and offer strong performance, yet still incur high inference-time costs due to iterative sampling, our approach alleviates these costs and ensures faster and more practical deployment.

    \section{Distinctions of our method} \label{sec:supp_method_distinction}
    
    \paragraph{Purpose.}
    \noindent
    \textit{Generator-centric regularization vs. surrogate-centric ILP.}
    Prior ILP works typically optimize pixels against a feature map from a \emph{static surrogate model}. We instead ask how to regularize the synthesis process \emph{within a learnable generator} to enforce semantic stability. This shifts the focus from (external) surrogate optimization (followed by iterative ILP updates) to \emph{internal} regularization inside the generator. We summarize the distinctions of SCGA (Sec. \S 3 in the main paper)  below.

    \begin{itemize}[leftmargin=*] \setlength\itemsep{-0.2mm}
        \item \textbf{Attack framework.}
        Prior surrogate–centric ILPs update pixels iteratively under a surrogate classifier, typically requiring many small steps and sometimes fine-tuning existing adversarial examples. 
        Our approach is generator–centric: a single forward pass through a learnable perturbation generator uses the full perturbation budget at once, without any fine-tuning of existing AEs. 
        This yields a qualitatively different attack pipeline with essentially zero additional inference cost.
        
        \item \textbf{Source of guidance features.}
        Conventional ILPs extract guidance from fixed intermediate layers of a frozen surrogate. 
        We instead draw guidance from the intermediate representations \emph{inside} the learnable perturbation generator, thus internalizing what the model should preserve or alter. 
        Semantic information is therefore obtained from the generator’s own dynamics rather than from an external model.
        
        \item \textbf{Role of intermediate features (optimization objective).}
        In surrogate–centric ILPs, the surrogate’s mid-level features directly define the objective, commonly by maximizing the feature distance between a benign image and its adversarial counterpart. 
        In our method, the generator’s intermediate features are regularized to stabilize core semantics via an EMA teacher, providing self-guided regularization during noise synthesis. 
        Whereas ILPs feed both benign and adversarial images to the surrogate, our generator consumes only the benign image, producing internal features with different embedded semantics.
        
        \item \textbf{Driving factor for transferability.}
        Classic ILPs largely seek stronger disturbance in surrogate mid-layer features to improve transfer. 
        Our method maintains salient object semantics throughout the noise-generation path, emphasizing semantic consistency in the generator’s early blocks. 
        This distinction shifts the driver of transfer from surrogate divergence to internally consistent noise generation.
        \item \textbf{Plug-in compatibility.}
        Traditional ILP attacks are usually self-contained and not designed for modular composition. 
        Our regularizer is a drop-in module that plugs into existing generators (e.g., BIA, GAMA) without changing their inference routine. 
        It thus offers a general-purpose axis of improvement for generative attacks.
    \end{itemize}    
    
    The seminal \emph{Intermediate-Level Attack (ILA)}~\cite{huang2019enhancing} begins with a baseline perturbation and amplifies its change at a single mid-layer of a frozen classifier, boosting cross-model transfer. \emph{ILA++}~\cite{li2020ilaplusplus} maximizes the scalar projection onto a learned discrepancy vector, making the amplification direction data-adaptive. \emph{ILPD}~\cite{li2023ilpd_attack} folds amplification into one stage and adds a decay schedule to damp spurious directions. \emph{TAP}~\cite{zhou2018transferable} enlarges clean--adversarial feature distances while imposing a smoothness prior on the noise. All rely on a \emph{fixed surrogate classifier} to define the intermediate layer and therefore lose potency when the victim architecture or modality changes.
    
    \medskip
    \noindent\textit{Generator-based ILPs.}
    A parallel line supervises a generator with surrogate features. \emph{LTP} splits a chosen surrogate layer into class-consistent and class-inconsistent channels and steers the generator toward the latter. \emph{BIA} manipulates early surrogate features to weaken low-level cues across domains. Recent works impose high-level semantic or contrastive losses (often via CLIP) on the \emph{output image} but leave the generator's internal layers largely unconstrained.
    
    \medskip
    \noindent\textit{Our focus on generator internals.}
    We align early-block feature maps to an EMA-smoothed teacher and distill onto the student, preserving coarse structure \emph{before} any surrogate-level adversarial loss is applied. This internal alignment is agnostic to the choice of surrogate or adversarial loss, and remains effective when transferring to unseen architectures. Empirically, adding our self-feature consistency to representative generator attacks further \emph{lowers average accuracy across all four cross-setting protocols}, indicating complementarity rather than redundancy.
    
    \paragraph{Structure.}
    \begin{wraptable}{r}{0.6\textwidth}
        \vspace{-4mm}
        
\centering
\caption{\textbf{Our method distinction.} Comparison of transfer-based generative adversarial attacks, highlighted by the method's targeted stage in the training pipeline (in order from left to right), and GT label requirement.}
\setlength{\tabcolsep}{1pt}
\vspace{-3mm}
\renewcommand{\arraystretch}{1}
\renewcommand{\aboverulesep}{1pt}
\renewcommand{\belowrulesep}{1pt}
\resizebox{\linewidth}{!}{
\begin{tabular}{ccccccccccccc}
    \toprule
        \textbf{Attack} & \textbf{\shortstack{Input data\\aug.}} && \cellcolor{Goldenrod!25} \textbf{\shortstack{Generator\\feature-level}} && \textbf{\shortstack{Perturbed\\image-level}} && \textbf{\shortstack{\underline{Surrogate}\\mid-level layer\\feature-level}} && \textbf{\shortstack{\underline{Surrogate}\\output\\logit-level}} && \textbf{\shortstack{GT label\\required?}}\\
    \midrule
        GAP~\cite{poursaeed2018generative} & - && \cellcolor{Goldenrod!25} - && - && - && \textbf{\cmark} && \cmark / \xmark\\ 
        CDA~\cite{wang2022cross} & - && \cellcolor{Goldenrod!25} - && - && - && \textbf{\cmark} && \cmark / \xmark \\
        LTP~\cite{salzmann2021learning} & - && \cellcolor{Goldenrod!25} - && - && \textbf{\cmark} && - && - \\
        BIA~\cite{zhang2022beyond} & - && \cellcolor{Goldenrod!25}  - && \textbf{\cmark}
        && \textbf{\cmark} && - && - \\
        GAMA~\cite{aich2022gama} & - && \cellcolor{Goldenrod!25} - && - &&  \textbf{\cmark} && - && \textbf{\cmark}\\
        FACL~\cite{yang2024facl} & \textbf{\cmark} && \cellcolor{Goldenrod!25}  - && -  && \textbf{\cmark} && - && -\\
        PDCL~\cite{yang2024pdcl} & - && \cellcolor{Goldenrod!25} - && - && \textbf{\cmark} && - && \textbf{\cmark} \\
    \midrule
        Our focus & - && \cellcolor{Goldenrod!25}  \textbf{\cmark} && - && - && - && -\\
    \bottomrule
\end{tabular}
}
\label{tab:supp_method_disctinction}

        \vspace{-4mm}
    \end{wraptable}
    Our method stands distinct from the focus of existing generative attacks in that we delve into the generator feature space, rather than the surrogate model space, as categorized in Table~\ref{tab:supp_method_disctinction}. 
    Previous works have targeted various stages of the generative attack pipeline~\cite{poursaeed2018generative, naseer2019cross, salzmann2021learning, zhang2022beyond}, including input data augmentation, pixel-level perturbation, surrogate model's logit- and feature-level manipulations. Nonetheless, no work has yet explicitly manipulated the internal features of the generative model to enhance transferability. In this work, we investigate how internal feature representations within generative models can be harnessed to enhance the transferability of AEs.

    \paragraph{Distinction from ensemble-based approach.}
    We assert that our proposed method is intended as a complementary add-on to the single perturbation generator already employed by existing generative attacks, rather than as a separate generator. As such throughout our experiments, we meticulously focus on the effect of attaching our method onto existing perturbation generators and show that our complementary add-on exhibits attack-beneficial effects. Though we do agree that an ensemble of different generators may be an interesting direction of work, the scope in our work focuses more on a complementary add-on for existing adversarial generative perturbation works.
    
    \paragraph{Full algorithm.}
    For a full picture of training and inference stage of our algorithm, we provide in Alg.~\ref{alg:alg1_full} outlining the procedure in both stages.

\begin{algorithm}[!h]
    \DontPrintSemicolon
    \caption{Full pseudo-code of SCGA}
    \label{alg:alg1_full}
    \KwData{Training dataset $\mathcal{D}_{src}$}
    \KwInput{Generator $\mathcal{G}_{\theta}(\cdot)$, a surrogate model trained on source data $\mathcal{F}^{s}(\cdot)$, projector $\mathcal{P}(\cdot)$, perturbation budget $\epsilon$}
    \KwOutput{Optimized teacher perturbation generator $\mathcal{G}_{\theta'}(\cdot)$}
    \smallskip
    \CommentSty{Training:}\;
    Initialize generators: \\
    student $\mathcal{G}_{\theta}(\cdot) \gets$ random init., 
    teacher $\mathcal{G}_{\theta'}(\cdot) \gets \mathcal{G}_{\theta}(\cdot)$ \;
    \Repeat{$\mathcal{G}_{\theta}(\cdot)$ converges}{
        Randomly sample a mini-batch $x_{i}$ from $\mathcal{D}_{\textrm{src}}$\;
        Acquire student generator intermediate features: \quad\quad  $\mathbf{g}_{i} \gets \mathcal{G}_{\theta}^{\textrm{enc}}(x_{i})$\;
        Acquire teacher generator intermediate features: \quad\quad  $\mathbf{g}_{i}' \gets \mathcal{G}_{\theta'}^{\textrm{enc}}(x_{i})$\;
        Generate unbounded adversarial examples from student generator intermediate features: \quad\quad $\tilde{x}_{i}^{\textrm{adv}} \gets \mathcal{G}_{\theta}^{\textrm{dec}}(\mathbf{g}_{i})$\;
        Bound (project) $\tilde{x}_{i}^{\textrm{adv}}$ using $\mathcal{P}$  within the perturbation budget such that $||\mathcal{P}(\tilde{x}_{i}^{\textrm{adv}}) - x_{i}||_{\infty} \le \epsilon
        $ to obtain $x_{i}^{\textrm{adv}}$\;
        Forward pass $x_{i}$ and $x_{i}^{\textrm{adv}}$ through the surrogate model, $\mathcal{F}^{s}(\cdot)$ at layer $k$, to acquire $f_{i}^{\textrm{benign}}, f_{i}^{\textrm{adv}}$\;
        Compute loss using $f_{i}^{\textrm{benign}}, f_{i}^{\textrm{adv}}$, $\mathbf{g}_{i}, \mathbf{g}_{i}'$: \quad\quad $\mathcal{L} = \mathcal{L}_{\textrm{adv}} + \lambda_{\textrm{cons.}}\cdot\mathcal{L}_{\textrm{cons.}}$ \hfill\tcp{Eq.\ref{eq:loss}}
        Update student generator parameters via backpropagation\;
        EMA update teacher weights with student weights: \quad\quad  $\theta \mapsto \theta'$\hfill\tcp{Eq.~\ref{eq:ema_update}}
    }
    \CommentSty{Inference:}\;
    Acquire an input image sample, $x_{\textrm{test}}$\;
    Forward pass $x_{\textrm{test}}$ through the trained teacher, $\mathcal{G}_{\theta'}(\cdot)$, to obtain an unbounded adversarial example, $\tilde{x}_{\textrm{test}}^{\textrm{adv}}$\;
    Bound (project) $\tilde{x}_{\textrm{test}}^{\textrm{adv}}$ using $\mathcal{P}$ within $\epsilon_{\textrm{test}}$ to obtain $x_{\textrm{test}}^{\textrm{adv}}$\;
    Forward pass $x_{\textrm{test}}$, $x_{\textrm{test}}^{adv}$ through pre-trained victim model $\mathcal{F}^{t}(\cdot)$ to obtain $p_{\textrm{test}}^{\textrm{out}}$, $p_{\textrm{test}}^{\textrm{out,adv}}$, respectively\;
    Compute metric scores by comparing the $\argmax \{p_{\textrm{test}}^{\textrm{out,adv}}\}$ against $\argmax \{p_{\textrm{test}}^{\textrm{out}}\}$ or GT label
    
\end{algorithm}

\section{Reasons for transferability} \label{sec:supp_reasons_for_transferability}

    As our work is motivated by the semantic variability across the intermediate blocks within the generator, we look further into how our early-block semantic anchoring drives the observed phenomenon in Sec. \S 2.2 of the main paper. 
    Our empirical analysis of intermediate feature activation maps from existing ResNet-based generators reveals that coarse, object-salient regions consistently emerge in the early residual blocks, and appear even more pronounced in models with higher black-box transferability. This insight suggests that these early‐block features play a pivotal role in shaping perturbations. To capitalize on this, we anchor our adversarial noise generation to the clean image’s semantic structure at these early stages. Lacking explicit semantic priors to retain the semantic integrity of the benign input images, we introduce a Mean Teacher mechanism: by maintaining an exponential moving average (EMA) of the student generator’s weights, the teacher generator yields temporally smoothed features that are \emph{largely free of adversarial noise.} We then fully leverage the Mean Teacher framework by further imposing a self-feature consistency loss between the student generator’s and the teacher’s early-intermediate block activations, \emph{filtering out spurious noise while preserving the coarse object shapes and boundaries} present in the teacher generator features. This strict semantic-consistency constraint focuses perturbation power on object-salient regions, thereby enhancing transferability without sacrificing efficiency.

\begin{figure*}[!h]
        \centering
        \includegraphics[width=\linewidth]{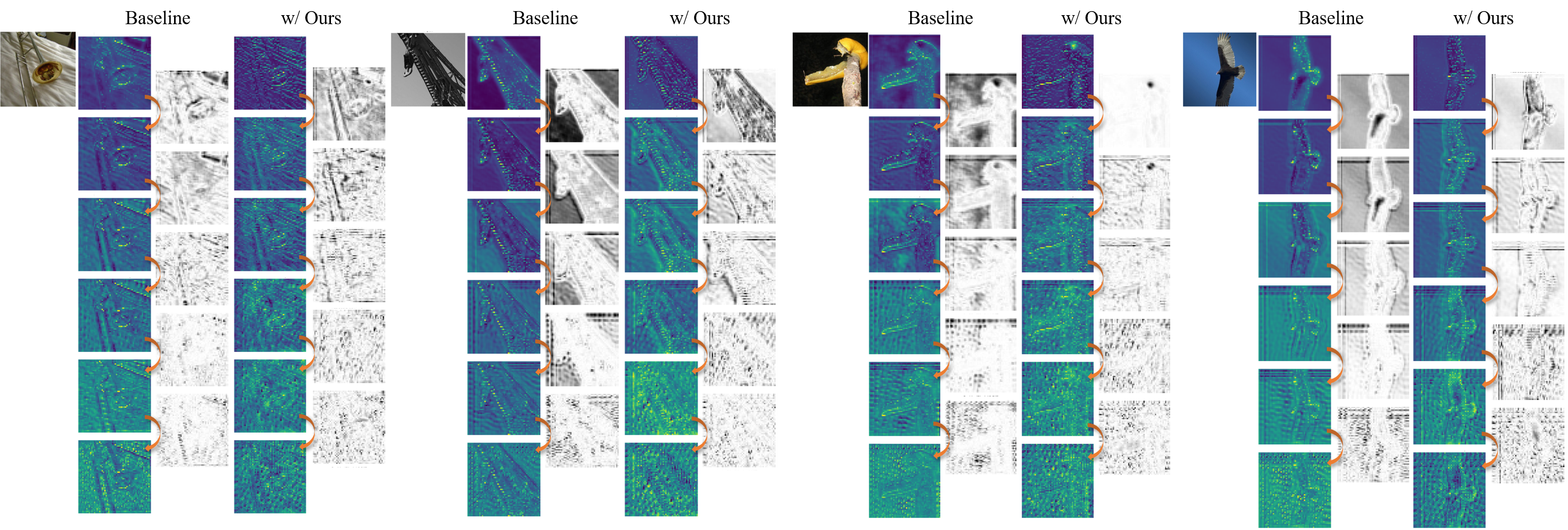}
        \caption{Comparison of the input image (column 1), the feature activation maps, and the added adversarial noise between two residual blocks (absolute difference between the feature maps of two residual blocks) of Ours (columns 4, 5) against the baseline~\cite{zhang2022beyond} (columns 2, 3).
        Although the baseline’s feature maps more vividly emphasize object boundaries and contours, this focus actually prevents perturbations from appearing in those highlighted regions. By contrast, our method produces relatively less pronounced early features than those of the baseline, yet focuses more perturbation power on object-salient regions towards the later blocks than the baseline, thereby allowing adversarial noise to be dispersed directly on and around those salient regions, as opposed to the baseline. For the feature activation maps and the added noise, the brighter and darker (respectively), the higher the value.
        }
        \label{fig:supp_block_ablation2}
    \end{figure*}

    In the figure below (Fig.~\ref{fig:supp_block_ablation2}), we directly compare the feature activation maps and the added adversarial noise per block (absolute difference of the input and output of each block) of Ours against the baseline~\cite{zhang2022beyond}. We particularly focus on the intermediate residual blocks (``Residual Learning"), as most of the adversarial noise is generated in these blocks~\cite{zhang2022beyond}, and the preceding (``Downsampling" layers) and succeeding blocks (``Upsampling" layers) serve to simply adjust the spatial resolution of the feature maps. 
    
    We preserve semantic integrity in the early blocks because these layers capture the coarse structure of the object, such as boundaries and shapes. By aligning the student model’s early block activations to a teacher reference, we remove incidental details and initial noise. This alignment compresses feature magnitudes and lowers the measured semantic quality in those early blocks. However, that simpler representation allows the generator to focus on stronger and more widespread noise in the later layers. 
    
    \paragraph{Deliberate compression of early features.}
    We enforce semantic consistency in the early blocks to focus the generator on true object outlines. Aligning student features to a smoothed teacher strips away incidental detail and any initial noise, which reduces the average magnitude of feature activations, and thus appears to degrade early‐block semantics. That lean representation then lets the network concentrate its available capacity deeper in the intermediate blocks, where it produces stronger and more widely dispersed perturbations. When comparing the absolute difference between two feature maps in Fig.~\ref{fig:supp_block_ablation2}, we see that the baseline avoids object‐salient regions and restricts noise to the peripheral regions, whereas our approach applies noise across the entire image, including the object itself. Although the early semantics seem more degraded relative to the baseline, this deliberate compression of early features enables a broader attack on diverse features. Although the baseline’s early blocks may exhibit stronger semantic activations, our method’s slightly muted early features enable a broader and more effective perturbation distribution towards the later intermediate blocks, thus achieving higher transferability than the baseline. Crucially, we observe that \emph{the baseline’s finely detailed early-block semantics add little benefit; retaining only the coarse semantic outline is sufficient to guide highly transferable perturbations.}


    \paragraph{Comparison of feature activation maps by block.}
    \begin{wraptable}{r}{0.4\textwidth}
        \vspace{-12mm}
        \centering
        \includegraphics[width=\linewidth]{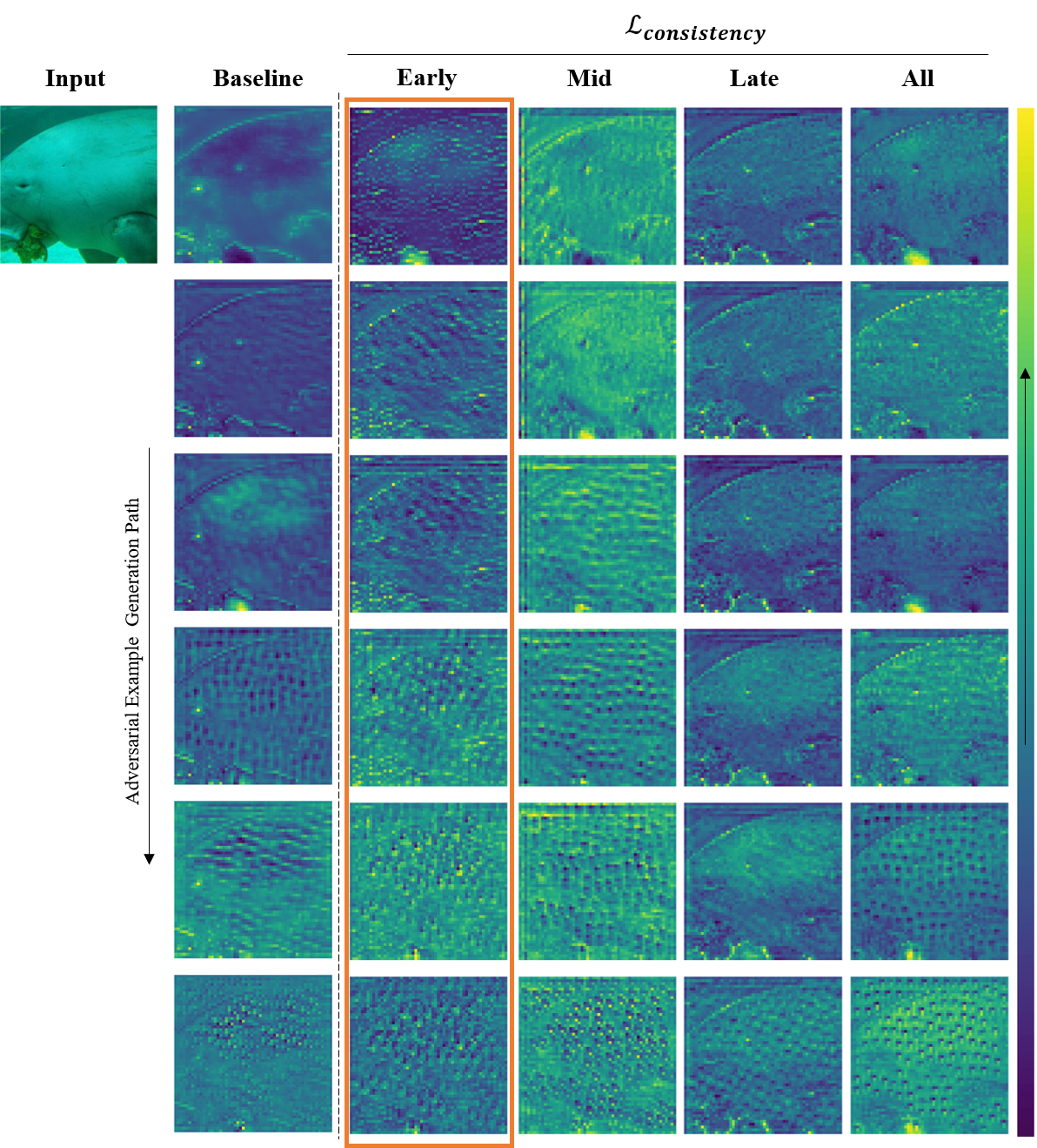}
        \vspace{-5mm}
        \captionof{figure}{Comparison of feature activation maps by block.}
        \label{fig:supp_block_ablation1}
        \vspace{-7mm}
    \end{wraptable}
    Compared to the baseline and to ablations that apply semantic consistency in mid layers, late layers, or across all layers, our approach of enforcing semantic consistency in the early block produces the most effective and transferable perturbations (Fig.~\ref{fig:supp_block_ablation1}). 
    When we inspect feature activation maps, we see that preserving the coarse semantic structure in the earliest layers anchors the noise to object-salient regions that tend to be shared across different models. This focus prevents the generator from wasting capacity on irrelevant details and guides it to concentrate its attack on universally important features. By contrast, semantic consistency applied later or across all intermediate blocks causes noise to be rather sparse or dispersed over regions away from those core cues, resulting in weaker transfer performance.


    \paragraph{Quantifying semantic information.} \label{sec:supp_quantifying_semantic}
     The very concept of \textit{semantic information} is a deep and challenging topic within information theory and computer vision, lacking a single, universally accepted, and operational definition. For example, foundational theories define meaning through the truthfulness of a representation to reality~\cite{floridi2002philosophy}, or the grammatical structure of its components~\cite{zhu2007stochastic}. While profound, these conceptual frameworks remain abstract. Attempts to formalize a quantifiable theory of semantics, such as the information-theoretic treatise by \cite{d2011quantifying}, highlight this challenge. While providing a valuable formal basis, such work has yet to yield operational metrics that can be optimized directly within deep learning models from the first principles.
   
    Given this well-documented gap between theory and practice, the standard and most rigorous approach in computer vision is to probe semantic coherence using concrete, multi-faceted empirical methods. To build a robust foundation for our \textit{semantic anchoring on the early generator intermediate block} claim, we employed two orthogonal analyses
    to test our hypothesis from different perspectives: 
    \vspace{-2mm}
    \begin{itemize}[leftmargin=*] \setlength \itemsep{-0.1mm}
        \item \textbf{Energy-Level Stability (Frequency Domain)}: To verify the \textit{coarse} structure embedded in the \textit{low} frequency components, we computed the spectral energy ratio compared to the total by band following Eq.~\ref{eq:supp_freq} (cutoff=0.2). Our spectral analysis confirms that our method preserves low-frequency energy more effectively than the baseline. This provides evidence that the coarse visual structure of the image is maintained throughout the generation process.        
        \item \textbf{Object-Level Coherence (Semantic Object)}: A closer look into semantic foreground IoU stability can be viewed in a complementary manner alongside the spectral energy analysis. We conducted k-means clustering on each intermediate block features, generated binary (object/background) object masks for each block features, and measured their IoU against the ground truth pixel-wise labels of ImageNet-S~\cite{gao2022imagenet-S} dataset. We then calculated the standard deviation of these  IoU scores across the generator's blocks--a metric we term as \textit{vararibility} in IoU. A lower variability score signifies that the object's core structure is preserved more coherently across generator blocks during generation. Our method achieves a significantly lower variance than the baseline, confirming superior object-level stability.
    \end{itemize}    
    \begin{align}
    \footnotesize
        \label{eq:supp_freq}
        F(u,v) &= \sum_{m=0}^{H-1}\sum_{n=0}^{W-1}
            f[m,n]\;\exp\!\Big(-\mathrm{i}2\pi\big(\tfrac{um}{H}+\tfrac{vn}{W}\big)\Big),\;\; \text{for feature map} \;f\qquad \text{(FFT)} \\ \nonumber
        S(u,v) &= \big|F(u,v)\big|^{2}, \qquad \text{(Power spectrum)}\\\nonumber
        R_{\mathrm{low}}(\tau) 
            &= \frac{\displaystyle \sum_{k,\ell}\,
                \mathbf{1}\!\big\{\sqrt{k^{2}+\ell^{2}}\le \tau\big\}\, S(k,\ell)}
               {\displaystyle \sum_{k,\ell} S(k,\ell)}, \label{eq:supp_freq3} \\\nonumber
        R_{\mathrm{high}}(\tau) 
            &= 1 - R_{\mathrm{low}}(\tau). 
    \end{align}
    Although a first-principles theory of semantics remains an open challenge for the broader scientific community, our work provides a strong and verifiable empirical basis for our claim. The combined evidence from both energy-level (frequency) and object-level (foreground IoU) analyses demonstrates that early-generator block anchoring effectively maintains semantic content, commonly shared by diverse architectures across domains. This coherence, in turn, provides a clear explanation for the observed boost in adversarial transferability across architectures, domains and tasks.

    \paragraph{Marginal improvement due to shared objective.}
    We acknowledge that PDCL~\cite{yang2024pdcl} is the strongest amongst the existing generative attack baselines. Our approach is complementary yet convergent with ideals of PDCL and GAMA~\cite{aich2022gama} in attacking the semantics. While PDCL and GAMA leverage CLIP~\cite{CLIP}-based semantics alignment in surrogate models, our method uniquely targets semantic distortion within the internal feature hierarchy of the generator via Mean Teacher smoothing and self-feature consistency. Though previous methods and our method work in different operational spaces (internal feature hierarchy of the generator vs surrogate model), the aligned goals as well as the generator and surrogate features interacting through backpropagation leads to diminishing returns when combined. This ceiling effect explains why numerical improvements over PDCL are modest in cross-model settings.

    However, we stress that while the cross-model results exhibit only minor degradation, our method yields more pronounced gains over the PDCL baseline in cross-domain and cross-task settings, where data and model shifts are more severe. Notably, in cross-domain results, while ASR, FR, and Accuracy consistently improve, we observe a drop in ACR with ours compared to PDCL alone. This suggests that PDCL’s strong surrogate alignment may overfit to high-level features, inadvertently leading to corrected predictions under domain shift. Our ACR metric uniquely captures this effect, which goes unnoticed by conventional metrics.   
    Ultimately, we believe this demonstrates that our method contributes in more realistic and challenging generalization settings (cross-domain, cross-task) than in the saturated cross-model setting. We acknowledge that this observation remains as a potential limitation of our method when layered atop strong vision-language aligned baselines such as PDCL. Thus, even when improvement margins are modest, our work introduces a conceptually novel and generator-centric mechanism that is compatible with PDCL, helps expose previously overlooked behaviors (via ACR), and delivers consistent and transferable performance boosts, especially under the strictest black-box constraints.
    
    \noindent In summary, our ablations reveal two situations in which the proposed early-block self-feature consistency provides only modest gains:
    \vspace{-2mm}
    \begin{itemize}[leftmargin=20pt] \setlength\itemsep{0.1em}
        \item \textbf{Transfer to dense–prediction task}: When evaluated on tasks that require pixel-level precision such as SS, our hinge-based self-consistent generator produces less disruptive AEs. In the cross-task setting, DeepLabV3+ IoU decreases by 1.35\%p and Faster R-CNN mAP50 falls by 0.20\%p. Figs. S8 and S9 show that the perturbations suppress large regions but rarely erode thin boundaries or very small objects. Because the hinge margin halts gradients once coarse alignment is achieved, optimization focuses on low-frequency structure, which benefits cross-model and cross-domain transfer more than cross-tasks that require fine-grained spatial details.
        \item \textbf{Combination with CLIP-guided attacks}: At $\epsilon_{\textrm{test}}=10$, augmenting GAMA with ours lowers cross-domain acc. by 2.47\%p but trims model accuracy by only 1.18\%p and mAP50 by 0.16\%p; PDCL changes are even smaller, all ≤ 1\%p. Because GAMA and PDCL already use a CLIP image-text similarity loss that enforces high-level global semantic alignment, our early-block anchor largely overlaps their objective, leaving limited headroom. Future work could involve adaptive margins that complement rather than replicate CLIP guidance.
    \end{itemize}

    \paragraph{Potentially compatible foundation model alternatives.}
    While CLIP is a powerful vision-language foundation model, its representations are heavily shaped by global semantic alignment across modalities. This design prioritizes coarse, aligned image-text associations and often abstracts away local or mid-level spatial structure, which is precisely the kind of structure our generator-centric perturbations are designed to manipulate. As a result, our structure-aware perturbations, which emphasize consistency in early generator layers (e.g., edges, object boundaries), may be less impactful when paired with CLIP-based classifiers, leading to the observed marginal improvements.

    Nonetheless, our method and CLIP-guided approaches share a high-level goal: disrupting semantic integrity in learned feature spaces. CLIP achieves this via global cross-modal alignment, whereas our method directly injects semantic distortions during perturbation synthesis. This overlap may lead to saturation effects when stacking the two, particularly when the downstream classifier lacks spatial sensitivity.
    To this end, we list below several foundation models that emphasize semantic structure, rather than purely global or language-aligned semantics may synergize better with our approach. Promising alternatives include:
    \vspace{-2mm}
    \begin{itemize}[leftmargin=20pt] \setlength\itemsep{0.1em}
        \item \textbf{DINOv2}~\cite{oquab2023dinov2} and \textbf{iBOT}~\cite{zhou2021ibot}: self-supervised, image-only models that retain strong spatial awareness and token-level feature localization. Their patch-level attention maps are often more structure-sensitive in early layers.
        \item \textbf{SAM} (Segment Anything)~\cite{kirillov2023sam} and \textbf{OWL-ViT}~\cite{minderer2022simple}: models trained with region-aware objectives or bounding box supervision, offering spatial grounding across scales.
        \item \textbf{SEEM}~\cite{zou2023segment}, \textbf{OpenSeg}~\cite{liang2023open}, and other dense prediction models: trained for segmentation or region-level tasks, these retain fine spatial resolution across semantic hierarchies.
    \end{itemize}

    In addition, we would like to highlight \textbf{TokenCut}~\cite{shin2022unsupervised}, which encourages part-aware decomposition in early transformer layers via spectral clustering. Similarly, \textbf{SegViT}~\cite{zhang2022segvit} and \textbf{DenseCLIP}~\cite{rao2022denseclip} enhance token-level representations with structural priors for region-specific prediction. These architectures inherently encode mid-level part semantics and object contours, aligning more naturally with our method’s emphasis on preserving and perturbing meaningful structure.
    Overall, these models offer varying degrees of structure sensitivity, depending on task supervision and architectural inductive biases. We view this as both a current limitation and a valuable opportunity: as foundation models increasingly integrate dense, structured objectives, our generator-centric perturbations could target and amplify the resulting spatial semantics in a more compatible way.   

    \paragraph{Practical value of the Accidental Correction Rate (ACR).}
    Here, we highlight the usefulness of the proposed ACR metric as previously outlined in Table 1 of the main paper. ACR was introduced specifically to expose cases in which a perturbation repairs an error already present in the clean prediction: a behavior that Accuracy, Attack‐Success Rate (ASR) and Fooling Rate (FR) cannot disentangle. In this light, we argue that \emph{this metric is of significance for both adversary and defender}, as it focuses on the reliability of the attack and haphazard defense due to an inadvertent transition to \textit{correct} predictions, respectively.
    In Table~\ref{tab:supp_eps_acr}, we report all four metrics for increasing test budgets $\epsilon_{\textrm{test}}$, averaged over the same cross-domain / cross-model settings and pinpoint key observations:
    \vspace{-2mm}
    \begin{itemize}[leftmargin=20pt] \setlength\itemsep{0.1em}
        \item \textbf{Complementary role to ASR.} At $\epsilon=4$, ACR (incorrect→correct) is ~14\% while ASR (correct→incorrect) is 8\%, yielding a net gain of ~6\% correct predictions. Because ASR accounts for only harmful flips, it overlooks this positive balance; ACR makes it explicit.
        \item \textbf{ACR is non-monotonic, unlike Acc/ASR/FR.} It peaks at ε = 4 and then falls as stronger noise overwhelms corrective effects. This trend offers a very different view from the defender’s side: evaluation should consider not only how many errors an attack creates but also how many it inadvertently corrects.
        \item \textbf{Actionable insight.} Defenders might tolerate or even harness low-budget perturbations that raise ACR, while attackers in safety-critical settings may need to penalize accidental corrections to avoid unintentionally improving model performance.
    \end{itemize}
    
    \begin{table}[!h]
        \centering
        \setlength{\tabcolsep}{8pt}
        \renewcommand{\arraystretch}{1}
        \renewcommand{\aboverulesep}{0.5pt}
        \renewcommand{\belowrulesep}{0.5pt}
        \caption{Performance evaluation under different $\varepsilon$ test values for other metrics than the accuracy.}
        \vspace{-2mm}
        \label{tab:supp_eps_acr}
        \resizebox{.7\linewidth}{!}{
        \begin{tabular}{c|cccc}
        \toprule
            $\varepsilon_{test}$ & Accuracy $\downarrow$ & ASR $\uparrow$ & FR $\uparrow$ & ACR $\downarrow$ \\
        \midrule
            2 & 89.93 / 74.67 & 1.88 / 3.32 & 3.70 / 6.94 & 9.33 / 6.08 \\
            4 & 84.95 / 70.34 & 7.90 / 10.22 & 11.01 / 16.42 & 14.37 / 9.26 \\
            8 & 60.13 / 53.82 & 30.94 / 32.67 & 38.09 / 39.35 & 11.95 / 9.36 \\
            \rowcolor{Goldenrod!25} 10 & 47.10 / 44.13 & 49.02 / 44.02 & 51.66 / 50.66 & 9.66 / 8.32 \\ 
            16 & 23.00 / 30.29 & 75.18 / 62.89 & 76.46 / 67.03 & 5.56 / 5.83 \\
        \bottomrule
        \end{tabular}}
    \end{table}

    ASR alone offers an incomplete and sometimes misleading view of adversarial attack effectiveness. It is \textbf{blind to beneficial flips}, failing to account for cases where perturbations actually improve predictions by turning an incorrect label into the correct one. This can lead to \textbf{inflated ASR scores under label noise}, where attacks appear successful by correcting existing errors. Furthermore, \textbf{ASR lacks granularity over the type of prediction change}, treating all flips as equivalent. This becomes especially problematic in cross-model comparisons, where identical ASR values can conceal substantial differences in behavior, particularly in how often attacks correct the clean model’s mistakes. \textbf{Among existing metrics, ASR captures only correct$\rightarrow$incorrect transitions, FR aggregates all changes without distinction, and only ACR isolates the critical incorrect$\rightarrow$correct transitions.}

    \paragraph{Summary of our approach.}
    We further summarize our findings and approach as follows. We empirically discover that existing generative works preserve the semantic integrity of the benign input image at the early intermediate blocks better than the later blocks, under the assumption that the downsampling blocks merely serve as feature extractors~\cite{zhang2022beyond}. To better structure adversarial noise generation in the intermediate blocks, we formulate our method to maintain at least coarse semantic structures in the \emph{early} intermediate blocks, thereby yielding much more perturbed features towards the end of the intermediate stage, which then results in enhanced transferability of the crafted AEs compared to the baseline. Through extensive cross-setting evaluations, we validate our approach of tuning the progression of adversarial noise generation in the generator's feature level as a compatible method to the existing generative attack framework~\cite{naseer2019cross, salzmann2021learning, zhang2022beyond, aich2022gama, yang2024facl, yang2024pdcl, nakka2025nat} without much overheads.

\section{Experimental Details}
\label{sec:additional_experiments}

    \subsection{Evaluation settings}
    
        We define three black-box evaluation scenarios that differ in the attacker’s knowledge of target data and models. In the cross-model setting, the adversary has access to the same data distribution used to train the unseen target models but must craft attacks using a substitute model rather than querying the targets directly. In the cross-domain setting, the attacker works solely with out-of-domain data and has no access to target-domain datasets (e.g., CUB-200-2011 \cite{cub}, Stanford Cars \cite{car}, FGVC Aircraft \cite{air}) or the ability to query target-domain models such as ResNet-50 \cite{res152}, SE-Net, or SE-ResNet101 \cite{senet}. Finally, in the cross-task setting, likewise, the adversary is completely agnostic to the target’s data, models, and even the task itself, representing the strict black-box challenge.


        
        \paragraph{Victim model specifications.}
        we selected a total of 22 different model architectures that span from CNNs to Vision Mamba variants, whose pre-trained model weights are available openly through TorchVision~\cite{marcel2010torchvision}, Timm~\cite{Wightman_PyTorch_Image_Models}, and the proprietary GitHub repositories. We list the sources in Table~\ref{tab:victim_models}.
        
    \subsection{Evaluated Models}
        
        \paragraph{Victim models.}
        For \emph{cross-model} evaluation, we employ ImageNet-1K ($224\times224$ resolution, 1,000 classes) pre-trained classification models of various architectures with their publicly available model weights. We source the pre-trained models from TorchVision~\cite{marcel2010torchvision} and Timm~\cite{Wightman_PyTorch_Image_Models} libraries. Compared to previous approaches~\cite{salzmann2021learning, yang2024facl, yang2024pdcl} demonstrating cross-model architecture transferability, we expand the evaluation to a wider scope of target model architectures for enhanced architecture-agnostic transferability.
        We tested our attack against a total of 21 different model architectures (11 CNN-based~\cite{res152,densenet,inc-v3,radosavovic2020regnet,tan2019mnasnet,iandola2016squeezenet,tan2104efficientnetv2}, 6 ViT-based~\cite{tu2022maxvit,liu2021swin,touvron2021deit,bao2021beit,cai2022efficientvit}, 2 Mixers~\cite{tolstikhin2021mlpmixer,trockman2022convmixer}, and 2 Vision Mamba-based~\cite{zhu2024visionmamba,hatamizadeh2025mambavision}) for cross-model evaluations. 
        
        For \emph{cross-domain} evaluation, we validate our attack against three different models (\ie ResNet50~\cite{res152}, SE-Net and SE-ResNet101~\cite{senet}) pre-trained on fine-grained datasets, CUB-200-2011~\cite{cub} (200 classes), Stanford Cars~\cite{car} (196 classes), FGVC Aircraft~\cite{air} (100 classes), of $448\times448$ resolution.
        %
        For \emph{cross-task}, we select a CNN-based and a ViT-based model for each task of semantic segmentation (SS) and object detection (OD), whose pre-trained weights are openly accessible as with the ImageNet-1K pre-trained weights. Specifically, we test against DeepLabV3+~\cite{chen2018deeplabv3plus}
        and SegFormer~\cite{xie2021segformer}
        for SS and Faster R-CNN~\cite{girshick2015fasterrcnn} on \texttt{detectron2}
        and DETR~\cite{carion2020detr}
        for OD.
        We validate on Cityscapes~\cite{cordts2016cityscapes} and COCO'17~\cite{lin2014mscoco} for SS and OD tasks, respectively.

        \begin{table*}[!h]
    \centering
    \setlength{\tabcolsep}{5pt}
    \renewcommand{\arraystretch}{1}
    \renewcommand{\aboverulesep}{1pt}
    \renewcommand{\belowrulesep}{1pt}
    \vspace{-3mm}
    \caption{Sources of victim models used to evaluate the attack performance, grouped by task.}
    \resizebox{\linewidth}{!}{
    \begin{tabular}{cll}
    \toprule
         \textbf{Task} & \textbf{Victim Model} & \textbf{Source} \\ 
    \midrule
         \multirow{22}{*}{\rotatebox{90}{Image Classification}}
         & (a) ResNet50~\cite{res152} & TorchVision~\cite{marcel2010torchvision} \\
         & (b) ResNet152~\cite{res152}& TorchVision~\cite{marcel2010torchvision} \\
         & (c) Dense121~\cite{densenet}& TorchVision~\cite{marcel2010torchvision} \\
         & (e) Dense169~\cite{densenet}& TorchVision~\cite{marcel2010torchvision} \\
         & (f) InceptionV3~\cite{inc-v3} & TorchVision~\cite{marcel2010torchvision} \\
         & (g) RegNetY~\cite{radosavovic2020regnet}& TorchVision~\cite{marcel2010torchvision} \\
         & (h) MNAsNet~\cite{tan2019mnasnet}& TorchVision~\cite{marcel2010torchvision} \\
         & (i) SqueezeNet~\cite{iandola2016squeezenet}& TorchVision~\cite{marcel2010torchvision} \\
         & (j) EfficientV2~\cite{tan2104efficientnetv2}& TorchVision~\cite{marcel2010torchvision} \\
         & (k) ConvNeXt-B~\cite{liu2022convnext}& TorchVision~\cite{marcel2010torchvision} \\
         & (l) ResNeXt~\cite{xie2017resnext}& TorchVision~\cite{marcel2010torchvision} \\
        & (m) ViT-B/16~\cite{dosovitskiy2020vit}& TorchVision~\cite{marcel2010torchvision} \\
        & (n) ViT-L/16~\cite{dosovitskiy2020vit}& TorchVision~\cite{marcel2010torchvision} \\
        & (o) Swin-B/16~\cite{liu2021swin}& TorchVision~\cite{marcel2010torchvision} \\
        & (p) DeiT-B~\cite{touvron2021deit} & Timm~\cite{Wightman_PyTorch_Image_Models} \\
        & (q) BEiT-B~\cite{bao2021beit}& Timm~\cite{Wightman_PyTorch_Image_Models} \\
        & (r) EfficientViT~\cite{cai2022efficientvit}& Timm~\cite{Wightman_PyTorch_Image_Models} \\
        & (s) MLP-Mixer-B~\cite{tolstikhin2021mlpmixer}& Timm~\cite{Wightman_PyTorch_Image_Models} \\
        & (t) ConvMixer-B~\cite{trockman2022convmixer}& Timm~\cite{Wightman_PyTorch_Image_Models} \\
        & (u) Vision Mamba-B~\cite{zhu2024visionmamba} & \textrm{https://github.com/hustvl/Vim} \\
        & (v) MambaVision-B~\cite{hatamizadeh2025mambavision} & \textrm{https://github.com/NVlabs/MambaVision} \\ 
    \midrule
        
         \multirow{2}{*}{Semantic Segmentation (SS)}
        & DeepLabV3+~\cite{chen2018deeplabv3plus} & \textrm{https://github.com/VainF/DeepLabV3Plus-Pytorch} \\
        & SegFormer~\cite{xie2021segformer}  & \textrm{https://github.com/NVlabs/SegFormer}\\
    \midrule
        \multirow{2}{*}{Object Detection (OD)}
        & Faster R-CNN~\cite{girshick2015fasterrcnn} & \textrm{https://github.com/facebookresearch/detectron2} \\
        & DETR~\cite{carion2020detr} & \textrm{https://github.com/facebookresearch/detr}\\

    \bottomrule
    \end{tabular}
    }
    \label{tab:victim_models}
\end{table*}
        
        \paragraph{Against robust models.}
        We tested our attack against robust models, \ie adversarially trained Inception-V3~\cite{kurakin2016bim}, ViT~\cite{dosovitskiy2020vit} and ConvNeXt~\cite{singh2023convstem} models, and robust input processing methods such as JPEG (75\%)~\cite{JPEG}, bit reduction (BDR; 4-bit)~\cite{xu2017bitreduction} and randomization (R\&P)~\cite{xie2017randomization}
        in Table 4.

    \subsection{Implementation details} 
        
        Throughout the experiments, we train the perturbation generator with $\epsilon=10$ using data from ImageNet-1K~\cite{imagenet} containing 1.2 M natural images of 224$\times$224 resolution, following~\cite{poursaeed2018generative, naseer2019cross, salzmann2021learning, zhang2022beyond, aich2022gama, yang2024facl, yang2024pdcl}, for one epoch using the Adam~\cite{adam} optimizer $\beta=(0.5, 0.99)$. We set the learning rate $lr=2e^{-4}$.
        We also use the mid-level layer feature at $k=16$ (Maxpooling.3) of VGG-16 surrogate for our baseline~\cite{zhang2022beyond}.
        We set $\lambda_{cons.}=0.7$ throughout our experiments for stable generator training at the feature level, and the EMA update parameter $\eta=0.999$ following~\cite{tarvainen2017meanteacher}.
        We selected $L_{early}=\{1,2\}$ for matching the first and second intermediate residual blocks within the generator.
        %
        We compare our attacks against the state-of-the-art baselines that rely on the same ResNet-based generator to craft adversarial examples, \ie 
        CDA~\cite{naseer2019cross}, LTP~\cite{salzmann2021learning}, BIA~\cite{zhang2022beyond}, GAMA~\cite{aich2022gama}, FACL-Attack~\cite{yang2024facl}, and PDCL-Attack~\cite{yang2024pdcl}.
        

        \paragraph{Dataset statistics.}
        We describe the statistics of the datasets used for training and evaluation in Table~\ref{tab:datasets}. Note that we do not use the training sets from CUB-200-2011~\cite{cub}, Stanford Cars~\cite{car}, or FGVC Aircraft~\cite{air} for the strict black-box cross-domain.
        
        \begin{table*}[!h]
            \vspace{-1mm}
            \centering
            \captionof{table}{Training and evaluation dataset statistics.}
            \vspace{-2mm}
            \setlength{\tabcolsep}{1pt}
            \renewcommand{\arraystretch}{1}
            \renewcommand{\aboverulesep}{3pt}
            \renewcommand{\belowrulesep}{3pt}
            \resizebox{.7\linewidth}{!}{
            \begin{tabular}{ccccc}
                \toprule
                    Dataset & \shortstack[c]{ImageNet-1K\\~\cite{imagenet}} & \shortstack[c]{CUB-200-2011\\~\cite{cub}} & \shortstack[c]{Stanford Cars\\~\cite{car}} & \shortstack[c]{FGVC Aircraft\\~\cite{air}} \\
                \midrule
                    Train & 1.2 M & \textcolor{gray!50}{5,994 (Not Used)} & \textcolor{gray!50}{8,144 (Not Used)} & \textcolor{gray!50}{6,667 (Not Used)} \\ 
                    Val. & 50,000 & 5,794 & 8,041 & 3,333 \\
                    \# Classes & 1,000 & 200 & 196 & 100 \\
                    Resolution & 224$\times$224 & 448$\times$448 & 448$\times$448 & 448$\times$448 \\
                \bottomrule
            \end{tabular}
            }
            \label{tab:datasets}
            \vspace{-2mm}
        \end{table*}
        
        

        
        \paragraph{Computational costs.}
        Since our approach only involves computational overheads during the training of the perturbation generator, there is \textit{no inference time overhead}. During training, we describe in the table on the right that the time for a single forward pass with a batch size of 1 incurs an additional +12.18 (ms) compared to the baseline~\cite{zhang2022beyond} time of 44.72 (ms), and an additional +28.31 (MB) in memory compared to that for the baseline of 1,404.13 (MB), which are averaged over 1,000 iterations (See 
        Tables~\ref{tab:supp_compute1}, \ref{tab:supp_compute2}
        ).

\begin{table}[t]
  \centering
  \setlength{\tabcolsep}{10pt}
  \renewcommand{\arraystretch}{1}
  \caption{
  Total training compute for the baseline and w/ Ours.}
  \label{tab:supp_compute1}
  \vspace{-3mm}
  \resizebox{\linewidth}{!}{
  \begin{tabular}{lcccc}
    \toprule
    \textbf{Method} & \textbf{Train time (hh:mm)} & \textbf{Peak memory (MB)} & \textbf{GPU type} & \textbf{Train batch size} \\
    \midrule
    Baseline & 5:00 & 1{,}384.62 & \multirow{2}{*}{NVIDIA RTX A6000 (1$\times$)} & \multirow{2}{*}{48} \\
    w/ Ours & 5:40 & 1{,}442.23 \\
    \bottomrule
  \end{tabular}}
\end{table}

\begin{table}[t]
  \centering
  \setlength{\tabcolsep}{6pt}
  \renewcommand{\arraystretch}{1}
  \caption{
  Average per-train-step compute over 10k steps with a batch size of 1.}
  \label{tab:supp_compute2}
  \vspace{-3mm}
  \resizebox{\linewidth}{!}{
  \begin{tabular}{lcccccc}
    \toprule
    \textbf{Method} &
    \textbf{Student fwd} (ms) &
    \textbf{Teacher fwd} (ms) &
    \textbf{Backward} (ms) &
    \textbf{Total} (ms) &
    \textbf{Backward cost} (GFLOPs) &
    \textbf{Backward CUDA time} (ms) \\
    \midrule
    Baseline & 7.1 & --  & 25.6 & 32.7 & 0.0012 & 19.714 \\
    w/ Ours & 6.8 & 3.9 & 28.4 & 39.1 & 0.0044 & 20.192 \\
    \bottomrule
  \end{tabular}}
\end{table}
        These slight increases owe to added computations for the forward (no backward) pass overhead on the teacher generator, $\mathcal{G}_{\theta'}(\cdot)$, in addition to the student,$\mathcal{G}_{\theta}(\cdot)$, and are significantly minor relative to the baseline costs.
        We note that the training was performed on a single NVIDIA RTX A6000 GPU.

\subsection{Additional quantitative results}
\label{sec:supp_additional_quantitative}
    \paragraph{Results with different surrogate models.}

    While we performed experiments against the VGG-16 surrogate model for fair comparison with previous works, we also provide, in Tables~\ref{tab:supp_different_surrogate_model} and \ref{tab:supp_different_surrogate_domain}, our improvements when trained against other surrogate models such as VGG-19, ResNet-152, and DenseNet-169 as practiced in \cite{zhang2022beyond, yang2024facl, yang2024pdcl}. Ours added to the baseline trained against all three surrogate models across models and domains, except for cross-model against Dense169, consistently enhances the attack transferability. As with our results against VGG-16 in Tables 1 and 2, our method effectively boosts the attack capacity regardless of the type of surrogate model used for training the generator. We believe the slight increase in cross-model average when using DenseNet-169 as the surrogate model is driven mainly by pronounced gains on a few architectures (\eg MNASNet and DeiT) while most other models also benefited, albeit to a lesser extent. In the cross-domain evaluation, adversarial examples generated with DenseNet-169 consistently deliver a substantial average performance boost, underscoring its effectiveness across differing data distributions.

\begin{figure*}[!t]
        \centering
        \includegraphics[width=\linewidth]{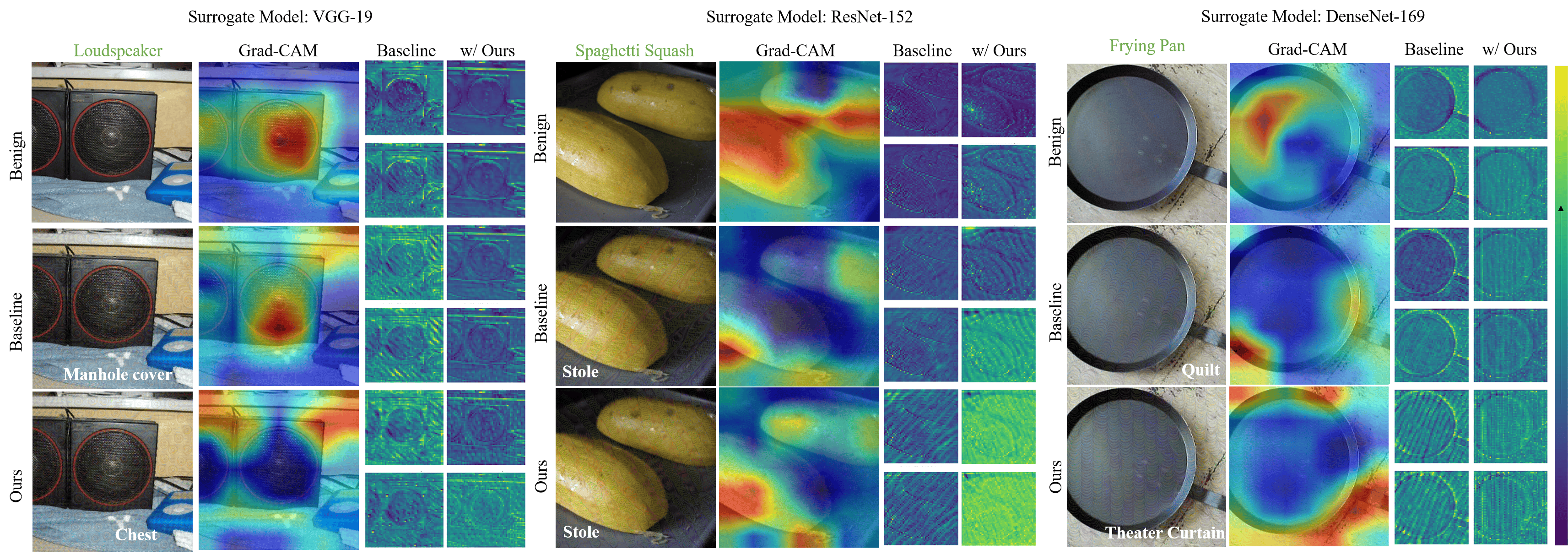}
        \caption{Comparison of the crafted AEs, Grad-CAMs, intermediate feature activation maps from the baseline, and with ours (columns 1--4, respectively). We present the qualitative results from training against other surrogate models:  VGG-19 (Left), ResNet-152 (Center), and DenseNet-169 (Right) as commonly compared in existing generative attacks~\cite{zhang2022beyond, aich2022gama, yang2024facl, yang2024pdcl}. The correct label and attacked prediction results are marked in \textcolor{ForestGreen}{\textbf{green}} and white, respectively.}
        \label{fig:supp_different_surrogate}
    \end{figure*}

    \begin{wraptable}{r}{0.65\textwidth}
        \vspace{-4mm}
        \centering
        \captionof{table}{Comparison of cross-task attack strength with ours added to each baseline. Ours further enhances the transferability consistently across semantic segmentation and object detection tasks. Boldface means better results.}

    \setlength{\tabcolsep}{1pt}
    \renewcommand{\arraystretch}{1}
    \renewcommand{\aboverulesep}{2pt}
    \renewcommand{\belowrulesep}{2pt}
    \resizebox{\linewidth}{!}{
    \begin{tabular}{cccccccc}
    \toprule
        \texttt{Cross-task} & \multicolumn{5}{c}{Task} \\ \cmidrule{2-8}
        & \multicolumn{2}{c}{Semantic Segmentation (mIoU $\downarrow$)}  & \multirow{2}{*}{Avg.} && \multicolumn{2}{c}{Object Detection (mAP50 $\downarrow$)} & \multirow{2}{*}{Avg.}   \\ \cmidrule{2-3} \cmidrule{6-7}
        & DeepLabV3+
        & SegFormer
        & && Faster R-CNN
        & DETR
        & \\ 
        
    \midrule
        \rowcolor{Goldenrod!25} Benign & 76.21 & 71.89 & 74.05 && 61.01 & 62.36 & 61.69 \\
    
    \midrule    
         CDA~\cite{naseer2019cross}                        & 25.63 & \textcolor{ForestGreen!75}{\textbf{20.16}} & 22.90 && 32.78 & 26.29 & 29.54 \\
          w/ Ours & \textcolor{ForestGreen!75}{\textbf{25.16}} & 20.26 & \textcolor{ForestGreen!75}{\textbf{22.71}} && \textcolor{ForestGreen!75}{\textbf{31.98}} & \textcolor{ForestGreen!75}{\textbf{25.68}} & \textcolor{ForestGreen!75}{\textbf{28.83}}  \\
    \midrule
         LTP~\cite{salzmann2021learning}                        & 23.71 & 26.97 & 25.34  && 29.39 & 22.41 & 25.90 \\
          w/ Ours & \textcolor{ForestGreen!75}{\textbf{22.27}} & \textcolor{ForestGreen!75}{\textbf{26.68}} & \textcolor{ForestGreen!75}{\textbf{24.48}} && \textcolor{ForestGreen!75}{\textbf{26.85}} & \textcolor{ForestGreen!75}{\textbf{22.18}} & \textcolor{ForestGreen!75}{\textbf{24.52}} \\
    \midrule
         BIA~\cite{zhang2022beyond}                        & 23.89 & 25.60 & 24.75 && 28.43 & 21.01 & 24.72 \\
          w/ Ours & \textcolor{ForestGreen!75}{\textbf{22.05}} & \textcolor{ForestGreen!75}{\textbf{24.75}} & \textcolor{ForestGreen!75}{\textbf{23.40}} && \textcolor{ForestGreen!75}{\textbf{28.34}} & \textcolor{ForestGreen!75}{\textbf{20.70}}& \textcolor{ForestGreen!75}{\textbf{24.52}}  \\
    \midrule
         GAMA~\cite{aich2022gama}                       & 24.10 & 27.53 & 25.82 && 28.01 & \textcolor{ForestGreen!75}{\textbf{20.71}} & 24.36  \\
          w/ Ours & \textcolor{ForestGreen!75}{\textbf{23.67}} & \textcolor{ForestGreen!75}{\textbf{25.59}} & \textcolor{ForestGreen!75}{\textbf{24.63}} && \textcolor{ForestGreen!75}{\textbf{27.60}} & 20.79 & \textcolor{ForestGreen!75}{\textbf{24.20}} \\
    \midrule
         FACL~\cite{yang2024facl}                       & 23.75 & 26.40 & 25.08 && 27.94 & 20.91 & 24.43 \\
          w/ Ours & \textcolor{ForestGreen!75}{\textbf{23.38}} & \textcolor{ForestGreen!75}{\textbf{25.01}} & \textcolor{ForestGreen!75}{\textbf{24.20}} && \textcolor{ForestGreen!75}{\textbf{27.64}} & \textcolor{ForestGreen!75}{\textbf{20.29}} & \textcolor{ForestGreen!75}{\textbf{23.97}}   \\
    \midrule
         PDCL~\cite{yang2024pdcl}                       & 24.42 & 26.05 & 25.24 && 28.48 & 21.38 & 24.93 \\
          w/ Ours & \textcolor{ForestGreen!75}{\textbf{22.51}} & \textcolor{ForestGreen!75}{\textbf{25.88}} & \textcolor{ForestGreen!75}{\textbf{24.20}} && \textcolor{ForestGreen!75}{\textbf{27.66}} & \textcolor{ForestGreen!75}{\textbf{20.73}} & \textcolor{ForestGreen!75}{\textbf{24.20}} \\
         
    \bottomrule
    \end{tabular}
    }
    \label{tab:supp_cross_task}
        \vspace{-4mm}
    \end{wraptable}
    In qualitative comparisons across different surrogate models in Fig.~\ref{fig:supp_different_surrogate}, our Grad-CAM~\cite{selvaraju2017gradcam} visualizations reveal activation patterns that depart significantly from the baseline. Rather than highlighting only the main object region, our method further amplifies those top responses and draws out additional high-sensitivity areas that the baseline misses. When we step through the stages of adversarial noise generation, we see that our approach consistently places perturbations along object edges and contours. By focusing noise on these shared, model-agnostic features instead of scattering it elsewhere, our method not only seeks to align the generated adversarial noise with the most semantically meaningful regions but also achieves stronger transferability.

\begin{table*}[!t] \centering
\caption{\textbf{Quantitative cross-model transferability results}. We report the average improvements ($\Delta$ \%p) of our method relative to each baseline, with better results marked in a darker color. For VGG-19, Res152, Dense169 surrogate, (a-d) correspond to \{Res50, Res152, Dense121, Dense169\}, \{VGG16, VGG19, Dense121, Dense169\}, and \{VGG16,VGG19,Res50,Res152\}, respectively, as black-box models.}
\setlength{\tabcolsep}{1pt}
\renewcommand{\arraystretch}{1.1}
\renewcommand{\aboverulesep}{5pt}
\renewcommand{\belowrulesep}{5pt}
\label{tab:supp_different_surrogate_model}
\vspace{-0.2cm}
{
\fontsize{18}{18}\selectfont 
\resizebox{\linewidth}{!}{%
\begin{tabular}{ccccccccccccccccccccccccc}
\toprule
\multicolumn{2}{c}{\texttt{Cross-model}} & \multicolumn{11}{c}{\bf CNN} & \multicolumn{6}{c}{\bf Transformer} & \multicolumn{2}{c}{\bf Mixer} & \multicolumn{2}{c}{\bf Mamba} & \\
\cmidrule[1pt](lr){3-13}
\cmidrule[1pt](lr){14-19}
\cmidrule[1pt](lr){20-21}
\cmidrule[1pt](lr){22-23}
\textbf{Method} & \textbf{Metric} & (a) & (b) & (c) & (d) & (e) & (f) & (g) & (h) & (i) & (j) & (k) & (l) & (m) & (n) & (o) & (p) & (q) & (r) & (s) & (t) & (v) & \multirow{1}{*}{Avg.}\\
\midrule
\rowcolor{Goldenrod!20} Benign & Acc. (\%) $\downarrow$
& \text{74.60} & \text{77.33} & \text{74.22} & \text{75.74} & \text{76.19} & \text{77.95} & \text{66.50} & \text{55.91} & \text{79.12} & \text{81.49} & \text{75.42} 
& 80.67	& 79.28 &	81.19	& 80.48	& 79.10	& 57.91	
& 69.90	& 66.53
& 66.53 &	73.21
& 73.77
\\
\midrule
\rowcolor{gray!25} \multicolumn{24}{l}{Surrogate model: VGG19} \\
\hline
\multirow{4}{*}{\shortstack{BIA\\w/ Ours}} 
& Acc. ($\Delta$\%p) $\downarrow$
& \textcolor{black}{-1.49} & \textcolor{gray!75}{+0.88} & \textcolor{black}{-1.70} & \textcolor{black}{-3.49} & \textcolor{black}{-4.04} & \textcolor{black}{-1.07} & \textcolor{gray!75}{+1.06} & \textcolor{black}{-2.31} & \textcolor{gray!75}{+0.46} & \textcolor{black}{-4.40} & \textcolor{gray!75}{+0.39} & \textcolor{black}{-0.23} & \textcolor{black}{-0.17} & \textcolor{black}{-1.14} & \textcolor{gray!75}{+0.04} & \textcolor{black}{-0.40} & \textcolor{gray!75}{+0.22} & \textcolor{black}{-0.46} & \textcolor{gray!75}{+0.11} & \textcolor{black}{-0.12} & \textcolor{black}{-0.02} & \textcolor{ForestGreen!75}{\textbf{-0.85}} \\
& ASR ($\Delta$\%p) $\uparrow$
& \textcolor{black}{+1.93} & \textcolor{gray!75}{-1.12} & \textcolor{black}{+2.17} & \textcolor{black}{+4.43} & \textcolor{black}{+5.18} & \textcolor{black}{+1.34} & \textcolor{gray!75}{-1.43} & \textcolor{black}{+3.98} & \textcolor{gray!75}{-0.65} & \textcolor{black}{+5.18} & \textcolor{gray!75}{-0.44} & \textcolor{black}{+0.35} & \textcolor{black}{+0.16} & \textcolor{black}{+1.19} & \textcolor{gray!75}{-0.06} & \textcolor{black}{+0.48} & \textcolor{gray!75}{-0.35} & \textcolor{black}{+0.65} & \textcolor{gray!75}{-0.15} & \textcolor{black}{+0.17} & \textcolor{black}{+0.11} & \textcolor{ForestGreen!75}{\textbf{+1.10}} \\
& FR ($\Delta$\%p) $\uparrow$
& \textcolor{black}{+1.72} & \textcolor{gray!75}{-0.89} & \textcolor{black}{+1.97} & \textcolor{black}{+3.95} & \textcolor{black}{+4.68} & \textcolor{black}{+1.24} & \textcolor{gray!75}{-1.19} & \textcolor{black}{+2.89} & \textcolor{gray!75}{-0.55} & \textcolor{black}{+4.77} & \textcolor{gray!75}{-0.37} & \textcolor{black}{+0.51} & \textcolor{black}{+0.07} & \textcolor{black}{+1.26} & \textcolor{gray!75}{-0.14} & \textcolor{black}{+0.27} & \textcolor{gray!75}{-0.56} & \textcolor{black}{+0.69} & \textcolor{gray!75}{-0.27} & \textcolor{black}{+0.13} & \textcolor{black}{+0.10} & \textcolor{ForestGreen!75}{\textbf{+0.97}} \\
& ACR ($\Delta$\%p) $\downarrow$
& \textcolor{black}{-0.23} & \textcolor{gray!75}{+0.11} & \textcolor{black}{-0.35} & \textcolor{black}{-0.56} & \textcolor{black}{-0.41} & \textcolor{black}{-0.13} & \textcolor{gray!75}{+0.34} & \textcolor{black}{-0.18} & \textcolor{black}{-0.20} & \textcolor{black}{-0.82} & \textcolor{gray!75}{+0.24} & \textcolor{gray!75}{+0.28} & \textcolor{black}{-0.19} & \textcolor{black}{-0.92} & \textcolor{black}{-0.02} & \textcolor{black}{-0.10} & \textcolor{gray!75}{+0.02} & \textcolor{black}{-0.04} & \textcolor{gray!75}{+0.03} & \textcolor{black}{-0.02} & \textcolor{gray!75}{+0.23} & \textcolor{ForestGreen!75}{\textbf{-0.14}} \\
\midrule
\textbf{Method} & \textbf{Metric} & (a) & (b) & (c) & (d) & (e) & (f) & (g) & (h) & (i) & (j) & (k) & (l) & (m) & (n) & (o) & (p) & (q) & (r) & (s) & (t) & (v) & \multirow{1}{*}{Avg.}\\
\midrule
\rowcolor{Goldenrod!20} Benign & Acc. (\%) $\downarrow$
& \text{70.15} & \text{70.95} & \text{74.22} & \text{75.74} & \text{76.19} & \text{77.95} & \text{66.50} & \text{55.91} & \text{79.12} & \text{81.49} & \text{75.42} 
& 80.67	& 79.28 &	81.19	& 80.48	& 79.10	& 57.91	
& 69.90	& 66.53
& 66.53 &	73.21
& 73.26

\\
\midrule
\rowcolor{gray!25} \multicolumn{24}{l}{Surrogate model: Res152} \\
\hline
\multirow{4}{*}{\shortstack{BIA\\w/ Ours}} 
& Acc. ($\Delta$\%p) $\downarrow$
& \textcolor{gray!75}{+0.38} & \textcolor{black}{-0.87} & \textcolor{black}{-2.96} & \textcolor{black}{-0.38} & \textcolor{black}{-5.13} & \textcolor{black}{-7.36} & \textcolor{black}{-5.35} & \textcolor{black}{-0.71} & \textcolor{black}{-3.87} & \textcolor{black}{-6.17} & \textcolor{black}{-3.75} & \textcolor{black}{-0.34} & \textcolor{black}{-0.57} & \textcolor{black}{-0.37} & \textcolor{gray!75}{+0.00} & \textcolor{black}{-0.38} & \textcolor{black}{-0.59} & \textcolor{black}{-3.77} & \textcolor{black}{-2.27} & \textcolor{black}{-0.23} & \textcolor{black}{-0.70} & \textcolor{ForestGreen!75}{\textbf{-2.16}} \\
& ASR ($\Delta$\%p) $\uparrow$
& \textcolor{gray!75}{-0.52} & \textcolor{black}{+1.22} & \textcolor{black}{+3.76} & \textcolor{black}{+0.39} & \textcolor{black}{+6.36} & \textcolor{black}{+9.09} & \textcolor{black}{+7.37} & \textcolor{black}{+1.32} & \textcolor{black}{+4.75} & \textcolor{black}{+7.16} & \textcolor{black}{+4.70} & \textcolor{black}{+0.41} & \textcolor{black}{+0.83} & \textcolor{black}{+0.51} & \textcolor{black}{+0.08} & \textcolor{gray!75}{-0.97} & \textcolor{black}{+0.47} & \textcolor{black}{+6.42} & \textcolor{black}{+2.02} & \textcolor{black}{+1.66} & \textcolor{black}{+9.36} & \textcolor{ForestGreen!75}{\textbf{+2.74}} \\
& FR ($\Delta$\%p) $\uparrow$
& \textcolor{gray!75}{-0.49} & \textcolor{black}{+0.95} & \textcolor{black}{+3.23} & \textcolor{black}{+0.27} & \textcolor{black}{+5.70} & \textcolor{black}{+8.14} & \textcolor{black}{+6.29} & \textcolor{black}{+0.97} & \textcolor{black}{+4.37} & \textcolor{gray!75}{-1.23} & \textcolor{black}{+7.24} & \textcolor{gray!75}{-1.22} & \textcolor{gray!75}{-0.45} & \textcolor{gray!75}{-0.94} & \textcolor{black}{+0.30} & \textcolor{gray!75}{-0.86} & \textcolor{black}{+0.57} & \textcolor{black}{+5.38} & \textcolor{black}{+1.60} & \textcolor{black}{+1.42} & \textcolor{black}{+8.78} & \textcolor{ForestGreen!75}{\textbf{+3.22}} \\
& ACR ($\Delta$\%p) $\downarrow$
& \textcolor{black}{-0.46} & \textcolor{black}{-0.74} & \textcolor{black}{-0.73} & \textcolor{black}{-1.65} & \textcolor{black}{-2.64} & \textcolor{black}{-1.19} & \textcolor{black}{-0.31} & \textcolor{black}{-0.78} & \textcolor{black}{-1.47} & \textcolor{black}{-0.40} & \textcolor{black}{-1.75} & \textcolor{black}{-0.10} & \textcolor{gray!75}{+0.07} & \textcolor{gray!75}{+0.67} & \textcolor{gray!75}{+0.53} & \textcolor{black}{-0.19} & \textcolor{gray!75}{+0.28} & \textcolor{black}{-0.86} & \textcolor{black}{-0.45} & \textcolor{black}{-0.25} & \textcolor{black}{-2.93} & \textcolor{ForestGreen!75}{\textbf{-0.74}} \\
\midrule
\textbf{Method} & \textbf{Metric} & (a) & (b) & (c) & (d) & (e) & (f) & (g) & (h) & (i) & (j) & (k) & (l) & (m) & (n) & (o) & (p) & (q) & (r) & (s) & (t) & (v) & \multirow{1}{*}{Avg.}\\
\midrule
\rowcolor{Goldenrod!20} Benign & Acc. (\%) $\downarrow$
& \text{70.15} & \text{70.95} & \text{74.60} & \text{77.33} & \text{76.19} & \text{77.95} & \text{66.50} & \text{55.91} & \text{79.12} & \text{81.49} & \text{75.42} 
& 80.67	& 79.28 &	81.19	& 80.48	& 79.10	& 57.91	
& 69.90	& 66.53
& 66.53 &	73.21
& 73.35
\\
\midrule
\rowcolor{gray!25} \multicolumn{24}{l}{Surrogate model: Dense169} \\
\hline
\multirow{4}{*}{\shortstack{BIA\\w/ Ours}} 
& Acc. ($\Delta$\%p) $\downarrow$
& \textcolor{black}{-1.63} & \textcolor{black}{-3.00} & \textcolor{black}{-2.25} & \textcolor{black}{-7.65} & \textcolor{black}{-12.99} & \textcolor{black}{-4.66} & \textcolor{black}{-0.13} & \textcolor{black}{-3.01} & \textcolor{black}{-7.68} & \textcolor{gray!75}{+1.19} & \textcolor{black}{-6.71} & \textcolor{gray!75}{+1.00} & \textcolor{gray!75}{+0.36} & \textcolor{gray!75}{+0.91} & \textcolor{black}{-0.10} & \textcolor{gray!75}{+0.74} & \textcolor{black}{-0.16} & \textcolor{black}{-4.74} & \textcolor{black}{-1.50} & \textcolor{black}{-1.19} & \textcolor{black}{-7.63} & \textcolor{ForestGreen!75}{\textbf{-2.90}} \\
& ASR ($\Delta$\%p) $\uparrow$
& \textcolor{black}{+2.13} & \textcolor{black}{+3.92} & \textcolor{black}{+2.77} & \textcolor{black}{+9.41} & \textcolor{black}{+16.23} & \textcolor{black}{+5.64} & \textcolor{black}{+0.04} & \textcolor{black}{+4.77} & \textcolor{black}{+9.32} & \textcolor{gray!75}{-1.55} & \textcolor{black}{+8.22} & \textcolor{gray!75}{-1.26} & \textcolor{gray!75}{-0.47} & \textcolor{gray!75}{-0.96} & \textcolor{black}{+0.24} & \textcolor{gray!75}{-0.97} & \textcolor{black}{+0.47} & \textcolor{black}{+6.42} & \textcolor{black}{+2.02} & \textcolor{black}{+1.66} & \textcolor{black}{+9.36} & \textcolor{ForestGreen!75}{\textbf{+3.69}} \\
& FR ($\Delta$\%p) $\uparrow$
& \textcolor{black}{+1.79} & \textcolor{black}{+3.27} & \textcolor{black}{+2.41} & \textcolor{black}{+8.48} & \textcolor{black}{+13.97} & \textcolor{black}{+4.98} & \textcolor{black}{+0.14} & \textcolor{black}{+3.42} & \textcolor{black}{+8.66} & \textcolor{gray!75}{-1.23} & \textcolor{black}{+7.24} & \textcolor{gray!75}{-1.22} & \textcolor{gray!75}{-0.45} & \textcolor{gray!75}{-0.94} & \textcolor{black}{+0.30} & \textcolor{gray!75}{-0.86} & \textcolor{black}{+0.57} & \textcolor{black}{+5.38} & \textcolor{black}{+1.60} & \textcolor{black}{+1.42} & \textcolor{black}{+8.78} & \textcolor{ForestGreen!75}{\textbf{+3.22}} \\
& ACR ($\Delta$\%p) $\downarrow$
& \textcolor{black}{-0.46} & \textcolor{black}{-0.74} & \textcolor{black}{-0.73} & \textcolor{black}{-1.65} & \textcolor{black}{-2.64} & \textcolor{black}{-1.19} & \textcolor{black}{-0.31} & \textcolor{black}{-0.78} & \textcolor{black}{-1.47} & \textcolor{black}{-0.40} & \textcolor{black}{-1.75} & \textcolor{black}{-0.10} & \textcolor{gray!75}{+0.07} & \textcolor{gray!75}{+0.67} & \textcolor{gray!75}{+0.53} & \textcolor{black}{-0.19} & \textcolor{gray!75}{+0.28} & \textcolor{black}{-0.86} & \textcolor{black}{-0.45} & \textcolor{black}{-0.25} & \textcolor{black}{-2.93} & \textcolor{ForestGreen!75}{\textbf{-0.74}} \\

\bottomrule
\end{tabular}%
}}
\end{table*}

\begin{table*}[!h]
\centering
\setlength{\tabcolsep}{2pt}
\renewcommand{\arraystretch}{1}
\renewcommand{\aboverulesep}{4pt}
\renewcommand{\belowrulesep}{4pt}
\caption{\textbf{Additional quantitative cross-domain transferability results}. We report the average improvement margins of our method added to each baseline, averaged over three models for each domain using different surrogate models (VGG-19, Res152, Dense169). We report the improvements ($\Delta$ \%p) with ours relative to the baseline~\cite{zhang2022beyond}. Better averaged results are marked in \textbf{boldface}.}
\label{tab:supp_different_surrogate_domain}
\vspace{-0.3cm}
\fontsize{24}{24}\selectfont
\resizebox{\linewidth}{!}{
\begin{tabular}{ccccccccccccccccc|ccccccc}
\toprule
      \multirow{1}{*}{\texttt{Cross-domain}} && \multicolumn{4}{c}{\textbf{CUB-200-2011}
      } 
      && \multicolumn{4}{c}{\textbf{Stanford Cars}
      } 
      && \multicolumn{4}{c}{\textbf{FGVC Aircraft}
      } & \multirow{2}{*}{\shortstack{Avg.\\Acc. (\%)}} 
      & \multicolumn{2}{c}{\textbf{SemSeg (SS)}} 
    & \multirow{2}{*}{\shortstack{Avg.\\mIoU $\downarrow$}} 
    & \multicolumn{2}{c}{\textbf{ObjDet (OD)}} 
    & \multirow{2}{*}{\shortstack{Avg.\\mAP50 $\downarrow$}} 
    \\  \cmidrule{3-6}\cmidrule{8-11}\cmidrule{13-16} \cmidrule{18-19} \cmidrule{21-22}
     
    \textbf{Method} 
      && Acc $\downarrow$ & ASR $\uparrow$ & FR $\uparrow$ & ACR $\downarrow$
      && Acc $\downarrow$ & ASR $\uparrow$ & FR $\uparrow$ & ACR $\downarrow$
      && Acc $\downarrow$ & ASR $\uparrow$ & FR $\uparrow$ & ACR $\downarrow$ 
          &
    & DeepLabV3+
    & SegFormer
    &
    & Faster R-CNN
    & DETR
    \\
\midrule
    \rowcolor{Goldenrod!20} Benign  
      && 86.91 & \textcolor{gray}{\tiny{N/A}} & \textcolor{gray}{\tiny{N/A}} & \textcolor{gray}{\tiny{N/A}} 
      && 93.56 & \textcolor{gray}{\tiny{N/A}} & \textcolor{gray}{\tiny{N/A}} & \textcolor{gray}{\tiny{N/A}}
      && 92.07 & \textcolor{gray}{\tiny{N/A}} & \textcolor{gray}{\tiny{N/A}} & \textcolor{gray}{\tiny{N/A}} & 90.85 
    & 76.21 & 71.89 & 74.05
    & 61.01 & 62.36 & 61.69 \\

\midrule
\rowcolor{gray!25} \multicolumn{23}{l}{Surrogate model: VGG19} \\
\midrule
    BIA~\cite{zhang2022beyond} (\%)
      && 52.47 & 41.08 & 45.63 & 9.77  
      && 71.09 & 25.16 & 27.32 & 17.15    
      && 52.28 & 44.14 & 46.96 & 11.03 & 58.61 
      & 28.11 & 25.86 & 26.99 & 28.85 & 21.77 & 25.31 \\
    w/ Ours  ($\Delta$\%p)
      && \textcolor{ForestGreen!75}{\textbf{-10.05}} & \textcolor{ForestGreen!75}{\textbf{+11.20}} & \textcolor{ForestGreen!75}{\textbf{+10.39}} & \textcolor{ForestGreen!75}{\textbf{-2.26}} 
      && \textcolor{ForestGreen!75}{\textbf{-11.48}} & \textcolor{ForestGreen!75}{\textbf{+12.04}} & \textcolor{ForestGreen!75}{\textbf{+11.73}} & \textcolor{ForestGreen!75}{\textbf{-2.91}} 
      && \textcolor{ForestGreen!75}{\textbf{-10.87}} & \textcolor{ForestGreen!75}{\textbf{+11.56}} & \textcolor{ForestGreen!75}{\textbf{+11.04}} & \textcolor{ForestGreen!75}{\textbf{-2.79}} & \textcolor{ForestGreen!75}{\textbf{-10.80}} 
      & \textcolor{ForestGreen!75}{\textbf{-2.59}} & \textcolor{ForestGreen!75}{\textbf{-0.69}} & \textcolor{ForestGreen!75}{\textbf{-1.64}} & +0.46 & +0.12 & +0.29 \\

\midrule
\rowcolor{gray!25} \multicolumn{23}{l}{Surrogate model: Res152} \\
\midrule
    BIA~\cite{zhang2022beyond} (\%)
      && 49.52 & 44.51 & 48.53 & 9.96 
      && 50.71 & 46.60 & 48.44 & 12.81
      && 40.43 & 56.83 & 59.14 & 9.01 & 46.89
       & 32.34 & 31.63 & 31.98 & 33.02 & 26.02 & 29.52 \\ 
    w/ Ours  ($\Delta$\%p)
      && \textcolor{ForestGreen!75}{\textbf{-6.27}} & \textcolor{ForestGreen!75}{\textbf{+7.07}} & \textcolor{ForestGreen!75}{\textbf{+6.62}} & \textcolor{ForestGreen!75}{\textbf{-0.82}}
      && +0.09 & -0.15 & -0.20 & \textcolor{ForestGreen!75}{\textbf{-0.43}}
      && \textcolor{ForestGreen!75}{\textbf{-7.17}} & \textcolor{ForestGreen!75}{\textbf{+7.72}} & \textcolor{ForestGreen!75}{\textbf{+7.07}} & \textcolor{ForestGreen!75}{\textbf{-0.87}} & \textcolor{ForestGreen!75}{\textbf{-4.45}} 
       & +2.44 & +2.55 & +2.50 & \textcolor{ForestGreen!75}{\textbf{-0.39}} & +0.99 & +0.30 \\

\midrule
\rowcolor{gray!25} \multicolumn{23}{l}{Surrogate model: Dense169} \\
\midrule
    BIA~\cite{zhang2022beyond} (\%)
      && 30.01 & 66.46 & 68.97 & 6.73
      && 34.08 & 64.13 & 65.34 & 9.29 
      && 23.23 & 75..24 & 5.74 & 76.44 & 29.11 
       & 27.70 & 30.31 & 29.01 & 31.53 & 26.89 & 29.21\\ 
    w/ Ours  ($\Delta$\%p)
      && \textcolor{ForestGreen!75}{\textbf{-3.68}} & \textcolor{ForestGreen!75}{\textbf{+4.04}} & \textcolor{ForestGreen!75}{\textbf{+3.92}} & \textcolor{ForestGreen!75}{\textbf{-1.20}}
      && \textcolor{ForestGreen!75}{\textbf{-8.54}} & \textcolor{ForestGreen!75}{\textbf{+8.99}} & \textcolor{ForestGreen!75}{\textbf{+8.66}} & \textcolor{ForestGreen!75}{\textbf{-2.32}}
      && \textcolor{ForestGreen!75}{\textbf{-11.60}} & \textcolor{ForestGreen!75}{\textbf{+12.32}} & \textcolor{ForestGreen!75}{\textbf{+11.62}} & \textcolor{ForestGreen!75}{\textbf{-3.36}} & \textcolor{ForestGreen!75}{\textbf{-7.94}}
       & +0.08 & \textcolor{ForestGreen!75}{\textbf{-2.74}} & \textcolor{ForestGreen!75}{\textbf{-1.33}} & \textcolor{ForestGreen!75}{\textbf{-0.37}} & \textcolor{ForestGreen!75}{\textbf{-0.76}} & \textcolor{ForestGreen!75}{\textbf{-0.56}} \\

\bottomrule
\end{tabular}
}
\end{table*}
    
\paragraph{Cross-task evaluations.}
\begin{wraptable}{r}{0.45\textwidth}
    \vspace{-5mm}
    \centering
    \includegraphics[width=\linewidth]{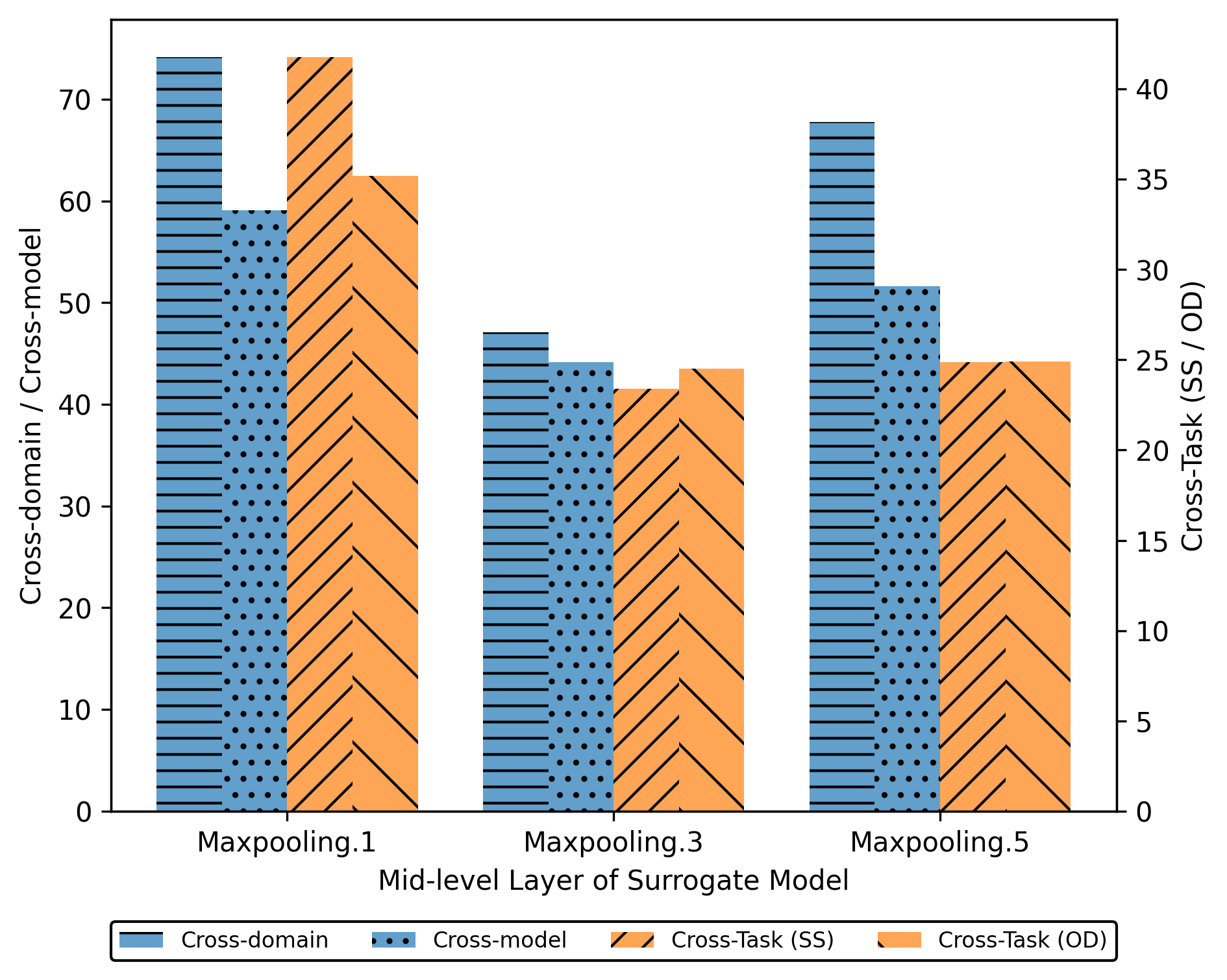}
    \vspace{-7mm}
    \captionof{figure}{Ablation study on the mid-level layer of the VGG-16 surrogate model.}
    \label{fig:supp_midlevellayer_ablation}    
    \vspace{-2mm}
\end{wraptable}
Cross-task results in Table~\ref{tab:supp_cross_task} show only minor differences among existing generative attack methods, with all approaches achieving similarly low attack success rates in this setting. These uniformly modest outcomes highlight the difficulty of transferring adversarial examples crafted on an image classification–oriented surrogate model to tasks with different objectives, since the perturbations fail to align with the target task’s feature representations. 
Nevertheless, integrating our early-block semantic consistency into each baseline yields small but consistent improvements, demonstrating that preserving coarse semantic cues still provides a performance boost even under the strict black-box evaluation.



\paragraph{Mid-level layer variations.}
%

To further verify that our proposed method is compatible with the baseline~\cite{zhang2022beyond} shown to perform best at the selected mid-level layer of the VGG-16 surrogate model (\ie Maxpooling.3), we conducted an ablation study of the mid-level layer in Fig.~\ref{fig:supp_midlevellayer_ablation}. For reference, Maxpooling.1--5 have resolutions of 112$^{2}$, 56$^{2}$, 28$^{2}$, 14$^{2}$, and 7$^{2}$, respectively.
Across domain, model, and two tasks (SS and OD), we observe that our method added to the baseline still maintains the best strength of the attack at the selected mid-level layer, Maxpooling.3, compared to the other early or late layers.

\begin{figure}[!h]
    \centering
    \begin{minipage}{.38\linewidth}
        \includegraphics[width=\linewidth]{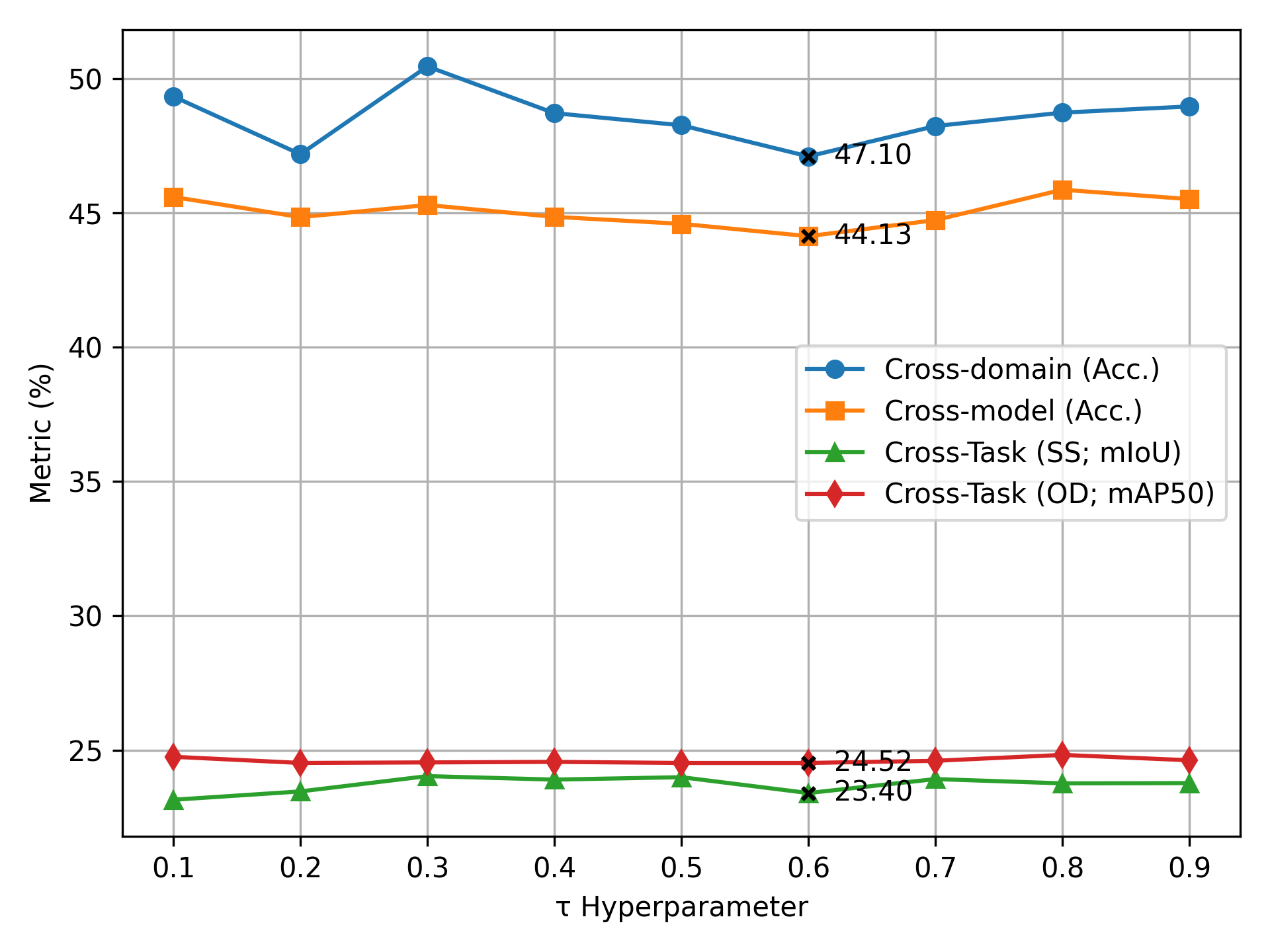}
        \vspace{-5mm}
        \caption{Sensitivity of $\tau$.}
        \label{fig:ablation_tau}
    \end{minipage}%
    \hspace{1mm}
    \begin{minipage}{0.6\linewidth}
        \centering
        \centering
\captionof{table}{Comparison of cross-setting performance and image perceptual quality of AEs.
}
\vspace{-0.3cm}
\setlength{\tabcolsep}{2pt}
\renewcommand{\arraystretch}{1}
\renewcommand{\aboverulesep}{3pt}
\renewcommand{\belowrulesep}{3pt}
\resizebox{\linewidth}{!}{
\begin{tabular}{ccccc|ccc}
\toprule
     \textbf{Method} 
     & \multicolumn{4}{c}{\textbf{Cross-setting} (Avg.)} 
     &  \multicolumn{3}{c}{\textbf{Perceptual Quality}} \\
 \cmidrule{2-8}
     & Domain (Acc.) & Model (Acc.) & Task (SS; mIoU) & Task (OD; mAP50) 
     & PSNR $\uparrow$  
     & SSIM $\uparrow$ 
     & LPIPS $\downarrow$ \\
\midrule
    CDA
    & 69.94 & 50.27 & 22.90 & 29.54
    & 29.11 
    & 0.78 
    & 0.43\\
\rowcolor{Goldenrod!25}  w/ Ours 
    & \textbf{54.82} & \textbf{43.38} & \textbf{22.71} & \textbf{28.83}
    & 29.17 {\tiny(\textcolor{ForestGreen}{+0.06})} & 0.78 {\tiny(--)}
    & 0.43 {\tiny(--)}  \\
    \hline
    
    LTP
    & 49.91 & 48.33 & 25.34 & 25.90 
    & 29.11 
    & 0.76 
    & 0.47 \\
\rowcolor{Goldenrod!25}  w/ Ours  
    & \textbf{40.76} & \textbf{41.99} & \textbf{24.48} & \textbf{24.52}
    & 29.26 {\tiny(\textcolor{ForestGreen}{+0.15})} & 0.77 {\tiny(\textcolor{ForestGreen}{+0.01})}
    & 0.49 {\tiny(+0.02)}  \\
     \hline
     
     BIA
     & 51.07 & 45.17 & 24.75 & 24.72
     & 28.08 
     & 0.75 
     & 0.49 \\
\rowcolor{Goldenrod!25}w/ Ours  
    & \textbf{47.10} &  \textbf{44.13} & \textbf{23.40} & \textbf{24.52}
    & 28.76 {\tiny(\textcolor{ForestGreen}{+0.68})} & 0.75 {\tiny(--)} 
    & 0.49 {\tiny(--)} \\ 
     \hline
     
     GAMA
     & 48.56 & 44.58 & 25.82 & 24.36 
     & 28.62 & 0.74 
     & 0.49 \\
\rowcolor{Goldenrod!25}w/ Ours 
    & \textbf{46.09} & \textbf{43.46} & \textbf{24.81} & \textbf{24.20}
    & 28.69 {\tiny(\textcolor{ForestGreen}{+0.07})} & 0.74 {\tiny(--)} 
    & 0.49 {\tiny(--)} \\
     \hline
     
     FACL
     & 44.05 & 42.00 & 25.08 & 24.43
     & 28.61 & 0.74 
     & 0.49 \\
 \rowcolor{Goldenrod!25}w/ Ours 
    & \textbf{41.78} & \textbf{40.92} & \textbf{24.20} & \textbf{23.97}
     & 28.67 {\tiny(\textcolor{ForestGreen}{+0.05})}& 0.74 {\tiny(--)}
     & 0.49 {\tiny(--)} \\
     \hline
     
     PDCL
     & 43.91 & 42.84 & 25.24 & 24.93
     & 28.68 & 0.74 
     & 0.48 \\
\rowcolor{Goldenrod!25}w/ Ours  
    & \textbf{43.06} & \textbf{42.77} & \textbf{24.20} & \textbf{24.20}
    & 28.70 {\tiny(\textcolor{ForestGreen}{+0.02})}& 0.74 {\tiny(--)}
    & 0.49 {\tiny(--)} \\
\bottomrule
\end{tabular}
}
\label{tab:supp_perceptual}

    \end{minipage}
\end{figure}

\paragraph{Ablation study on the hyperparameter $\tau$.}
To assess the sensitivity of $\tau$ in Eq.~2, we conducted a sweep of hyperparameter values in Fig.~\ref{fig:ablation_tau}. Across the range of values from 0 to 1, we find that our optimal value of $\tau=0.6$ best balances the strength of the attack across all four cross-settings.

\paragraph{Slight improvements in perceptual quality.}
In Table~\ref{tab:supp_perceptual}, we compare PSNR, SSIM, and LPIPS scores for each baseline alone versus the same baseline augmented with our early‐block semantic consistency mechanism. Across all baselines, adding our method results in a slight PSNR increase while SSIM and LPIPS remain effectively unchanged. These minimal or positive changes confirm that our approach does not introduce any perceptual degradation. Instead, it preserves, and in some cases slightly enhances, the visual fidelity of adversarial examples even as it strengthens their transferability.

\subsection{Additional qualitative results}
\begin{figure*}[!h]
    \centering
    \includegraphics[width=\linewidth]{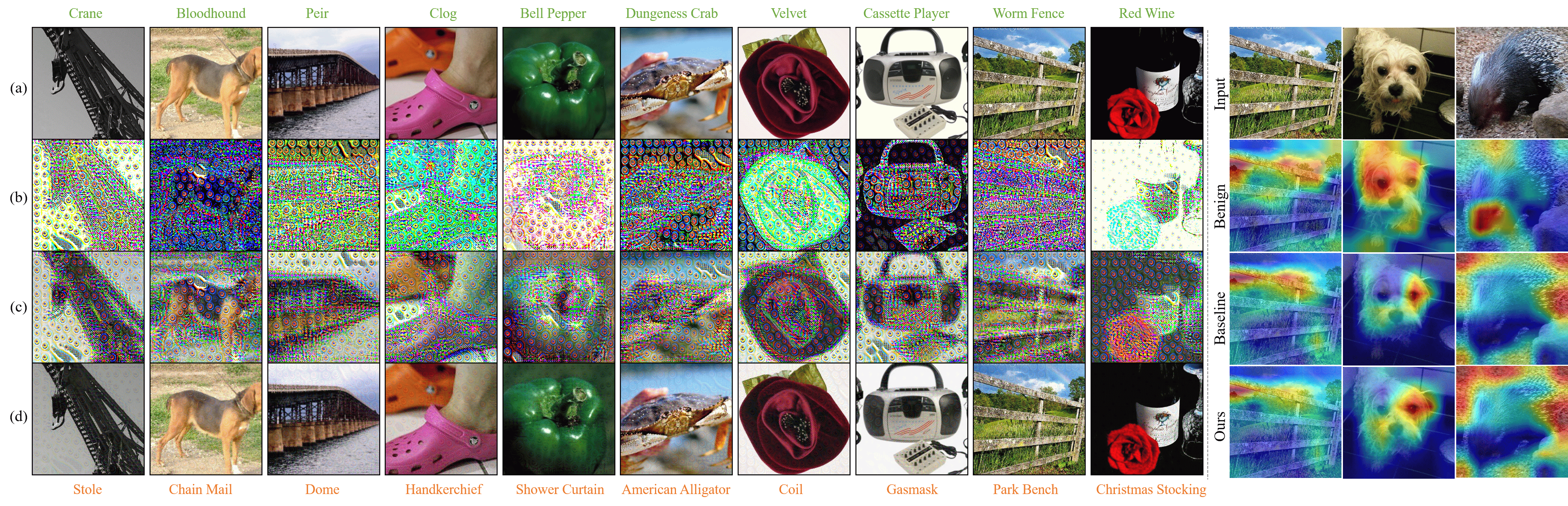}
    \vspace{-6mm}
    \caption{\textbf{Additional qualitative results.} 
    Our semantic structure-aware attack successfully guides the generator to focus perturbations particularly on the semantically salient regions, effectively fooling the victim classifier. 
    \textit{Left:} (a) benign input image, (b) generated perturbation (normalized for visual purposes only), (c) unbounded adversarial image, and (d) bounded adversarial image. The label on top (\textcolor{ForestGreen}{green}) and bottom (\textcolor{Orange}{orange}) denote the correct label and prediction after the attack, respectively.
    \textit{Right:} 
    We highlight that our method induces Grad-CAM~\cite{selvaraju2017gradcam} to focus on \textit{drastically different regions} in our adversarial examples compared to both the benign image and the adversarial examples crafted by the baseline~\cite{zhang2022beyond}. Moreover, our approach \textit{noticeably spreads and reduces the high activation regions} observed in the benign and baseline cases, enhancing the transferability of our adversarial perturbations.
    }
    \label{fig:supp_qual}
\end{figure*}

In these additional qualitative results in Fig.~\ref{fig:supp_qual}, we observe two clear patterns in the adversarial masks. First, straight lines trace the edges of objects, reinforcing the primary structural cues. Second, circular ring shapes appear in the background, helping to disperse noise across non‐object regions. Grad‐CAM visualizations on the right show that our method also drives adversarial activations to much higher levels than those seen in the benign image and boosts areas that exhibited only modest responses under baseline attacks. By combining precise noise along the boundaries with amplified feature activations, our approach anchors noise to the most meaningful contours while strengthening weaker signals, producing stronger and more transferable adversarial examples.

\begin{figure*}[!h]
    \centering
    \includegraphics[width=\linewidth]{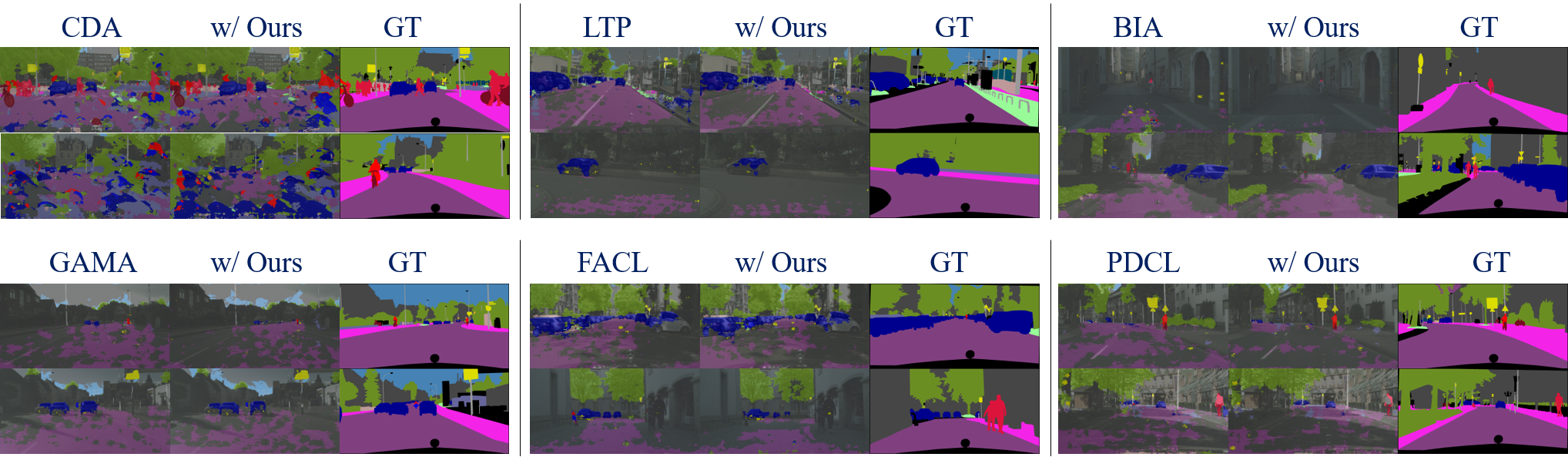}
    \caption{\textbf{Additional cross-task (SS) qualitative results.} 
    }
    \label{fig:supp_qual_task_ss}
\end{figure*}
\begin{figure*}[!h]
    \centering
    \includegraphics[width=\linewidth]{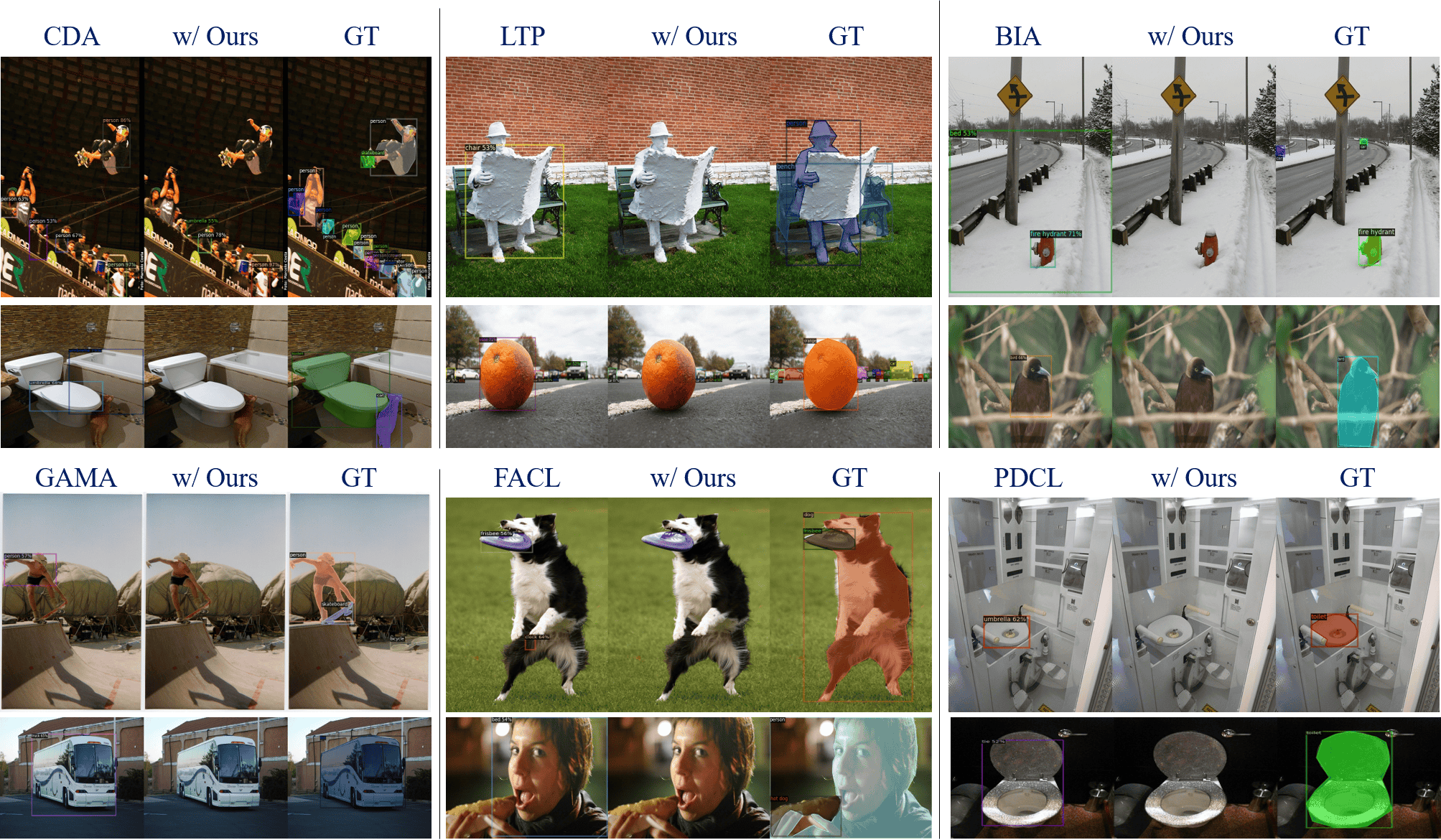}
    \caption{\textbf{Additional cross-task (OD) qualitative results.}
    }
    \label{fig:supp_qual_task_od}
\end{figure*}

We also provide additional qualitative results on the cross-task (SS and OD) settings. 
In our additional segmentation examples in Fig.~\ref{fig:supp_qual_task_ss}, we observe that our method does not just blur or hide parts of the road, but it actually makes the model stop recognizing entire road areas and even small objects like pedestrians or cars. The baseline attack might only erase a few isolated pixels or blend edges, but ours turns whole stretches of road into “ignore,” wiping out those predictions in one go. In other words, our method uniformly removes both large surfaces and tiny details, so the segmented map ends up missing key pieces of the scene that the baseline leaves untouched.
Similarly, in our additional object detection examples in Fig.~\ref{fig:supp_qual_task_od}, our attack causes the model to stop predicting any localized boxes around objects (RoI), completely removing every predicted region of interest, whereas the baseline often leaves boxes in place or only shifts them slightly.

\begin{figure*}[!ht]
    \centering
    \includegraphics[width=0.9\linewidth]{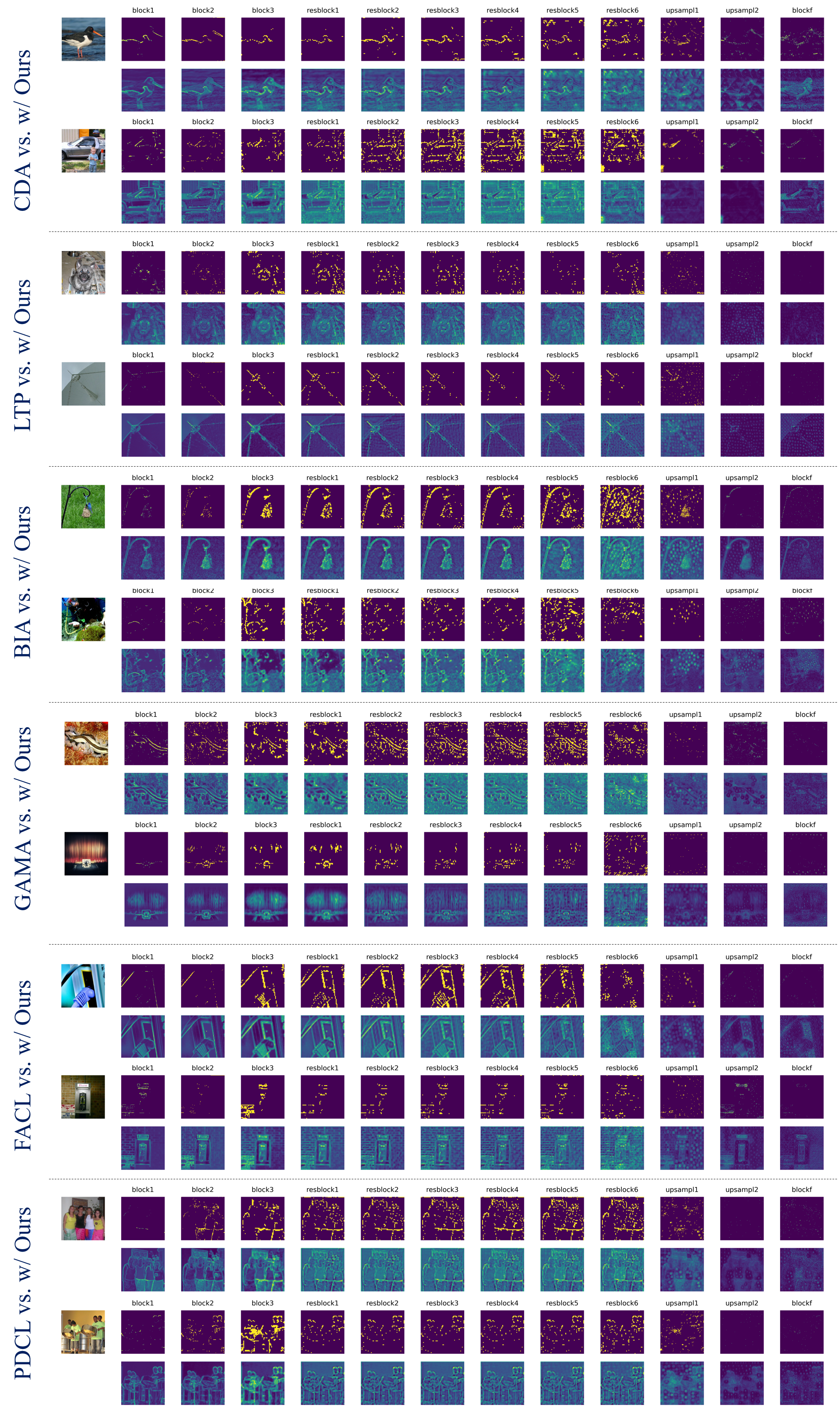}
    \vspace{-3mm}
    \caption{\textbf{Feature difference map comparisons.} Our method noticeably adds noise to object-salient regions in the generator intermediate features, as visible by the distinctive difference from each baseline. For each input on the leftmost column, we visualize the output feature map of each block in the generator in each column (\textit{left}$\rightarrow$\textit{right}): thresholded ($\tau=0.6$) binary mask (\textit{row 1}) after min-max normalization and feature activation difference maps (\textit{row 2}). In the in the \texttt{resblocks} in particular, our method further guides perturbations around the object semantic structure, enhancing transferable noise generation.}
    \label{fig:supp_diffmap}
\end{figure*}

\paragraph{Feature difference map analysis.}
Going further from the investigation of the difference map in the existing work in \texttt{ block3} only, which is the input to the residual blocks, we expanded the analysis into all the blocks, paying particular attention to the intermediate blocks (\texttt{resblocks}). To this end, we compare each baseline to ours and visualized block-wise feature difference in Fig.~\ref{fig:supp_diffmap}. We observed that our method further boosts the object-centric regions (foreground) towards the early intermediate blocks (\texttt{reblocks}), and gradually induces perturbation to be generated towards background, or regions away from the objects directly.

\paragraph{Applicability to state-of-the-art generator-based targeted attacks.}
We conduct state-of-the-art targeted black-box attack experiments following M3D~\cite{zhao2023minimizing} and CGNC~\cite{fang2024clip} on ImageNet for a single training epoch, using the same VGG-family models as surrogates. Perturbations are bounded by $\ell_\infty$ with $\epsilon = 16/255$. Higher TSR indicates a stronger targeted attack. For M3D, TSR is averaged over target classes
$\{24, 99, 245, 344, 471, 555, 661, 701, 802, 919\}$. For CGNC, TSR is averaged over target classes $\{150, 426, 843, 715, 952, 507, 590, 62\}$ in the normal mode.

\begin{table}[!h]
    \centering
    \caption{
    Target success rate (TSR, \%) for CLIP-guided CGNC attack (surrogate: VGG19). Higher TSR indicates a stronger targeted attack. TSR is averaged over target classes $\{150, 426, 843, 715, 952, 507, 590, 62\}$ in normal mode. All experiments use ImageNet, VGG-family surrogates, and $\ell_\infty$ perturbations with $\epsilon = 16/255$.}
    \label{tab:cgnc_vgg19}
    \resizebox{\linewidth}{!}{
    \begin{tabular}{lcccccccc}
        \toprule
            & \multicolumn{7}{c}{\textbf{Victim model}} & \multirow{2}{*}{\textbf{Avg.}}\\ \cmidrule{2-8}
            Method  & VGG16 & GoogLeNet & Inc-v3 & Res152 & Dense121 & Inc-v4 & IncRes-v2 &  \\
        \midrule
            CGNC~\cite{fang2024cgnc}     & 14.71 & 2.03 & 2.77 & 2.68 & 8.31 & 2.41 & 0.96 & 4.84 \\
           \rowcolor{Goldenrod!20} w/ Ours  & 47.50 & 7.90 & 10.96 & 12.24 & 31.29 & 13.36 & 3.98 & \textbf{18.18} \\
        \bottomrule
    \end{tabular}}
\end{table}

\textbf{Key observations.} \vspace{-3mm}
\begin{itemize}[leftmargin=*] \setlength\itemsep{-0.2mm}
    \item \textbf{Large average gain.} Our method improves the average TSR from \textbf{4.84\%}~$\rightarrow$~\textbf{18.18\%}, achieving a \textbf{3.7$\times$ relative increase}, despite the single-epoch training constraint.
    \item \textbf{Consistent gains across victim architectures.} Every victim model benefits, with especially strong improvements on \textbf{DenseNet121 (+22.98\%)} and \textbf{Inception-v4 (+10.95\%)}, indicating that generator-internal semantic consistency remains effective even under CLIP guidance.
\end{itemize}
\begin{table}[htbp]
\centering
\caption{Performance comparison of M3D and M3D w/Ours across architectures. Final column shows average.}
\vspace{-2mm}
\resizebox{\textwidth}{!}{
\begin{tabular}{lcccccccccccc}
\toprule
\textbf{Model \ Target} & \textbf{24} & \textbf{99} & \textbf{245} & \textbf{344} & \textbf{471} & \textbf{555} & \textbf{661} & \textbf{701} & \textbf{802} & \textbf{919} & \textbf{Avg.} \\
\midrule
\multicolumn{12}{l}{\textbf{DenseNet121}} \\
M3D~\cite{zhao2023m3d} & 71.24 & 77.12 & 73.35 & 85.37 & 71.75 & 73.79 & 60.63 & 78.70 & 69.66 & 17.34 & 67.90 \\
\rowcolor{Goldenrod!20} w/ Ours & 76.53 & 77.11 & 82.64 & 87.74 & 82.81 & 78.72 & 74.12 & 78.29 & 79.43 & 47.20 & \textbf{76.46} \\
\midrule
\multicolumn{12}{l}{\textbf{ResNet50}} \\
M3D~\cite{zhao2023m3d} & 68.54 & 75.48 & 77.67 & 79.43 & 77.64 & 80.05 & 54.48 & 89.04 & 55.85 & 9.51 & 66.77 \\
\rowcolor{Goldenrod!20} w/ Ours & 71.02 & 73.60 & 80.69 & 88.04 & 87.64 & 83.88 & 68.97 & 84.88 & 69.60 & 45.36 & \textbf{75.37} \\
\midrule
\multicolumn{12}{l}{\textbf{ResNet152}} \\
M3D~\cite{zhao2023m3d} & 50.73 & 62.76 & 63.15 & 69.53 & 60.96 & 57.82 & 37.28 & 73.87 & 37.82 & 11.29 & 52.52 \\
\rowcolor{Goldenrod!20} w/ Ours & 60.35 & 58.80 & 68.61 & 79.73 & 73.64 & 65.47 & 56.12 & 70.29 & 52.77 & 39.63 & \textbf{62.54} \\
\midrule
\multicolumn{12}{l}{\textbf{WRN-50-2}} \\
M3D~\cite{zhao2023m3d} & 64.55 & 68.95 & 73.51 & 69.53 & 73.04 & 66.53 & 46.85 & 84.39 & 45.17 & 13.15 & 60.57 \\
\rowcolor{Goldenrod!20} w/ Ours & 72.41 & 69.22 & 77.94 & 79.61 & 82.97 & 69.83 & 66.16 & 76.32 & 58.20 & 41.01 & \textbf{69.37} \\
\bottomrule
\end{tabular}
}
\end{table}

\textbf{Key observations.} \vspace{-3mm}
\begin{itemize}[leftmargin=*] \setlength\itemsep{-0.2mm}
    \item \textbf{Consistent gains on a strong targeted attack.} The overall average TSR across all four victims increases from \textbf{61.94\%}~$\rightarrow$~\textbf{70.94\%}, with absolute gains of \textbf{+8.56\%p (DenseNet121)}, \textbf{+8.6\%p (ResNet50)}, \textbf{+10.02\%p (ResNet152)}, and \textbf{+8.8\%p (WRN-50-2)}.
    \item \textbf{Across-target robustness.} Improvements are observed across a randomized set of 10 target classes, not just “easy” ones. For example, on \textbf{ResNet50}, class 919 improves from 9.51\%~$\rightarrow$~45.36\%, showing that generator-centric regularization effectively steers features toward diverse target semantics even under tight training.
\end{itemize}


\paragraph{
Attack robustness against purification defense.}
Further looking into how robust our attack performs against purification methods such as NRP~\cite{NRP}, we report the improvements with Ours in Table~\ref{tab:supp_purification}. On most of the baselines, our method addition maintains lower Accuracy, ASR and FR scores than the baseline alone, while on the recent advanced methods (\eg FACL~\cite{yang2024facl} and PDCL~\cite{yang2024pdcl}), our method very slightly maintains similar scores with those of the baseline alone on the three metrics. On the other hand, on the more challenging ACR metric, we observe that our method slightly improves from the baseline, entailing that our generative feature-level tuning further reduces the number of inadvertently corrected samples.

Running DiffPure on ImageNet is computationally intensive and the cost grows with the 
number of samples, so within the rebuttal time frame, we conducted preliminary experiments 
on a random subset of 1k validation images. We plan to extend to the full 50k val set 
given sufficient time.

Our framework is designed to improve adversarial transferability on undefended models, 
following standard protocols in CDA, LTP, BIA, GAMA, FACL, and PDCL, rather than to 
construct an adaptive attack specifically tailored to circumvent purification defenses.

Under DiffPure with $\epsilon = 10/255$ and $t = 150$, incorporating our method yields 
robust accuracies that remain comparable to the baselines on both victims. Fluctuations 
are small in magnitude and are plausibly explained by the stochastic diffusion process 
and the limited 1k sample size, rather than by a systematic loss of robustness.


\begin{table}[!h]
    \centering
    \setlength{\tabcolsep}{1pt}
    \renewcommand{\arraystretch}{1.12}
    \renewcommand{\aboverulesep}{3pt}
    \renewcommand{\belowrulesep}{3pt}
    \caption{Preliminary robustness evaluations against DiffPure ($\epsilon = 10/255$, $t = 150$) on 1k samples (Cls. Acc., \%).}
    \label{tab:diffpure}
    \vspace{-2mm}

    \resizebox{\linewidth}{!}{
    \begin{tabular}{l cc cc cc cc}
        \toprule
        \textbf{Victim} & CDA~\cite{naseer2019cross} & \cellcolor{Goldenrod!20} \textbf{w/ Ours} & LTP~\cite{salzmann2021learning} & \cellcolor{Goldenrod!20} \textbf{w/ Ours} & BIA~\cite{zhang2022beyond} & \cellcolor{Goldenrod!20} \textbf{w/ Ours} \\
        \midrule
        \textbf{Res152}   & 67.7 & \cellcolor{Goldenrod!20}66.3 & 66.1 & \cellcolor{Goldenrod!20}68.0 & 66.1 & \cellcolor{Goldenrod!20}66.8 \\
        \textbf{Dense121} & 62.0 & \cellcolor{Goldenrod!20}60.5 & 62.9 & \cellcolor{Goldenrod!20}64.0 & 62.1 & \cellcolor{Goldenrod!20}61.4 \\
        \bottomrule
    \end{tabular}}

    \vspace{-1mm} 

    \resizebox{\linewidth}{!}{
    \begin{tabular}{l cc cc cc cc}
        \toprule
        \textbf{Victim} & GAMA~\cite{aich2022gama} & \cellcolor{Goldenrod!20} \textbf{w/ Ours} & FACL~\cite{yang2024facl} & \cellcolor{Goldenrod!20} \textbf{w/ Ours} & PDCL~\cite{yang2024pdcl} & \cellcolor{Goldenrod!20} \textbf{w/ Ours} \\
        \midrule
        \textbf{Res152}   & 65.8 & \cellcolor{Goldenrod!20}66.6 & 65.4 & \cellcolor{Goldenrod!20}66.2 & 65.3 & \cellcolor{Goldenrod!20}65.2 \\
        \textbf{Dense121} & 62.0 & \cellcolor{Goldenrod!20}61.6 & 61.8 & \cellcolor{Goldenrod!20}62.1 & 61.5 & \cellcolor{Goldenrod!20}61.4 \\
        \bottomrule
    \end{tabular}}

\end{table}

\noindent\textbf{Key observations:} \vspace{-2mm}
\begin{itemize}[leftmargin=*] \setlength\itemsep{-0.2mm}
    \item On ResNet152, accuracies with Ours stay close to the baselines, with modest 
    increases (e.g., LTP, BIA, GAMA, FACL) and modest decreases (e.g., CDA), all within 
    a narrow band.
    \item On DenseNet121, deviations are similarly small, and several pairs 
    (e.g., LTP, FACL) show slight improvements.
    \item Overall, these preliminary results indicate that our method remains largely 
    compatible with strong diffusion based purification. The semantic consistency 
    enforced during generation improves transfer on standard models without causing a 
    significant loss of robustness under DiffPure compared to the original baselines.
\end{itemize}


\paragraph{Additional results on baselines and w/ Ours against against input processing methods.}
In addition to the results in Table 4 against robustly trained models, we further report the results on the other baselines that are left out due to page limitations in Table~\ref{tab:supp_defense_eval}.

\begin{table}[!h]
    \centering
    \caption{
    Defense evaluation comparisons against other baselines.}
    \label{tab:supp_defense_eval}
    \vspace{-3mm}
    \resizebox{\linewidth}{!}{%
    \begin{tabular}{l l c c c c c c c}
        \toprule
        Method  & Metric & \textbf{Adv.IncV3} & \textbf{Adv.ViT} & \textbf{Adv.ConvNeXt} & \textbf{JPEG} & \textbf{BDR} & \textbf{R\&P} & \textbf{Avg.} \\

        \midrule
              Benign & Acc. (\%) $\downarrow$ & 76.33 & 48.82 & 58.44 & 74.68 & 74.68 & 76.58 &  68.26 \\
        \midrule
            \multirow{4}{*}{\textbf{BIA}} & 
            Acc. (\%)  $\downarrow$       & 68.54 & 45.64 & 53.88 & 63.49 & 47.82 & 44.78 & 54.03 \\
            
            & ASR (\%) $\uparrow$                & 14.95 & 11.72 & 10.26 & 20.24 & 40.76 & 44.59 &  23.75 \\
           
            & FR (\%) $\uparrow$  & 24.02 & 25.48 & 19.40 & 28.09 & 48.06 & 51.60 &  32.78 \\
        \cmidrule{2-9}
            & ACR (\%) $\downarrow$          & 15.30 &  4.96 &  3.46 & 11.45 & 11.30 & 10.56 &   9.51 \\
        \midrule
            \rowcolor{Goldenrod!20}& Acc. (\%) $\downarrow$         & 67.92 & 45.33 & 53.62 & 60.83 & 44.07 & 39.01 & \textbf{51.80} \\            
            \rowcolor{Goldenrod!20}& ASR (\%) $\uparrow$                & 15.75 & 11.95 & 10.65 & 23.74 & 45.37 & 51.63 & \textbf{26.52} \\
            \rowcolor{Goldenrod!20}& FR (\%) $\uparrow$ & 24.83 & 25.31 & 19.60 & 31.61 & 52.22 & 57.86 & \textbf{35.28} \\
        \cmidrule{2-9}
            \rowcolor{Goldenrod!20} \multirow{-4}{*}{\textbf{BIA w/ Ours}} & ACR (\%) $\downarrow$          & 15.23 &  4.57 &  3.38 & 11.48 & 10.29 &  9.08 &  \textbf{9.01} \\
        
        \midrule
            \multirow{4}{*}{\textbf{GAMA}}
             & Acc. (\%) $\downarrow$ & 66.71 & 45.74 & 53.76 & 59.27 & 41.08 & 37.60 & 50.69 \\
             & ASR (\%) $\uparrow$       & 17.27 & 11.65 & 10.50 & 25.66 & 49.01 & 53.34 & 27.91 \\
             & FR (\%) $\uparrow$        & 26.37 & 25.70 & 19.68 & 33.51 & 55.42 & 59.42 & 36.68 \\
             \cmidrule{2-9}
             & ACR (\%) $\downarrow$     & 15.06 &  5.09 &  3.50 & 11.07 &  9.38 &  8.62 &  8.79 \\
        \midrule
             \rowcolor{Goldenrod!20}& Acc. (\%) $\downarrow$ & 66.37 & 45.50 & 53.81 & 58.99 & 39.07 & 36.00 & \textbf{49.96} \\
             \rowcolor{Goldenrod!20}& ASR (\%) $\uparrow$       & 17.76 & 11.78 & 10.45 & 26.07 & 51.49 & 55.36 & \textbf{28.82} \\
             \rowcolor{Goldenrod!20}& FR (\%) $\uparrow$        & 26.78 & 25.34 & 19.61 & 33.86 & 57.63 & 61.18 & \textbf{37.40} \\
             \cmidrule{2-9}
             \rowcolor{Goldenrod!20} \multirow{-4}{*}{\textbf{GAMA w/ Ours}}& ACR (\%) $\downarrow$     & 15.19 &  4.74 &  3.56 & 11.19 &  8.88 &  8.36 &  \textbf{8.65} \\
        \midrule
            \multirow{4}{*}{\textbf{FACL}}
             & Acc. (\%) $\downarrow$ & 65.68 & 45.17 & 53.12 & 47.25 & 38.36 & 33.31 & 47.15 \\
             & ASR (\%) $\uparrow$       & 18.68 & 12.44 & 11.70 & 41.08 & 52.43 & 58.78 & 32.52 \\
             & FR (\%) $\uparrow$        & 27.75 & 26.24 & 21.19 & 47.99 & 58.39 & 64.16 & 40.95 \\
             \cmidrule{2-9}
             & ACR (\%) $\downarrow$     & 15.22 &  4.73 &  3.66 &  9.90 &  8.91 &  8.03 &  8.41 \\
        \midrule
            \rowcolor{Goldenrod!20}
             \rowcolor{Goldenrod!20}& Acc. (\%) $\downarrow$ & 65.49 & 44.89 & 53.14 & 47.93 & 33.24 & 28.64 & \textbf{45.56} \\
             \rowcolor{Goldenrod!20}& ASR (\%) $\uparrow$       & 18.88 & 12.74 & 11.58 & 40.25 & 58.85 & 64.60 & \textbf{34.48} \\
             \rowcolor{Goldenrod!20}& FR (\%) $\uparrow$        & 27.90 & 26.12 & 20.88 & 47.28 & 64.03 & 69.19 & \textbf{42.57} \\
             \cmidrule{2-9}
             \rowcolor{Goldenrod!20} \multirow{-4}{*}{\textbf{FACL w/ Ours}}& ACR (\%) $\downarrow$     & 15.07 &  4.46 &  3.54 & 10.10 &  7.94 &  7.00 &  \textbf{8.01 }\\
        \midrule
            \multirow{4}{*}{\textbf{PDCL}}
             & Acc. (\%) $\downarrow$ & 67.57 & 45.24 & 53.61 & 58.10 & 39.84 & 37.01 & 50.23 \\
             & ASR (\%) $\uparrow$       & 16.20 & 12.10 & 10.74 & 27.15 & 50.56 & 54.09 & 28.47 \\
             & FR (\%) $\uparrow$        & 25.29 & 25.43 & 19.84 & 34.92 & 56.77 & 59.98 & 37.04 \\
             \cmidrule{2-9}
             & ACR (\%) $\downarrow$     & 15.24 &  4.55 &  3.47 & 10.91 &  9.12 &  8.54 &  \textbf{8.64} \\
        \midrule
            \rowcolor{Goldenrod!20}
             \rowcolor{Goldenrod!20}& Acc. (\%) $\downarrow$ & 67.53 & 45.13 & 53.48 & 57.63 & 39.53 & 35.67 & \textbf{49.83} \\
             \rowcolor{Goldenrod!20}& ASR (\%) $\uparrow$       & 16.24 & 12.38 & 10.88 & 27.82 & 51.06 & 55.78 & \textbf{29.03} \\
             \rowcolor{Goldenrod!20}& FR (\%) $\uparrow$        & 25.27 & 25.69 & 19.92 & 35.54 & 57.29 & 61.50 & \textbf{37.54} \\
             \cmidrule{2-9}
             \rowcolor{Goldenrod!20} \multirow{-4}{*}{\textbf{PDCL w/ Ours}}& ACR (\%) $\downarrow$     & 15.16 &  4.59 &  3.36 & 11.09 &  9.42 &  8.31 &  8.66\\
        \bottomrule
    \end{tabular}%
    }
\end{table}

\paragraph{Additional results on input processing defense.}
Beyond the main experiments, we further evaluate our method under standard input-processing defenses. Rotation (deg) applies small random rotations sampled from a bounded angle range to preserve semantic content while perturbing pixel-level alignment~\cite{guo2017countering, xie2017randomization}. 

\textit{Smoothing} uses standard spatial smoothing filters (Gaussian, median, and mean) to attenuate small high-frequency adversarial perturbations while largely preserving coarse structure, at the cost of some local blurring~\cite{guo2017countering, xu2017bitreduction}. 

\textit{Total variation minimization} (TVM) performs TV-based denoising by approximately minimizing a reconstruction loss with a total-variation regularizer, yielding a piecewise-smooth reconstruction that preserves major edges while suppressing small oscillatory perturbations~\cite{guo2017countering, rudin1992nonlinear}. 

\textit{Pixel deflection} (PD) randomly selects a subset of pixels and replaces each with the value of a randomly chosen neighbor within a local window to stochastically disrupt finely tuned adversarial patterns without destroying global semantics~\cite{prakash2018deflecting}. 

Taken together, JPEG-style compression (JPEG), bit-depth reduction (BDR), random resizing--padding (R\&P), and smoothing, as well as the transformations reported in the additional table, correspond to the standard family of input-processing defenses widely used in prior work on transferable attacks, including our baselines and recent studies such as TransferAttackEval~\cite{zhao2025revisiting} and CGNC~\cite{fang2024clip}. TVM and pixel deflection can be regarded as stronger, yet conceptually similar, input transformations in this family. Evaluating under this common protocol makes our results directly comparable to existing attacks, and across these defenses we observe that adding our generator-side regularizer consistently lowers accuracy and increases FR/ASR relative to each underlying baseline, indicating that our method enhances attack effectiveness against the standard input pre-processing defenses.

\begin{table}[t]
    \centering
    \setlength{\tabcolsep}{3pt}
    \renewcommand{\arraystretch}{1}
    \renewcommand{\aboverulesep}{1pt}
    \renewcommand{\belowrulesep}{1pt}
    \caption{
    Additional input processing defenses.}
    \label{tab:input_processing_defenses}
    \vspace{-3mm}
    \resizebox{\linewidth}{!}{%
    \begin{tabular}{l l c c c c c c c c c c}
        \toprule
            & & &
            \multicolumn{4}{c}{\textbf{Random Rotation (deg)}} &
            \multicolumn{3}{c}{\textbf{Smoothing (Kernel)}} &
            \multicolumn{1}{c}{\textbf{Total Var. Min.}} &
            \multicolumn{1}{c}{\textbf{Random Pixel}} \\
            \cmidrule(lr){4-7}\cmidrule(lr){8-10}\cmidrule(lr){11-11}\cmidrule(lr){12-12}
            & & \textbf{Avg.} & 30 & 50 & 70 & 90 & Gaussian & Median & Mean & TVM & PD \\
        \midrule
            \multirow{4}{*}{\textbf{BIA}}
            & Acc. (\%) $\downarrow$ & 37.08 & 33.59 & 29.20 & 26.27 & 24.92 & 58.75 & 54.27 & 59.06 & 41.17 & 41.62 \\
            & ASR (\%) $\uparrow$      & 53.95 & 58.38 & 63.86 & 67.52 & 69.26 & 26.66 & 32.44 & 26.15 & 49.05 & 48.48 \\
            & FR (\%) $\uparrow$       & 59.56 & 63.89 & 68.66 & 71.81 & 73.34 & 34.81 & 40.36 & 34.46 & 55.38 & 54.98 \\
            & ACR (\%) $\downarrow$    &  8.36 &  7.91 &  6.97 &  6.40 &  6.31 & 11.94 & 11.77 & 11.76 &  9.87 &  9.96 \\
        \midrule
            \rowcolor{Goldenrod!20} 
            \rowcolor{Goldenrod!20} & Acc. (\%) $\downarrow$ & \textbf{34.05} & 28.61 & 28.43 & 22.27 & 21.18 & 55.55 & 49.61 & 57.20 & 37.42 & 38.15 \\
            \rowcolor{Goldenrod!20} & ASR (\%) $\uparrow$      & \textbf{57.75} & 64.52 & 64.79 & 72.44 & 73.86 & 30.84 & 38.44 & 28.63 & 53.61 & 52.77 \\
            \rowcolor{Goldenrod!20} & FR (\%) $\uparrow$       & \textbf{63.01} & 69.32 & 69.50 & 76.14 & 77.43 & 38.88 & 45.82 & 36.70 & 59.60 & 58.86 \\
            \rowcolor{Goldenrod!20} \multirow{-4}{*}{\textbf{BIA w/ Ours}}& ACR (\%) $\downarrow$    & \textbf{7.80}  &  6.62 &  6.76 &  5.36 &  5.34 & 12.01 & 11.37 & 11.88 &  8.76 &  9.09 \\
        \midrule
            \multirow{4}{*}{\textbf{GAMA}}
            & Acc. (\%) $\downarrow$  & 31.66 & 26.10 & 26.25 & 20.18 & 19.10 & 52.47 & 45.53 & 54.05 & 35.08 & 35.78 \\
            & ASR (\%) $\uparrow$      & 60.67 & 67.59 & 67.48 & 75.03 & 76.46 & 34.61 & 43.45 & 32.55 & 56.41 & 55.52 \\
            & FR (\%) $\uparrow$       & 65.60 & 71.93 & 71.88 & 78.35 & 79.57 & 42.41 & 50.40 & 40.47 & 62.00 & 61.17 \\
            & ACR (\%) $\downarrow$    &  7.12 &  5.90 &  6.19 &  4.88 &  4.88 & 11.15 & 10.32 & 11.18 &  7.88 &  7.97 \\
        \midrule
            \rowcolor{Goldenrod!20} 
            \rowcolor{Goldenrod!20}& Acc. (\%) $\downarrow$ & \textbf{30.30} & 25.31 & 21.75 & 19.41 & 18.41 & 52.09 & 43.89 & 53.91 & 32.60 & 33.61 \\
            \rowcolor{Goldenrod!20}& ASR (\%) $\uparrow$      & \textbf{62.37} & 68.63 & 73.06 & 76.01 & 77.29 & 35.09 & 45.46 & 32.76 & 59.55 & 58.28 \\
           \rowcolor{Goldenrod!20} & FR (\%) $\uparrow$       & \textbf{67.08} & 72.92 & 76.65 & 79.26 & 80.33 & 42.71 & 52.09 & 40.62 & 64.77 & 63.66 \\
            \rowcolor{Goldenrod!20} \multirow{-4}{*}{\textbf{GAMA w/ Ours}}& ACR (\%) $\downarrow$    & \textbf{6.86}  &  5.93 &  5.16 &  4.77 &  4.65 & 11.09 &  9.80 & 11.25 &  7.48 &  7.68 \\
        \midrule
            \multirow{4}{*}{\textbf{FACL}}
            & Acc. (\%) $\downarrow$ & 27.12 & 22.00 & 18.64 & 16.68 & 15.84 & 45.91 & 36.29 & 48.38 & 32.40 & 33.11 \\
            & ASR (\%) $\uparrow$      & 66.41 & 72.75 & 77.08 & 79.49 & 80.59 & 42.91 & 55.01 & 39.73 & 59.88 & 59.00 \\
            & FR (\%) $\uparrow$       & 70.66 & 76.42 & 80.21 & 82.20 & 83.13 & 49.88 & 60.58 & 46.97 & 64.60 & 64.31 \\
            & ACR (\%) $\downarrow$    &  6.44 &  5.22 &  4.95 &  4.43 &  4.42 & 10.14 &  8.46 & 10.34 &  7.74 &  7.89 \\
        \midrule
            \rowcolor{Goldenrod!20} 
            \rowcolor{Goldenrod!20}& Acc. (\%) $\downarrow$ & \textbf{25.30} & 19.68 & 16.91 & 15.20 & 14.61 & 43.08 & 33.43 & 46.18 & 30.56 & 31.32 \\
            \rowcolor{Goldenrod!20}& ASR (\%) $\uparrow$      & \textbf{68.69} & 75.78 & 79.19 & 81.35 & 82.13 & 46.35 & 58.54 & 42.50 & 62.15 & 61.28 \\
            \rowcolor{Goldenrod!20}& FR (\%) $\uparrow$       & \textbf{72.68} & 79.02 & 81.99 & 83.83 & 84.49 & 52.95 & 63.78 & 49.47 & 67.02 & 66.33 \\
            \rowcolor{Goldenrod!20} \multirow{-4}{*}{\textbf{FACL w/ Ours}}& ACR (\%) $\downarrow$    & \textbf{6.08}  &  5.15 &  4.47 &  4.15 &  4.16 &  9.29 &  7.77 &  9.97 &  7.25 &  7.66 \\
        \midrule
            \multirow{4}{*}{\textbf{PDCL}}
            & Acc. (\%) $\downarrow$ & 31.34 & 26.04 & 22.50 & 21.84 & 19.02 & 52.68 & 45.87 & 56.05 & 32.97 & 34.40 \\
            & ASR (\%) $\uparrow$      & 60.87 & 67.76 & 72.16 & 73.00 & 76.50 & 31.79 & 43.04 & 29.95 & 59.06 & 57.78 \\
            & FR (\%) $\uparrow$       & 65.75 & 72.10 & 75.91 & 76.60 & 79.65 & 39.75 & 50.02 & 38.04 & 64.29 & 63.17 \\
            & ACR (\%) $\downarrow$    &  7.09 &  6.20 &  5.39 &  5.35 &  4.67 & 11.42 & 10.41 & 11.30 &  7.46 &  7.87 \\
        \midrule
            \rowcolor{Goldenrod!20} 
            \rowcolor{Goldenrod!20}& Acc. (\%) $\downarrow$ & \textbf{30.80} & 25.20 & 21.84 & 19.58 & 18.63 & 53.77 & 45.34 & 55.29 & 32.64 & 33.71 \\
            \rowcolor{Goldenrod!20}& ASR (\%) $\uparrow$      & \textbf{61.76} & 68.80 & 73.00 & 75.85 & 77.04 & 32.87 & 43.68 & 30.98 & 59.54 & 58.26 \\
            \rowcolor{Goldenrod!20}& FR (\%) $\uparrow$       & \textbf{66.54} & 73.03 & 76.60 & 79.06 & 80.16 & 40.66 & 50.63 & 38.92 & 64.34 & 63.74 \\
            \rowcolor{Goldenrod!20} \multirow{-4}{*}{\textbf{PDCL w/ Ours}}& ACR (\%) $\downarrow$    & \textbf{7.02}  &  6.00 &  5.35 &  4.97 &  4.80 & 11.03 & 10.22 & 11.39 &  7.63 &  8.06 \\

        \bottomrule
    \end{tabular}%
    }
\end{table}

\paragraph{Attack robustness on zero-shot image classification.}
We also evaluated our method on the zero-shot image classification task with the well-known CLIP~\cite{CLIP} vision-language model in Table~\ref{tab:supp_zsclassification}. Here, we observe that, except for BIA~\cite{zhang2022beyond} and FACL~\cite{yang2024facl}, we observe boosted attacked accuracy when we add our method to the baselines~\cite{naseer2019cross, salzmann2021learning, aich2022gama, yang2024pdcl}. We conjecture that the slight attack strength degradation owes to the baseline method that has already been well-fitted to generate adversarial examples effective for the zero-shot setting based on the relatively lower accuracy scores than the rest. We posit that the respective well-trained generator is already adept enough that our method may interfere with the learned generator weights negatively, and there may exist a maximum capacity at which AEs from generative model-based attacks can attack victim models. 

\paragraph{Random trials for Baseline~\cite{zhang2022beyond} with Ours.}
In Table~\ref{tab:supp_randomtrials}, we further show that our method exhibits stable training results (mean$\pm$std.dev.) as shown from multiple random seed trials evaluated on all four cross-settings.

\begin{table*}[!t]
\centering
\begin{minipage}{0.48\linewidth}
    \centering
    \vspace{-5mm}
    \caption{Adversarial transferability results with our method against the purification method tested on the Inc-V3 victim model, and random seed testing. Better results in \textbf{boldface}.}
    \label{tab:supp_purification}
    \setlength{\tabcolsep}{10pt}
    \renewcommand{\arraystretch}{1}
    \renewcommand{\aboverulesep}{.1pt}
    \renewcommand{\belowrulesep}{.1pt}
    \resizebox{\linewidth}{!}{
        \begin{tabular}{ccccc}
            \toprule                
                & & \multicolumn{2}{c}{\textbf{Purification}} & \multirow{2}{*}{\textbf{Avg.}} \\ \cmidrule{3-4}
                \textbf{Method} & \textbf{Metric} & NRP
                & NRP-ResNet
                \\
            \midrule
                 Benign & Acc. \% & \multicolumn{3}{c}{76.19} \\
            \midrule    
                \multirow{4}{*}{CDA} 
                & Acc. (\%) $\downarrow$ & 71.14 & 67.04 & 69.09\\
                & ASR (\%) $\uparrow$ & 10.29 & 15.94 & 13.11\\
                & FR (\%) $\uparrow$ & 17.45 & 24.00 & 20.73 \\ \cmidrule{2-5}
                & ACR (\%) $\downarrow$ & 11.73 & 12.53 &  \textbf{12.13} \\
            \midrule
                \rowcolor{Goldenrod!20} & Acc. (\%) $\downarrow$ & 70.90 & 66.16 &  \textbf{68.53} \\
                \rowcolor{Goldenrod!20} & ASR (\%) $\uparrow$ & 10.73 & 17.07 &  \textbf{13.90}  \\
                \rowcolor{Goldenrod!20} & FR (\%) $\uparrow$ & 18.15 & 25.25 &  \textbf{21.70} \\\cmidrule{2-5}
                \rowcolor{Goldenrod!20} \multirow{-4}{*}{w/ Ours} & ACR (\%) $\downarrow$ &  12.11 & 12.51 & 12.31 \\ 
            \midrule
                \multirow{4}{*}{LTP}
                & Acc. (\%) $\downarrow$ & 72.19 & 67.65 & 69.92 \\
                & ASR (\%) $\uparrow$ & 8.82 & 15.22 & 12.02 \\
                & FR (\%) $\uparrow$ & 15.59 & 23.15 & 19.37 \\\cmidrule{2-5}
                & ACR (\%) $\downarrow$ & 11.42 & 12.83 &  \textbf{12.13} \\ 
            \midrule
                \rowcolor{Goldenrod!20} & Acc. (\%) $\downarrow$ & 71.78 & 65.51 &  \textbf{68.65} \\
                \rowcolor{Goldenrod!20} & ASR (\%) $\uparrow$ & 9.45 & 17.97 &  \textbf{13.71}\\
                \rowcolor{Goldenrod!20} & FR (\%) $\uparrow$ & 16.34 & 25.98 &  \textbf{21.16} \\\cmidrule{2-5}
                \rowcolor{Goldenrod!20} \multirow{-4}{*}{w/ Ours} & ACR (\%) $\downarrow$ & 11.69 & 12.63 & 12.16 \\ 
            \midrule
                \multirow{4}{*}{BIA}
                & Acc. (\%) $\downarrow$ & 73.84 & 71.93 & 72.89 \\
                & ASR (\%) $\uparrow$ & 6.37 & 9.20 & 7.79 \\
                & FR (\%) $\uparrow$ & 12.46 & 16.35 & 14.41 \\\cmidrule{2-5}
                & ACR (\%) $\downarrow$ & 19.51 & 11.53 &  \textbf{11.02} \\ 
            \midrule 
                \rowcolor{Goldenrod!20} & Acc. (\%) $\downarrow$ &  73.85 & 71.40 &  \textbf{72.63}\\
                \rowcolor{Goldenrod!20} & ASR (\%) $\uparrow$ & 6.36 & 10.00 &  \textbf{8.18} \\
                \rowcolor{Goldenrod!20} & FR (\%) $\uparrow$ &  12.63 & 17.31 &  \textbf{14.97} \\\cmidrule{2-5}
                \rowcolor{Goldenrod!20} \multirow{-4}{*}{w/ Ours} & ACR (\%) $\downarrow$ & 10.52 & 11.89 & 11.21 \\ 
            \midrule
                \multirow{4}{*}{GAMA}
                & Acc. (\%) $\downarrow$ & 74.42 & 72.30 & 73.36 \\
                & ASR (\%) $\uparrow$ & 5.35 & 8.53 & 6.94 \\
                & FR (\%) $\uparrow$ &  10.96 & 15.35 & 13.16\\\cmidrule{2-5}
                & ACR (\%) $\downarrow$ & 9.67 & 10.96 &  \textbf{10.32} \\ 
            \midrule
                \rowcolor{Goldenrod!20} & Acc. (\%) $\downarrow$ &  74.31 & 71.78 &  \textbf{73.05}\\
                \rowcolor{Goldenrod!20} & ASR (\%) $\uparrow$ &  5.65 & 9.32 &  \textbf{7.49}\\
                \rowcolor{Goldenrod!20} & FR (\%) $\uparrow$ & 11.60 & 16.32 &  \textbf{13.96} \\\cmidrule{2-5}
                \rowcolor{Goldenrod!20} \multirow{-4}{*}{w/ Ours} & ACR (\%) $\downarrow$ & 10.17 & 11.29 & 10.73 \\ 
            \midrule
                \multirow{4}{*}{FACL}
                & Acc. (\%) $\downarrow$ & 74.23 & 71.95 &  \textbf{73.09} \\
                & ASR (\%) $\uparrow$ & 5.68 & 9.07 &  \textbf{7.38} \\
                & FR (\%) $\uparrow$ & 11.57 & 15.94 &  \textbf{13.76} \\\cmidrule{2-5}
                & ACR (\%) $\downarrow$ & 9.94 & 11.22 & 10.58 \\ 
            \midrule
                \rowcolor{Goldenrod!20} & Acc. (\%) $\downarrow$ & 74.21 & 72.00 & 73.11 \\
                \rowcolor{Goldenrod!20} & ASR (\%) $\uparrow$ & 5.69 & 8.98 & 7.34 \\
                \rowcolor{Goldenrod!20} & FR (\%) $\uparrow$ & 11.42 & 15.81 & 13.62 \\\cmidrule{2-5}
                \rowcolor{Goldenrod!20} \multirow{-4}{*}{w/ Ours} & ACR (\%) $\downarrow$ & 9.90 & 11.11 &  \textbf{10.51} \\ 
            \midrule
                \multirow{4}{*}{PDCL}
                & Acc. (\%) $\downarrow$ & 74.04 & 71.55 &  \textbf{72.80} \\
                & ASR (\%) $\uparrow$ & 6.16 & 9.63 &  \textbf{7.90} \\
                & FR (\%) $\uparrow$ &  12.26 & 16.82 &  \textbf{14.54}\\\cmidrule{2-5}
                & ACR (\%) $\downarrow$ & 10.66 & 11.34 & 11.00 \\ 
            \midrule
                \rowcolor{Goldenrod!20} & Acc. (\%) $\downarrow$ & 74.17 & 71.76 & 72.97 \\
                \rowcolor{Goldenrod!20} & ASR (\%) $\uparrow$ &  5.92 & 9.32 & 7.62\\
                \rowcolor{Goldenrod!20} & FR (\%) $\uparrow$ & 11.91 & 16.43 & 14.17 \\\cmidrule{2-5}
                \rowcolor{Goldenrod!20} \multirow{-4}{*}{w/ Ours} & ACR (\%) $\downarrow$ & 10.45 & 11.20 &  \textbf{10.83} \\ 
            \bottomrule
        \end{tabular}
    }
\end{minipage}
\hspace{0.01mm}
\begin{minipage}{0.48\linewidth}

    \centering
    \caption{Real-world system evaluations (400 random images).}
    \label{tab:supp_zsclassification}
    \setlength{\tabcolsep}{15pt}    
    \renewcommand{\arraystretch}{1}
    \renewcommand{\aboverulesep}{1pt}
    \renewcommand{\belowrulesep}{1pt}
    \resizebox{\linewidth}{!}{
        \begin{tabular}{ccccccccc}
            \toprule
                \textbf{Task} & \multicolumn{4}{c}{Zero-Shot Cls. (Acc. \%)}  \\ 
            \cmidrule{2-5}
                \textbf{Data subset}
                & ImageNet-R 
                && \multicolumn{2}{c}{ImageNet-1K} \\ \cmidrule{2-2}\cmidrule{4-5}
                \textbf{Model} 
                & GPT-4o-mini 
                && GPT-4o-mini & GPT-4.1 \\
            \midrule
                 Benign 
                 & 27.25
                 && 34.25 & 17.50 \\
            \midrule
                \multirow{1}{*}{Baseline} 
                & 6.75
                && 5.00 & 3.00 \\
                \rowcolor{Goldenrod!20} \multirow{1}{*}{w/ Ours} 
                & 4.75
                && 1.25 & 1.00 \\
            
            \bottomrule
        \end{tabular}
    }
    \hspace{5cm}%
    \vspace{1em}
    \centering
    \caption{Multi-modal large language model evaluations with a prompt `\texttt{Provide a short caption for this image}'.}
    \setlength{\tabcolsep}{10pt}
    \renewcommand{\arraystretch}{1}
    \renewcommand{\aboverulesep}{1pt}
    \renewcommand{\belowrulesep}{1pt}
    \resizebox{\linewidth}{!}{
        \begin{tabular}{cccccccc}
            \toprule
                \textbf{Task} & \multicolumn{5}{c}{Image Captioning (MS COCO) on LLaVA 1.5-7B}  \\ \cmidrule{2-6}
                \textbf{Metric} & BLEU-4 & METEOR & ROUGE-L & CIDEr & SPICE \\
            \midrule
                 Benign & 34.9 & 29.1 & 58.0 & 122.7 & 22.9 \\
            \midrule
                BIA
                & 35.0 & 28.6 & 57.4 & 124.9 & 22.7 \\
                \rowcolor{Goldenrod!20} \multirow{1}{*}{w/ Ours} 
                & 34.5 & 28.4 & 57.0 & 123.1 & 22.2\\
            
            
            \bottomrule
        \end{tabular}
    }
    \label{tab:mllm_eval}
    \hspace{5cm}%
    \vspace{1em}
    \centering
    \caption{Random trials for Ours added to the baseline~\cite{zhang2022beyond}.}
    \label{tab:supp_randomtrials}
    \vspace{-0.1mm}
    \setlength{\tabcolsep}{5pt}
    \renewcommand{\arraystretch}{1}
    \renewcommand{\aboverulesep}{3pt}
    \renewcommand{\belowrulesep}{3pt}
    \resizebox{\linewidth}{!}{
        \begin{tabular}{cccccccc}
            \toprule
                Trial & Cross- & Accuracy $\downarrow$ & ASR $\uparrow$ & FR $\uparrow$ & ACR $\downarrow$ \\
            \midrule\midrule
                \multirow{4}{*}{1}
                    & Domain & 47.99 & 48.14 & 50.81 & 10.24 \\ 
                    & Model & 46.12 & 42.67 & 48.69 & 8.34  \\ \cmidrule{2-6}
                    & Task (OD) & \multicolumn{4}{c}{24.51 (mAP50)} \\
                    & Task (SS) & \multicolumn{4}{c}{23.02 (mIoU)} \\
            \midrule
                \multirow{4}{*}{2} 
                    & Domain & 47.95 & 48.19 & 50.86 & 10.31 \\ 
                    & Model & 46.26 & 42.53 & 48.49 & 8.32\\\cmidrule{2-6}
                    & Task (OD) & \multicolumn{4}{c}{24.52 (mAP50)} \\
                    & Task (SS) & \multicolumn{4}{c}{24.33 (mIoU)} \\
            \midrule
                \multirow{4}{*}{3} 
                    & Domain & 47.10 & 49.02 & 51.66 & 9.66 & \\ 
                    & Model & 45.86 & 43.04 & 49.00 & 8.28 \\\cmidrule{2-6}
                    & Task (OD) & \multicolumn{4}{c}{24.52 (mAP50)} \\
                    & Task (SS) & \multicolumn{4}{c}{23.20 (mIoU)} \\
            \midrule\midrule
                \multirow{4}{*}{\textbf{Avg.}} 
                    & Domain & \textbf{47.58} $\pm$ 0.41 & \textbf{48.45} $\pm$ 0.40 & \textbf{51.11} $\pm$ 0.39 & \textbf{10.07} $\pm$ 0.29 \\
                    & Model  & \textbf{46.08} $\pm$ 0.17 & \textbf{42.75} $\pm$ 0.22 & \textbf{48.73} $\pm$ 0.21 & \textbf{8.31} $\pm$ 0.03 \\\cmidrule{2-6}
                    & Task (OD) & \multicolumn{4}{c}{\textbf{24.52} $\pm$ 0.01 (mAP50)} \\
                    & Task (SS) & \multicolumn{4}{c}{\textbf{23.52} $\pm$ 0.58 (mIoU)} \\
            \bottomrule
        \end{tabular}
    }
\end{minipage}
\end{table*}

\paragraph{Attack robustness against real-world systems.}
In order to be on the path to the paradigm of commercial models that are currently being actively deployed, we tested our attack strategy against multi-modal large language models on zero-shot image classification and image captioning in Tables~\ref{tab:supp_zsclassification} and \ref{tab:mllm_eval}, respectively.
For the zero-shot image classification task with the well-known CLIP~\cite{CLIP} vision language model, we observe that, except for BIA~\cite{zhang2022beyond} and FACL~\cite{yang2024facl}, we observe improved attacked accuracy when we add our method to the baselines~\cite{naseer2019cross, salzmann2021learning, aich2022gama, yang2024pdcl}. We conjecture that the slight attack strength degradation owes to the baseline method that has already been well-fitted to generate adversarial examples effective for the zero-shot setting based on the relatively lower accuracy scores than the rest. We posit that the respective well-trained generator is already adept enough that our method may interfere with the learned generator weights negatively, and there may exist a maximum capacity at which AEs from generative model-based attacks can attack victim models. 
Against LLaVA 1.5-7B~\cite{liu2023llava}, our attack on the image captioning task, compared with the baseline, shows competitive attack potential. Although our attack is crafted using an image-only CNN surrogate model and its impact on similar architectures is most notable, we observe that attacking the image branch of the concurrent multi-modal models can also be a viable option for adversarial attacks. We defer this to future exploration along with the text-side attacks.
Our method thus well demonstrates the potential to further impair the recognition capabilities of large, deployed multimodal models, including vision-language systems such as LLaVA and GPT-4o.

\begin{figure}[!h]
    \centering
    \includegraphics[width=\linewidth]{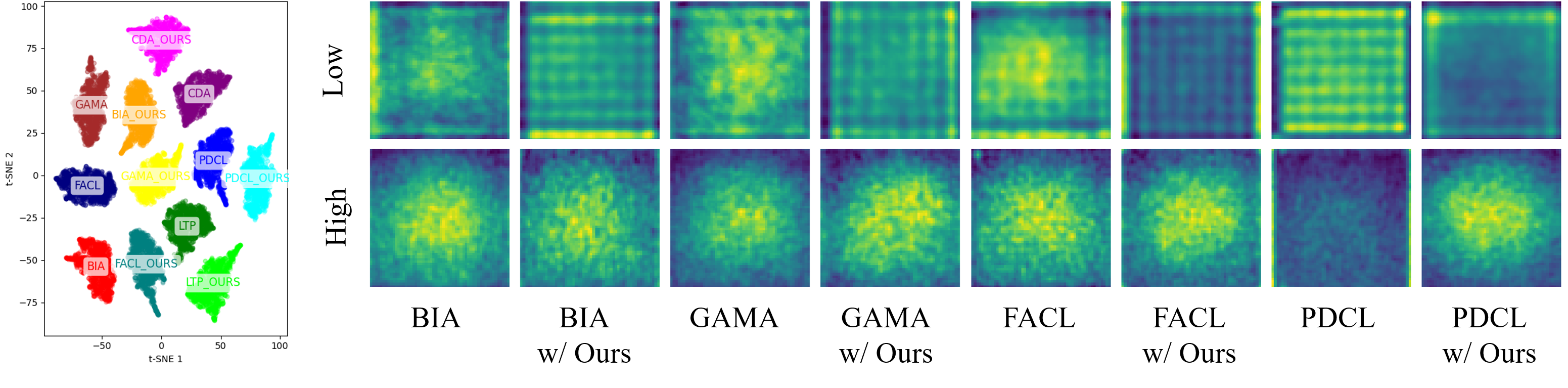}
    \caption{
    \textbf{Left}: t-SNE visualization of the intermediate generator features after \texttt{resblock3} for all methods, showing how the additions of Ours (SCGA) shift the internal feature geometry. The largest displacement appears for our primary baseline, BIA. \textbf{Right}: Low- and extreme high-band components of the same intermediate features, reconstructed in the image space by applying radial masks to their 2-D FFTs ($\rho<0.2$ and $\rho\ge 0.8$) and averaging over the validation set. For each baseline and its SCGA-augmented variant, SCGA yields low-band reconstructions with brighter and more compact regions and high-band reconstructions that are less peaky and more spatially dispersed. This joint behavior agrees with the spectral ratios in Table~6 and indicates a redistribution of energy toward semantically aligned low frequencies together with attenuation of sharp surrogate-specific high-frequency artifacts.}
    \label{fig:lowhigh_freq}
\end{figure}

\paragraph{
Spectral energy comparison by band.}
We emphasize that the spectral ratios in Table~6 and the band-wise t-SNE~\cite{maaten2008visualizingtsne} visualizations of generator intermediate features (after \texttt{resblock3}) in Fig.~\ref{fig:lowhigh_freq} are obtained in two different but complementary ways. For Table~6, we work purely in the frequency domain: for each generator output, we compute the 2D FFT, partition the spectrum into a low band ($\rho<0.2$) and its complement ($\rho\ge 0.2$) based on the normalized radial distance~$\rho$ from the DC component, and then integrate the power over each band. The reported numbers are therefore scalar ratios of Fourier-domain energy between low and the remaining higher frequencies. In contrast, for Fig.~\ref{fig:lowhigh_freq} we first apply radial masks to the FFT of the intermediate feature maps after \texttt{resblock3}, isolate either the low band ($\rho<0.2$) or an extreme high band ($\rho\ge 0.8$), and then perform an inverse FFT. This yields band-limited reconstructions back in the image space of the features, which we average over the validation set. The visualization then shows the spatial low- and extreme-high-frequency content, rather than showing raw power spectrum of each band.

The qualitative effects in Fig.~\ref{fig:lowhigh_freq} are consistent with the quantitative trends in Table~6. Across baselines, SCGA increases the low-band energy ratio and decreases the high-band ratio in Table~6. In the visualization, this appears as low-band reconstructions with brighter and more compact regions for SCGA, compared to the more diffuse patterns of the baselines. These bright areas indicate that a larger fraction of feature variance is carried by smoothly varying, semantically aligned components in the low band, which explains the higher low-band energy ratios. In contrast, the high-band reconstructions with SCGA are less peaky and more spatially dispersed, with fewer intensely bright spots. This visual pattern matches the reduced and less concentrated high-frequency energy in Table~6 and suggests that sharp surrogate-specific artifacts are attenuated and spread out. The PDCL case is a mild exception. Table~6 still reports a net shift of energy toward the low band, but in Fig.~\ref{fig:lowhigh_freq} the remaining extreme-high-frequency content becomes slightly more localized, which is consistent with the relatively modest gains of PDCL+SCGA in the transferability results.

\begin{figure}[!h]
    \centering
    \includegraphics[width=\linewidth]{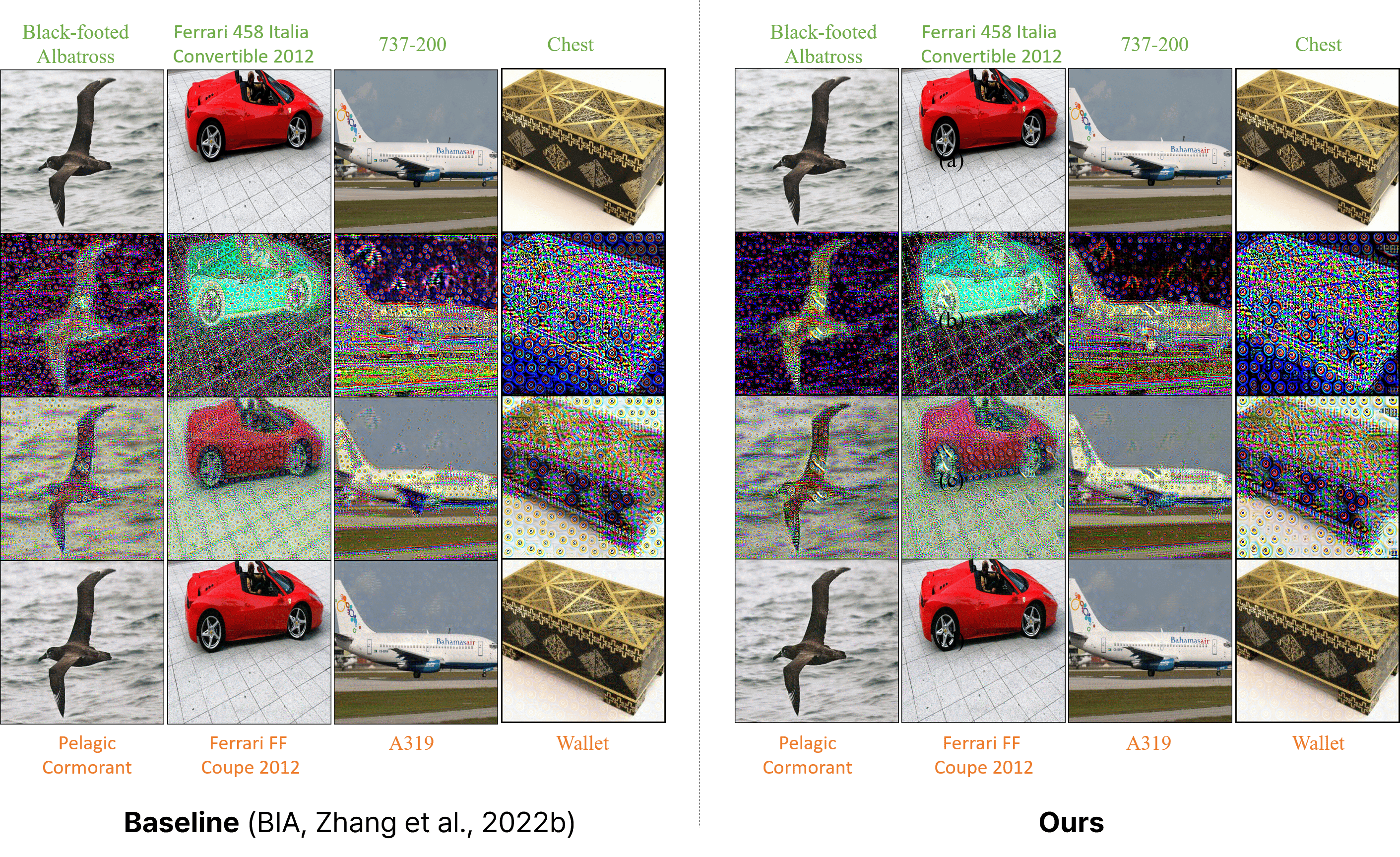}
    \caption{
    Side-by-side visualization of results on CUB-200-2011, Stanford Cars, FGVC Aircraft, and ImageNet. The attacked predictions are similar, yet the perturbation patterns are visibly different.}
    \label{fig:sidebyside_viz}
\end{figure}

\paragraph{
Side-by-side visualization with the baseline.}
To directly compare our method against the baseline qualitatively, we visualize them in Fig.~\ref{fig:sidebyside_viz}. Here, we observe that the predictions after each attack are highly similar, yet the object-aligned patterns in the perturbations are vastly different. Our perturbation demonstrates more vivid perturbations concentrate on the foreground regions without blurs in the noise pattern, suggesting that our intended object-aligned perturbations improved compared to those of the baseline.

\subsection{Limitations and Broader Societal Impacts}
Our method exposes vulnerabilities in generative attack pipelines, yet its transferability gains remain bounded by the underlying generator architecture. 
By revealing these constraints in publicly available generative models, we contribute to exposing safety vulnerabilities of neural networks. The demonstrated transferability of generator internal semantic-aware perturbations underscores the need for adversarial robustness and motivates integrating safety measures, such as early‐block regularization or semantic‐consistency checks, into future network designs. Moreover, our approach targeting the adversarial perturbation process directly differs in principle from those that explicitly target benign-adversarial divergence in the surrogate model level. Therefore, our method stands as a compatible method to enhance those methods further, not to be assessed on the same grounds.

\end{document}